%% file: arxiv.tex
\pdfoutput=1

\documentclass{article} %
\usepackage{amd}

\newcommand{\AMDlogo}{\includegraphics[height=0.28in]{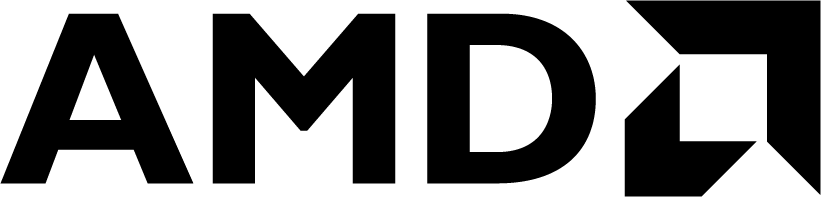}}

\input{math_commands.tex}

\usepackage{hyperref}
\usepackage{url}
\usepackage{graphicx}
\usepackage{booktabs}
\usepackage{multirow}
\usepackage{colortbl}
\usepackage{flafter}
\usepackage{placeins}
\usepackage{enumitem}
\definecolor{zebralavender}{HTML}{F1ECFA}
\definecolor{modelgray}{HTML}{E8EAED}

\newif\ifrevhl
\revhlfalse
\definecolor{revblue}{RGB}{0,90,170}
\newenvironment{revision}{\ifrevhl\color{revblue}\fi}{}

\newcommand{\ruijing}[1]{{#1}}
\newcommand{\method}{\textsc{Zebra3D }}
\newcommand{\methodns}{\textsc{Zebra3D}}
\newcommand{\eqcontribsymbol}{\textcolor{amd_gold}{\ensuremath{\clubsuit}}}
\newcommand{\corrsymbol}{{\boldmath$\ddagger$}}

\begin{document}
\onecolumn
\amdtitlerunning{Zebra3D: Efficient Hybrid Foundation
Models For Scalable 3D Reconstruction}

\amdtitle{%
\makebox[\textwidth][c]{%
    \raisebox{-0.3\height}{%
        \includegraphics[height=1.35cm]{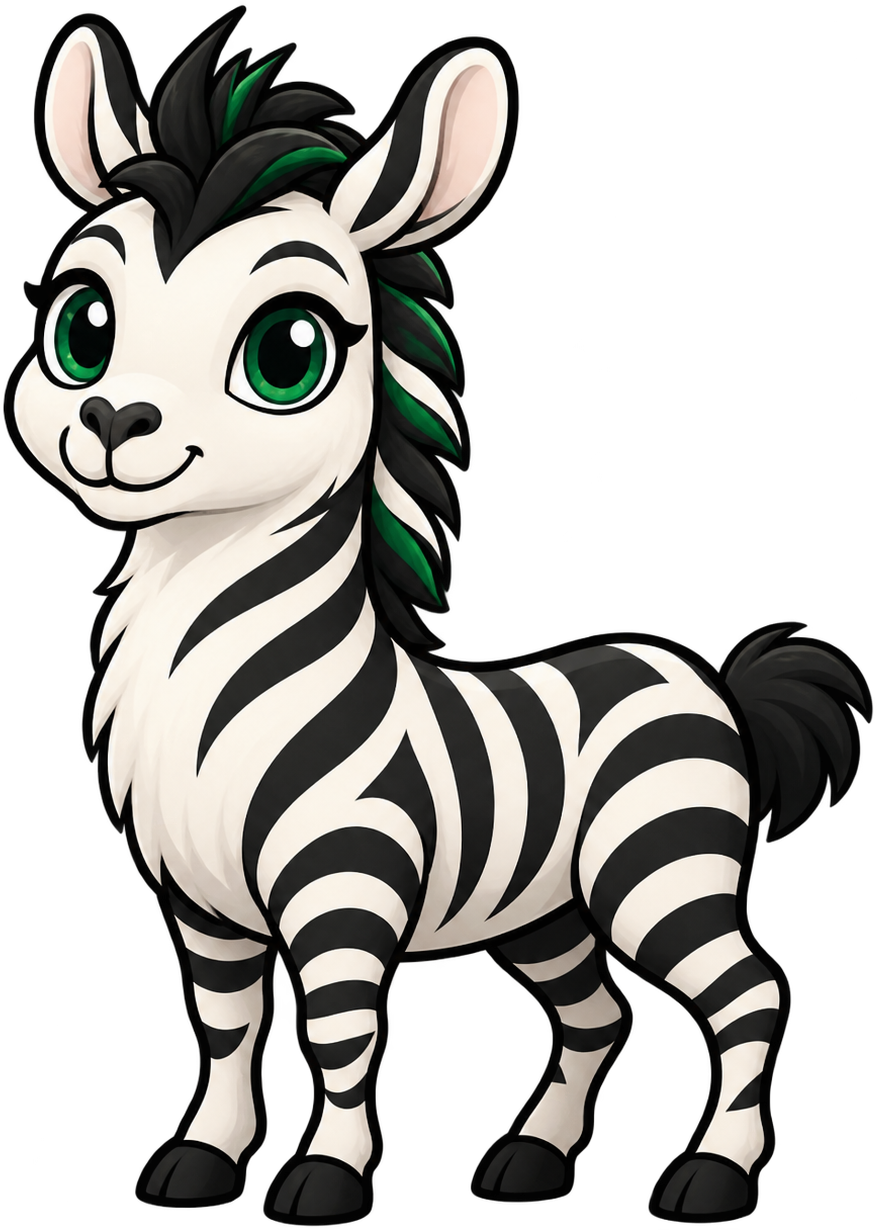}%
    }%
    \begin{tabular}{c}
        \textsc{Zebra3D}: Efficient Hybrid Foundation \\ Models For Scalable 3D Reconstruction
    \end{tabular}%
}%
}

\vspace*{-0.5\baselineskip}
{\centering\setlength{\parskip}{0pt}%
  {\bfseries
    Utkarsh Saxena\textsuperscript{\eqcontribsymbol\,\corrsymbol}\qquad
    Ruijing Yang\textsuperscript{\eqcontribsymbol\,\corrsymbol}\qquad
    Mehdi Rezagholizadeh\textsuperscript{\corrsymbol}\\[0.3em]
    Akash Haridas\qquad
    Vikram Appia\qquad
    Emad Barsoum\par}
  \vspace{0.75em}
  Advanced Micro Devices Inc. (AMD)\par
  \vspace{0.6em}
  {\small
    \textsuperscript{\eqcontribsymbol}\,Equal contribution\par
    \vspace{0.15em}
    \textsuperscript{\corrsymbol}\,Correspondence to:
    \texttt{\char`\{utkarsh.saxena, ruijing.yang, mehdi.rezagholizadeh\char`\}@amd.com}\par}
}
\amdkeywords{3D reconstruction, Sparse attention, Linear attention, knowledge distillation}
\vskip 0.15in

\newcommand{\fix}{\marginpar{FIX}}
\newcommand{\new}{\marginpar{NEW}}

\begin{abstract} 
Feed-forward 3D foundation models reconstruct scenes from unposed images, but the quadratic cost of global attention limits their scalability to large view sets. We present \textsc{Zebra3D}, a layerwise sparse--linear hybrid attention framework that improves the efficiency of pretrained 3D foundation models without training from scratch. Across VGGT, HunyuanWorld-Mirror, and $\pi^3$, we observe a broadly consistent depth-wise pattern: early stages are relatively tolerant to linear replacement, while sensitive middle-to-late stages tend to exhibit more concentrated attention. \textsc{Zebra3D} exploits this structure by assigning linear-centered attention to tolerant stages and tile-sparse attention to sensitive stages, using landmark summaries to preserve unselected context. Closed-form affine fitting enables training-free conversion, while optional post-conversion training combines task supervision with output and relational feature distillation. Across all three backbones, \textsc{Zebra3D} achieves strong reconstruction and camera-pose accuracy at high attention sparsity, with training-free variants remaining close to their dense counterparts. End-to-end speedups increase with input view count, reaching up to $7.1\times$ at 1,024 views. After post-conversion training, \textsc{Zebra3D}-VGGT reaches 80.5\% RealEstate10K AUC@30 at 94\% mean attention sparsity, within 1.1 percentage points of dense VGGT fine-tuned on the same data and schedule.

\end{abstract}

\section{Introduction}
Recovering scene geometry and camera poses from images is fundamental to 3D vision, supporting autonomous systems, immersive AR/VR
\citep{cadena2016slam,marchand2016pose}, and providing geometric representations for 3D world modeling \citep{team2026hy}. Classical structure-from-motion (SfM) and multi-view stereo pipelines combine correspondence estimation, camera optimization, and dense reconstruction \citep{schoenberger2016sfm,schoenberger2016mvs}. Neural scene representations such as NeRF require iterative per-scene optimization, typically from images with known camera poses~\citep{mildenhall2021nerf}.
The computational cost of scene-specific optimization motivates a shift toward feed-forward prediction. DUSt3R directly regresses dense point maps from image pairs, but reconstructing an image collection still requires global alignment~\citep{wang2024dust3r}. Recent feed-forward 3D (FF3D) foundation models, including VGGT~\citep{wang2025vggt}, HunyuanWorld-Mirror~\citep{liu2026worldmirror}, and $\pi^3$~\citep{wang2026pi3}, jointly predict camera poses and dense scene geometry from multiple unposed views in a single forward pass, enabling reconstruction without per-scene optimization. These models alternate frame-wise attention with global attention for cross-view information exchange, allowing each image to benefit from observations elsewhere in the scene.

\begin{figure}[t]
    \centering
    \includegraphics[width=\linewidth]{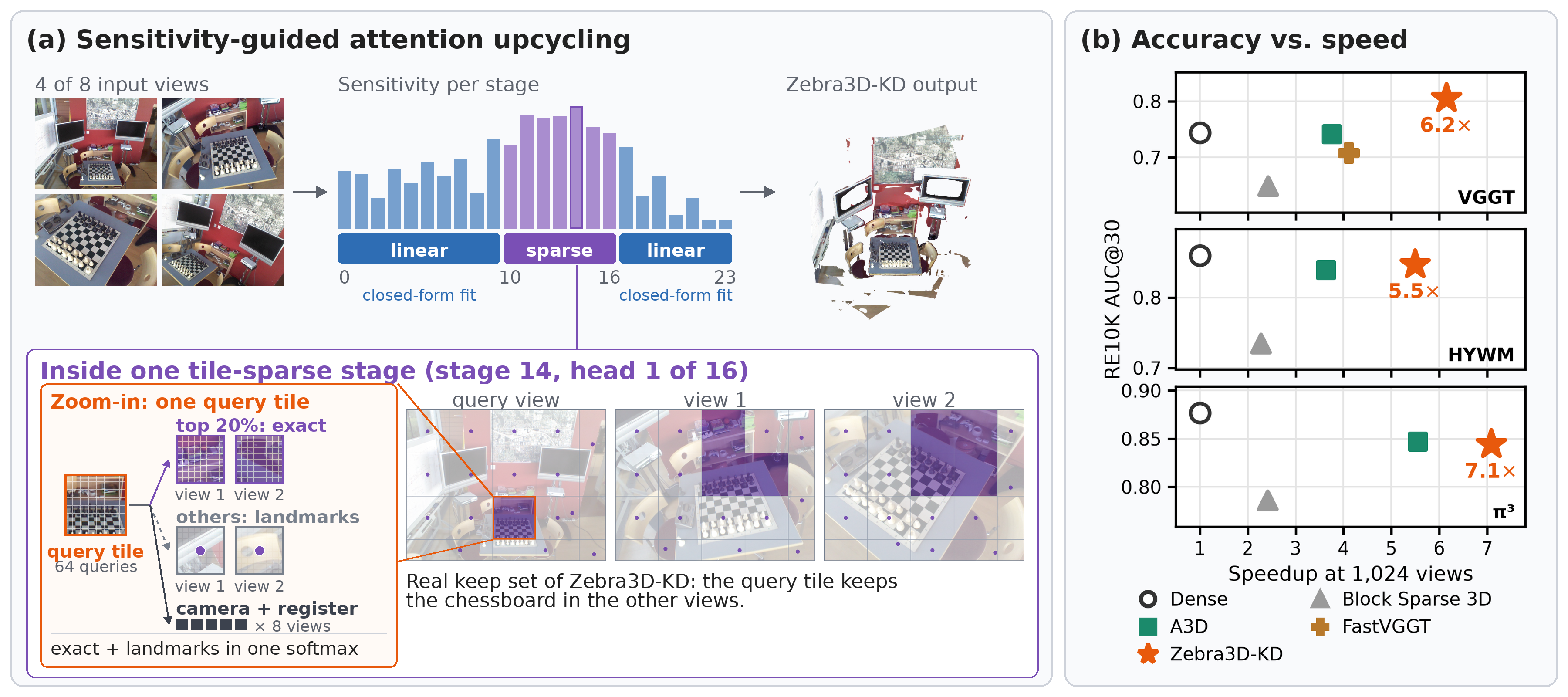}
    \caption{\textbf{Zebra3D overview.}
    \textbf{(a)} Sensitivity-guided assignment of linear and tile-sparse
    attention. Selected tiles retain full softmax interactions; omitted tiles
    become mean-key/value landmarks.
    \textbf{(b)} RE10K pose AUC@30 versus end-to-end speedup at $1{,}024$
    views.}
    \label{fig:overview}
\end{figure}

\begin{figure}[t]
    \centering
    \begin{minipage}[t]{0.33\linewidth}
        \centering
        \includegraphics[width=0.9\linewidth]{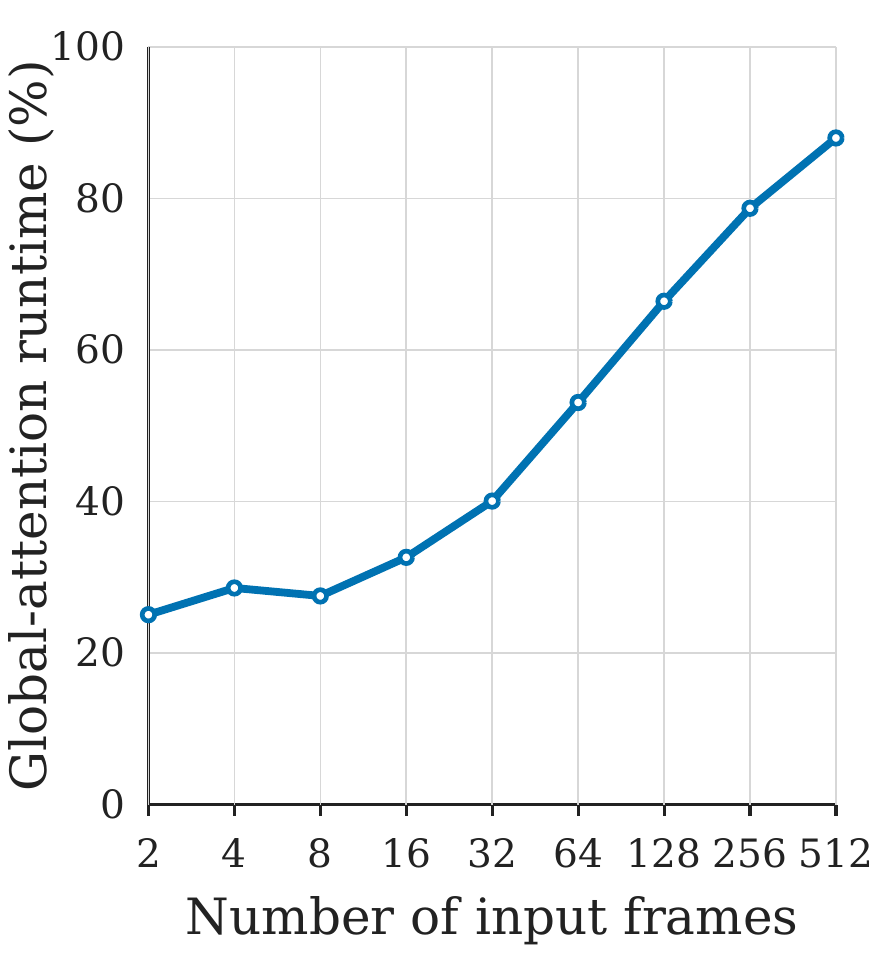}
        \smallskip
        \textbf{(a)} Global-attention runtime.
    \end{minipage}
    \hfill
    \begin{minipage}[t]{0.32\linewidth}
        \centering
        \includegraphics[width=0.9\linewidth]{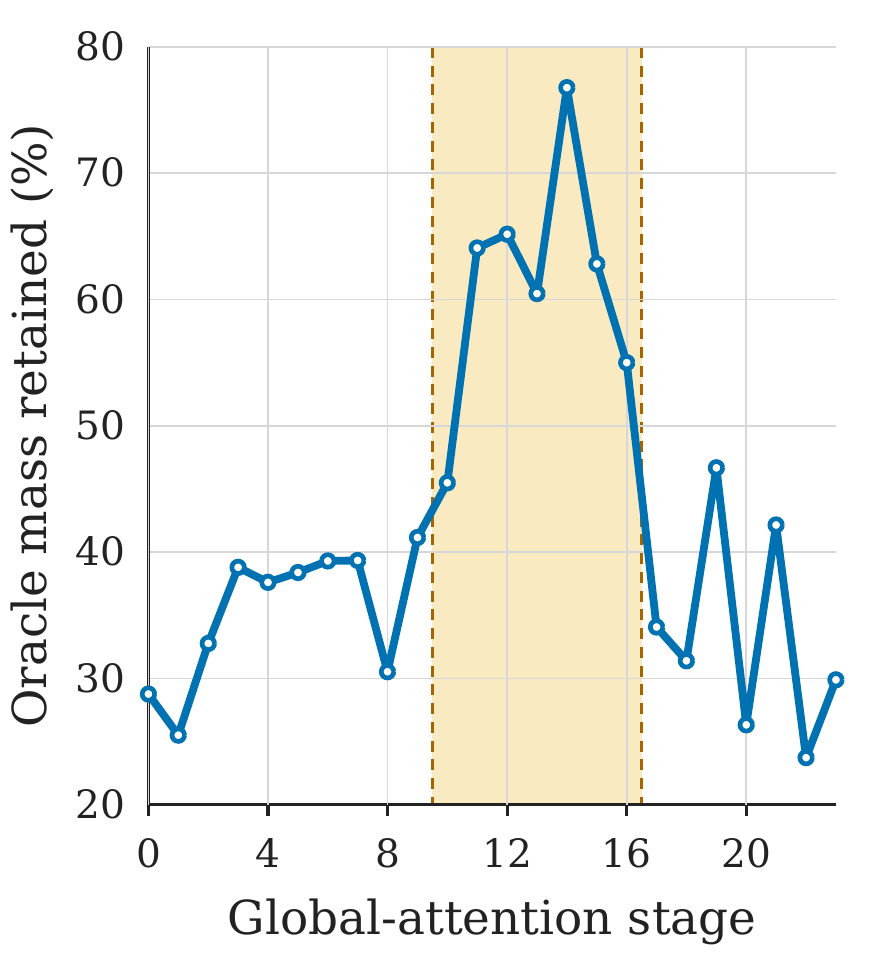}
        \smallskip
        \textbf{(b)} Oracle top-$20\%$ block mass.
    \end{minipage}
    \hfill
    \begin{minipage}[t]{0.32\linewidth}
        \centering
        \includegraphics[width=0.9\linewidth]{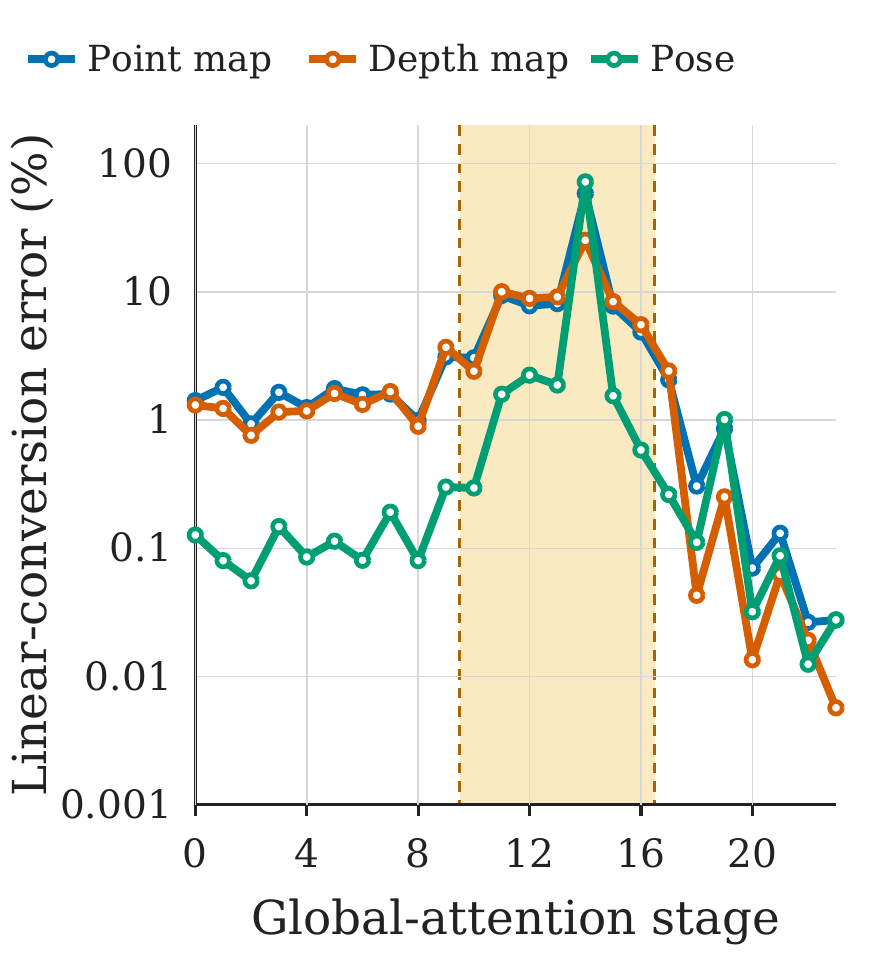}
        \smallskip
        \textbf{(c)} Per-stage teacher deviation.
    \end{minipage}
    \caption{\textbf{Measurements motivating the Zebra3D design.} (a) Global attention
    reaches $88.0\%$ of VGGT runtime at 512 views. (b) Attention Mass retained by an
    oracle Top-$20\%$ key/value blocks. (c) Relative
    output deviation after converting one global-attention stage to 
    linear attention.}
    \label{fig:background-efficiency}
\end{figure}

 Eliminating per-scene optimization, however, does not eliminate the computational bottleneck: dense global attention scales quadratically with the number of input views. In our measurements, global attention accounts for $88\%$ of VGGT's end-to-end runtime at $512$ views (Figure~\ref{fig:background-efficiency}a). Existing acceleration methods reduce this cost by merging image tokens, subsampling key/value context, or replacing dense attention with sparse or linear operators \citep{shen2026fastvggt,sun2026avggt,zhang2026lite3r,wang2026blocksparse}.
Among attention replacements, sparse and linear attention offer complementary ways to reduce this cost. Sparse attention preserves token-level softmax interactions over selected regions, whereas linear attention aggregates global context at linear cost. Applying either operator uniformly, however, does not account for differences in how pretrained stages respond to approximation. This motivates a central question: can the pretrained model itself reveal which stages tolerate linear replacement and which require more faithful softmax interactions?

We investigate this question by profiling every global-attention stage in pretrained VGGT, HunyuanWorld-Mirror (HYWM), and $\pi^3$ using two independent diagnostics: sensitivity to linear conversion and attention concentration. Across all three backbones, we observe a recurring depth-wise structural pattern: the middle stages are especially sensitive to linear replacement, yet their attention is highly concentrated.  In the preceding and following stages, attention mass is distributed across many tokens, and these stages are surprisingly tolerant to
linear approximation.  This alignment turns an apparent trade-off into a
common allocation principle:
use sparse attention where concentrated interactions matter and linear
attention everywhere else.

Building on this insight, we introduce \textsc{Zebra3D}, a sensitivity-guided framework for upcycling pretrained dense FF3D models
(Figure~\ref{fig:overview}). For conversion-tolerant stages, we introduce \emph{linear-centered attention}, which reuses pretrained projections and applies a centered affine correction fitted to the dense operator by a closed-form least-squares solve per stage. For sensitive stages, \emph{tile-sparse attention} evaluates selected spatial tiles with softmax and summarizes omitted tiles with landmarks to preserve global context. Together, the two operators replace the entire dense global-attention stack. The converted model supports immediate use without gradient updates. To further improve the quality--efficiency trade-off, we introduce a lightweight refinement which combines task supervision with distillation of outputs, centered features, and sampled cross-view relations from the frozen dense teacher.

Across three backbones and multiple benchmarks, \textsc{Zebra3D} provides strong accuracy--efficiency trade-offs. At $79\%$ mean attention sparsity, training-free variants remain within $0.4$--$0.7$ percentage points of dense RealEstate10K pose AUC@30 and achieve the highest CO3Dv2 AUC@30 among the evaluated non-dense methods for each backbone. At $1{,}024$ views, \method delivers end-to-end speedups of $6.2\times$, $5.5\times$, and $7.1\times$ on VGGT, HYWM, and $\pi^3$, respectively. After $3{,}000$ refinement steps, \method-VGGT achieves $80.5\%$ RealEstate10K pose AUC@30 at $94\%$ mean attention sparsity, within $1.1$ percentage
points of dense VGGT fine-tuned on the same data and training schedule.

Our contributions are:
\begin{itemize}
    \item An empirical analysis of depth-wise sensitivity to linear replacement, showing that sensitive stages tend to exhibit more concentrated attention, and a sensitivity-guided strategy that assigns linear attention to tolerant stages and sparse attention to sensitive ones.
    \item A pretrained-model conversion framework combining linear-centered attention and tile-sparse attention with landmark summaries. Closed-form affine fitting enables training-free conversion, while optional post-conversion training incorporates output and relational feature distillation.
    \item Validation across three 3D backbones, demonstrating strong accuracy at high attention sparsity and $5.5$--$7.1\times$ end-to-end speedups at $1{,}024$ views with packed Triton kernels.
\end{itemize}

\section{Key Related Works}

\textbf{Token and context compression.}
FastVGGT, LiteVGGT, HTTM, and Co-Me reduce token redundancy through
saliency-, geometry-, temporal-, or confidence-guided merging
\citep{shen2026fastvggt,shu2026litevggt,wang2026httm,chen2026come}.
AVGGT combines early frame attention with key/value subsampling
\citep{sun2026avggt}, FlashVGGT compresses views into descriptor tokens
\citep{wang2026flashvggt}, and RegimeVGGT combines merging and
key/value downsampling across depth \citep{you2026regimevggt}.

\textbf{Efficient attention and representations.}
Block-Sparse VGGT evaluates selected attention blocks
\citep{wang2026blocksparse}, Lite3R distills sparse linear attention
\citep{zhang2026lite3r}, and Speed3R combines compressed context with
selected full-resolution interactions \citep{ren2026speed3r}.
VGG-T$^3$ instead adapts a fixed-size scene MLP at test time
\citep{elflein2026vggt3}.
\method uses measured conversion sensitivity to assign linear and
tile-sparse attention to different stages of a pretrained model. Detailed related works can be found in Appendix~\ref{app:extended_related_works}.

\section{Method}
\label{sec:method}
\method converts pretrained FF3D models into sparse–linear hybrids, assigning tile-sparse attention to stages sensitive to linearization and linear-centered attention to tolerant stages (Figure 2b,c). Here, stages denote global-attention layers and are indexed independently of interleaved frame-attention layers. We describe the allocation, both operators, and relational distillation below.

\subsection{Sensitivity-Guided Stage Allocation}
\label{sec:structure}

\textbf{Measuring sensitivity.}
We replace one global-attention stage $\ell$ at a time with the fitted
linear-centered operator described in Section~\ref{sec:linear}, keeping all other
stages unchanged. For each output modality $u$, we measure the deviation
from the dense teacher over a validation set $\mathcal{D}$:
\begin{equation}
 e_{\ell}^{(u)}
 = \frac{1}{|\mathcal{D}|}
   \sum_{x\in\mathcal{D}}
   d_u\!\left(\widehat{y}_{\ell}^{(u)}(x),y_T^{(u)}(x)\right),
 \qquad
 r_{\ell}
 = \frac{1}{|\mathcal{U}|}
   \sum_{u\in\mathcal{U}}
   \operatorname{rank}_{\ell}\!\left(e_{\ell}^{(u)}\right),
 \label{eq:layer-conversion-error}
\end{equation}
where $d_u$ is the modality-specific discrepancy and lower ranks indicate
smaller deviations. We average ranks rather than raw errors to combine
modalities with different units and scales. On VGGT, \ruijing{the output deviations from linear replacement
peak in stages $11$--$15$ and fall off toward both ends}
(Figure~\ref{fig:background-efficiency}c).

\begin{figure}[t]
    \centering
    \includegraphics[width=0.8\linewidth]{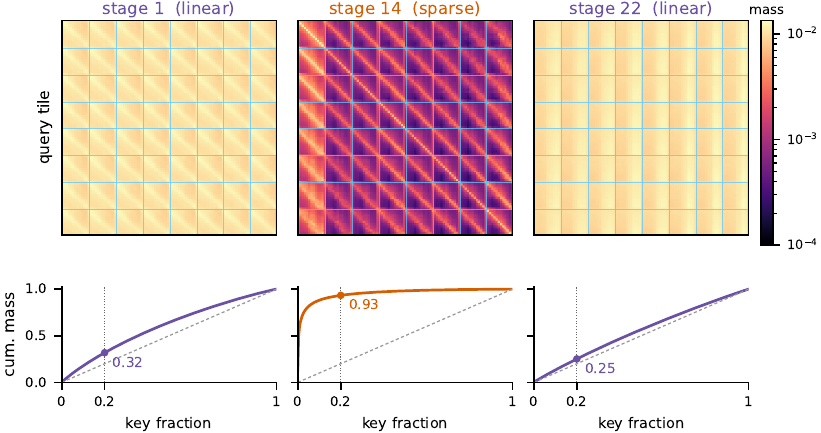}
    \caption{ \textbf{Attention feature map across depth (VGGT).} Top: attention mass on one calibration scene, with cyan lines at view boundaries. Bottom: cumulative attention mass, with the value at a $20\%$ key budget marked.
    A linear tolerant stage (1, 22) exhibits more diffuse attention while stage in the sensitive band (14)
    concentrates on few tokens. More visualizations in Figure~\ref{fig:appendix-attention-maps}}
    \label{fig:attention-maps}
\end{figure}

We measure concentration by the dense-attention mass retained by oracle
selection of the top $20\%$ of 64-token key/value blocks. Our key observation is that \emph{stages sensitive to linear conversion retain more attention mass, whereas tolerant stages approach uniform attention},
supporting sparse and linear attention, respectively
(Figures~\ref{fig:background-efficiency}b and~\ref{fig:attention-maps}).
This pattern extends to HYWM and $\pi^3$ (Figure~\ref{fig:appendix-background-additional}).

\textbf{Choosing the allocation.}
For a target linear fraction $p$, we assign the $K=\operatorname{round}(pL)$ stages with the lowest sensitivity scores $r_\ell$ to linear attention and the remaining stages
to tile-sparse attention. Although this ranking does not account for interactions between simultaneous replacements, Table~\ref{tab:attention-operator-ablations}(d)
shows that this allocation works well in practice and achieves good performance.

\subsection{Linear-Centered Attention with Closed-Form Initialization}
\label{sec:linear}
In conversion-tolerant stages, we replace softmax attention with a
linear operator, reusing the pretrained QKV projections,
positional encoding, normalization, and output projection from pretrained model.
For one head of width $d_h$, let $\phi$ apply \texttt{softmax} along the channel
dimension. We compute
\begin{equation}
 O_{\mathrm{lin}}
 = \frac{\phi(Q)\left(\phi(K)^\top V\right)}
        {\phi(Q)\left(\phi(K)^\top\mathbf{1}\right)+\epsilon},
 \label{eq:linear-attention}
\end{equation}
with row-wise division. Linear attention avoids quadratic compute of softmax attention and requires $O(Nd_h^2)$ computation. Further, to compensate for the mismatch with pretrained softmax attention, we fit
an affine map ($P_{\ell}$) and apply it to the
residual around the value mean:
\begin{equation}
 \overline{v}=\frac{1}{N}\mathbf{1}^\top V,
 \qquad \overline{V}=\mathbf{1}\overline{v},
 \qquad
 \widetilde{O}_{\ell}
 = \overline{V}+(O_{\mathrm{lin}}-\overline{V})P_{\ell}
   +\mathbf{1}b_{\ell}^{\top}.
 \label{eq:linear-centered}
\end{equation}
Centering around mean of values decomposes the output into the global value mean and token-specific residuals. Uniform attention reduces exactly to value-mean pooling, so applying the fitted projection only to the residuals anchors the approximation at this natural baseline while correcting fine-grained token differences. This performs better than fitting an affine map directly to the uncentered output. When $P_\ell=I$ and $b_\ell=0$, we recover raw linear attention; when $P_\ell=0$ and $b_\ell=0$, we recover value-mean pooling.

\textbf{Closed-form fitting.}
Using frozen-teacher activations from a calibration dataset, we pool tokens across scenes
and heads and form $X_\ell=[O_{\mathrm{lin}}-\overline{V},\mathbf{1}]$
and $Y_\ell=O_{\mathrm{softmax}}-\overline{V}$.
With $\Theta_\ell=[P_\ell;b_\ell^\top]$, ridge regression gives
\begin{equation}
 \Theta_\ell^*
 = \arg\min_{\Theta}
   \left\|X_\ell\Theta-Y_\ell\right\|_F^2
   +\lambda_{\mathrm{fit}}\|\Theta\|_F^2
 = \left(X_\ell^\top X_\ell+\lambda_{\mathrm{fit}}I\right)^{-1}
   X_\ell^\top Y_\ell.
 \label{eq:linear-projection-fit}
\end{equation}
One solve per converted stage initializes the correction without
backpropagation or updates to inherited weights. Fitting uses $13$
calibration scenes and a float64 solve
(Appendix~\ref{app:linear-init}).

\subsection{Tile-Sparse Attention with a Landmark Tail}
\label{sec:tile}
Stages sensitive to linear conversion retain token-level softmax interactions through tile-sparse attention (Figure~\ref{fig:overview}a). 
We reshape each view’s patch tokens onto their 2D image grid and partition the grid into non-overlapping $8\times8$ tiles ($B=64$ tokens). Each tile therefore represents a contiguous image region, unlike blocks of consecutive positions in the flattened multi-view sequence, which may not form a coherent spatial neighborhood.

\textbf{Selecting exact interactions.}
For each head, let $a$ index a query tile and $b$ a key tile, with mean query $\bar{\mathbf q}_a$ and mean key $\bar{\mathbf k}_b$. We score tile pairs and select the highest-scoring key tiles:
\begin{equation}
 r_{ab}=\frac{\overline{q}_a^\top\overline{k}_b}{\sqrt{d_h}},
 \qquad
 \mathcal{S}_a=\operatorname{TopK}_{b}(r_{ab};\rho),
 \label{eq:tile-selection}
\end{equation}
where the retention ratio $\rho$ sets the tile budget $k$. Queries in tile $a$ share $\mathcal S_a$, but evaluate individual query--key scores within the selected tiles.

\textbf{Preserving unselected context.}
Each unselected tile contributes a landmark consisting of its mean key $\bar{\mathbf k}_b$, mean value $\bar{\mathbf v}_b$, and valid patch-token count $n_b$. For patch query $i$ in tile $a(i)$, let $\mathcal E_i$ contain the selected patch keys, all special-token
keys, and its self key. The mean-landmark operator is:
\begin{equation}
\begin{aligned}
 Z_i
 &= \sum_{j\in\mathcal{E}_i}e^{s_{ij}}
    +\sum_{b\notin\mathcal{S}_{a(i)}}n_b e^{\overline{s}_{ib}},\\
 \widetilde{o}_i
 &= \frac{
       \sum_{j\in\mathcal{E}_i}e^{s_{ij}}v_j
       +\sum_{b\notin\mathcal{S}_{a(i)}}
          n_b e^{\overline{s}_{ib}}\overline{v}_b
      }{Z_i},
\end{aligned}
\label{eq:landmark-tail}
\end{equation}
where $s_{ij}=q_i^\top k_j/\sqrt{d_h}$ and
$\overline{s}_{ib}=q_i^\top\overline{k}_b/\sqrt{d_h}$.
The occupancy factor $n_b$, implemented as a $\log n_b$ logit offset,
accounts for the number of summarized tokens. Selected tokens and landmarks share one softmax normalization, preserving detailed interactions in selected regions and compressed
context elsewhere. Special-token queries attend densely. If a self key lies in an unselected tile, its explicit contribution supplements that tile's unchanged landmark. 

\textbf{Computational cost.}
Ignoring boundary and special-token terms, selected tiles require approximately $\rho N^2$ token-pair evaluations, landmarks require $(1-\rho)N^2/B$ query--landmark evaluations, and tile scoring uses $(N/B)^2$ descriptor pairs. Sparse stages substantially reduce the cost of global attention at fixed $\rho$ and $B$. We implement a packed tile-sparse attention kernel in Triton for efficient implementation without materializing the dense attention matrix (Appendix~\ref{app:triton}).

\subsection{Lightweight Refinement via Knowledge Distillation}
\label{sec:relkd}
The converted model can be used without gradient updates. Optional refinement combines task supervision with knowledge distillation from the frozen dense model. To preserve cross-view feature relations, we apply additional supervision at sampled linear-centered stages. Patch features are mean-centered within each view and $\ell_2$-normalized to obtain $z$. A cosine loss $\mathcal{L}_{\mathrm{cos}}$ aligns corresponding student and teacher tokens, while a relational loss matches cross-view similarities:
\begin{equation}
 \mathcal{L}_{\mathrm{rel}}
 = \mathbb{E}_{(a,c)\in\mathcal{P}}
   \left[\operatorname{SmoothL1}
   \left(z_{S,a}^{\top}z_{S,c},z_{T,a}^{\top}z_{T,c}\right)\right].
 \label{eq:relkd}
\end{equation}
Here, $S$ and $T$ denote student and teacher, and $\mathcal{P}$ contains token pairs $(a,c)$ from different views. For each anchor $a$, candidates $c$ include both teacher-selected high-similarity tokens and randomly sampled tokens from other views. Stage and pair sampling avoid constructing dense relation matrices.

Complementing these linear-stage losses, intermediate-feature distillation $\mathcal{L}_{1}$ supervises backbone-specific feature taps, including tile-sparse stages, while output distillation $\mathcal{L}_{\mathrm{out}}$ matches final predictions shared by student and teacher. Both use teacher-magnitude-normalized $\ell_1$ matching.
The refinement objective is
\begin{equation}
\begin{aligned}
 \mathcal{L}
 &= \mathcal{L}_{\mathrm{task}}
   +\lambda_{\mathrm{KD}}(t)
    \left(\mathcal{L}_{\mathrm{out}}+\mathcal{L}_{\mathrm{feat}}\right),\\
 \mathcal{L}_{\mathrm{feat}}
 &= 0.25\mathcal{L}_{1}
   +0.50\mathcal{L}_{\mathrm{cos}}
   +0.25\mathcal{L}_{\mathrm{rel}},
\end{aligned}
\label{eq:kd-objective}
\end{equation}
where $\mathcal{L}_{\mathrm{task}}$ provides backbone-specific geometry supervision. Refinement lasts $3{,}000$ optimizer updates, with $\lambda_{\mathrm{KD}}$ decreasing from $1.0$ to approximately $0.6$. Distillation requires an additional frozen-teacher forward pass during training but adds no inference-time computation. Exact loss definitions, feature taps, sampling, and optimization
settings are given in Appendices~\ref{app:relkd} and~\ref{app:kd-training}.

\section{Experiments}
\label{sec:experiments}
\subsection{Experimental Setup}

\textbf{Models and training.}
We use released VGGT, HYWM, and $\pi^3$ checkpoints
\citep{wang2025vggt,liu2026worldmirror,wang2026pi3}.
Training-free \method uses 13 held-out calibration scenes for
closed-form projection fitting without gradient updates.
\methodns-KD starts from the converted model and refines all student
parameters for 3,000 optimizer updates using a frozen dense teacher
(Section~\ref{sec:relkd}).
All training runs use 20\% subset of training data used in HunyuanWorld-Mirror
training, with BF16 mixed precision on eight MI355X GPUs
and at most 16 views per GPU microbatch.

\textbf{Evaluation.}
We evaluate reconstruction on 7Scenes, NRGBD, and DTU
\citep{shotton2013seven,azinovic2022neuralrgbd,anaes2016dtu},
and camera pose on RealEstate10K and CO3Dv2 using ten-view inputs
\citep{zhou2018stereomagnification,reizenstein2021co3d}.
Long-sequence reconstruction is evaluated on ScanNet-50 \citep{dai2017scannet} with
100--1,000 views. All evaluation datasets are strictly excluded from the refinement training mixture except CO3Dv2, for which we use disjoint training and test splits. To evaluate reconstruction beyond densely sampled video sequences, we additionally evaluate wide-baseline reconstruction on Tanks and Temples and ETH3D
\citep{knapitsch2017tanks,schoeps2017eth3d} at $N\in\{8,16,32,64\}$ views.
Timing uses compiled BF16 inference on one MI355X (Appendix~\ref{app:runtime}).

\textbf{Baselines.}
We compare against the original dense backbones, FastVGGT
\citep{shen2026fastvggt}, SAF3R \citep{deng2026saf3r},
RegimeVGGT \citep{you2026regimevggt}, and Lite3R
\citep{zhang2026lite3r}.
We additionally evaluate architecture-matched implementations of
AVGGT-f4 \citep{sun2026avggt} and Block-Sparse VGGT
\citep{wang2026blocksparse}, denoted A3D and Block Sparse 3D,
respectively.
\ruijing{In addition, we fine-tune the dense VGGT backbone
on the same training data, for the same $3{,}000$ steps,
and with the same ground-truth supervision used for refinement
(Appendix~\ref{app:kd-training}).  We denote this control VGGT + GT-FT
and report it alongside the dense backbones.}
Unsupported backbone--baseline combinations are omitted.
Further details are in Appendix~\ref{app:implementation-details}.

\textbf{Configuration notation.}
$x$L/$y$S uses linear attention in fraction $x$ of stages and omits fraction
$y$ of key tiles in the remaining sparse stages. Thus, $0.5$L/$0.8$S is half
linear and retains the top $20\%$ of tiles elsewhere, with landmarks
summarizing omitted tiles. Reported mean sparsity also accounts for landmarks,
protected tokens, and linear stages (Appendix~\ref{app:mean-sparsity}).

\subsection{Main Results}
\input{tables/main_combined_results}

\method provides configurable quality–speed trade-offs through the linear-stage fraction and sparse-stage retention budget. The conservative 0.5L/0.6S configuration prioritizes accuracy retention, whereas the more aggressive 0.7L/0.8S configuration delivers the largest measured speedups among our evaluated settings. 

\begin{figure}[!htbp]
    \centering
    \begin{minipage}[t]{0.29\textwidth}
        \centering
        \includegraphics[width=\linewidth]{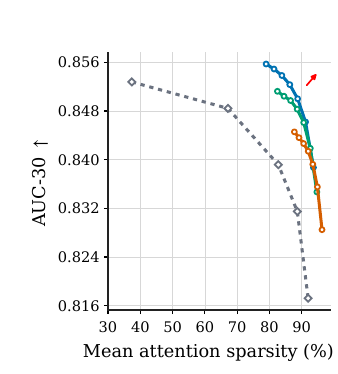}
        \smallskip
        \textbf{(a)} RE10K camera pose.
    \end{minipage}
    \hfill
    \begin{minipage}[t]{0.29\textwidth}
        \centering
        \includegraphics[width=\linewidth]{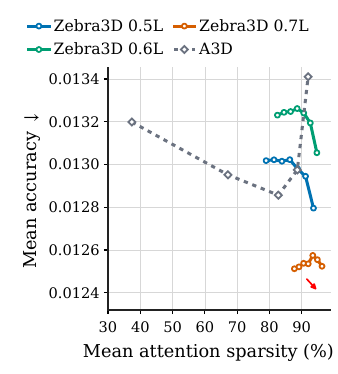}
        \smallskip
        \textbf{(b)} 7Scenes point maps.
    \end{minipage}
    \hfill
    \begin{minipage}[t]{0.29\textwidth}
        \centering
        \includegraphics[width=\linewidth]{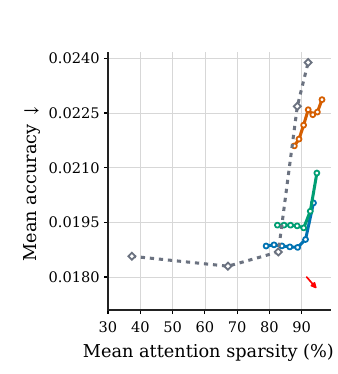}
        \smallskip
        \textbf{(c)} NRGBD point maps.
    \end{minipage}
    \caption{\textbf{HYWM quality--sparsity trade-off.}
    Zebra3D varies sparse retention within fixed placements; A3D varies
    subsampling.}
    \label{fig:hywm-performance-sparsity}
\end{figure}

\textbf{Training-free conversion.}
We evaluate \method across three
backbones on reconstruction and camera-pose benchmarks. Table~\ref{tab:main-combined-results} shows that substantial attention sparsity is achievable without gradient updates across all three backbones.
At 0.5L/0.6S, training-free conversion retains Re10K AUC-30 within
$0.007$, $0.004$, and $0.006$ of dense VGGT, HYWM, and $\pi^3$,
respectively, at approximately $79\%$ mean attention sparsity.
It also achieves the highest CO3Dv2 AUC-30 among the evaluated
non-dense methods for each backbone, while matching dense 7Scenes accuracy errors at the reported precision.
Increasing sparsity from 0.6S to 0.8S at the same linear-stage allocation raises sparsity to approximately 89\%, with additional Re10K losses of only 0.7, 0.6, and 0.5 percentage points. These results demonstrate that sparse retention is a useful control for adjusting efficiency without changing the stage allocation. Figure~\ref{fig:hywm-performance-sparsity} further examines this quality--sparsity trade-off against A3D on HYWM.

\input{tables/main_scannet50_results}

\begin{figure}[!t]
    \centering
    \includegraphics[width=0.92\textwidth]{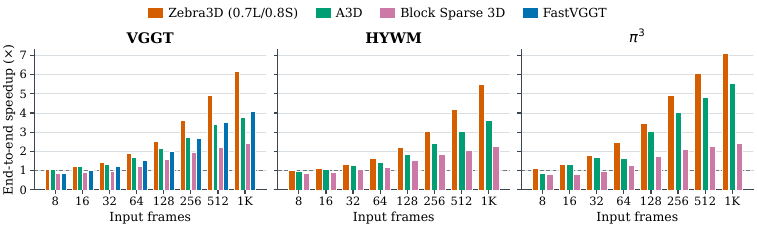}
    \caption{\textbf{End-to-end inference speedup relative to dense backbone.}
    At 1,024 frames, Zebra3D reaches
    $6.2{\times}$ on VGGT, $5.5{\times}$ on HYWM, and $7.1{\times}$ on
    $\pi^3$, with gains increasing as global attention becomes the dominant
    cost. Speedups measured on a single MI355X GPU.}
    \label{fig:inference-speedup}
\end{figure}

\textbf{Post-conversion training.}
Training benefits depend on both the backbone and the benchmark. For VGGT, the complete training objective raises Re10K AUC@30 from 0.737 to 0.805 at 0.5L/0.6S. The more aggressive 0.7L/0.8S configuration also reaches 0.805 at approximately 94\% mean attention sparsity, within 1.1 percentage points of dense VGGT fine-tuned on the same data and schedule. However, CO3Dv2 and NRGBD performance declines after training. We observe similar regressions in the dense control, indicating that the regressions are not specific to attention replacement. At 0.5L/0.6S, the refined hybrid remains within 0.002 CO3Dv2 AUC-30 of this control and matches its NRGBD accuracy error at the reported precision. This may reflect the adaptation to the training mixture may shift the model's performance toward some benchmarks at the expense of others. For $\pi^3$, training reduces the aggressive configuration's NRGBD accuracy error from 0.043 to 0.017, but does not improve its conservative configuration consistently. These results suggest that post-conversion training is particularly useful for recovering accuracy under aggressive replacement, whereas training-free conversion provides a strong conservative operating point.

\begin{figure}[!t]
    \centering
    \includegraphics[width=\linewidth]{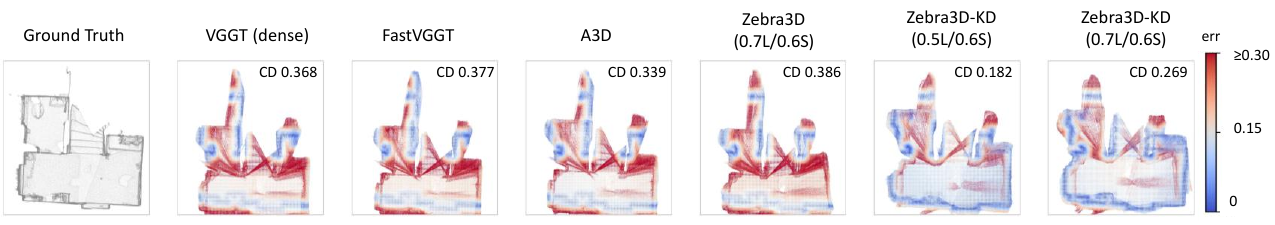}
    \caption{\textbf{Long-sequence reconstruction at $1{,}000$ views on
    ScanNet-50.} Visualization of the views of the predicted point maps, coloured by distance to the ground-truth mesh.}
    \label{fig:qual-scannet}
\end{figure}

\input{tables/attention_ablation_main}

\textbf{Long-sequence reconstruction.}
We further evaluate the quality and speed of \method on long-sequence reconstruction scenarios. At 1,000 views, \methodns-KD (0.7L/0.6S) achieves the lowest
reported Chamfer distance (CD) of $0.447$, compared with $0.466$
for dense VGGT, while achieving $3.54\times$ speedup. The more aggressive 0.7L/0.8S variant achieves a CD at 0.460 and a $5.6\times$ speedup, improving both quality and speed over dense VGGT, FastVGGT, and RegimeVGGT at this sequence length. Finetuned dense VGGT (VGGT + GTFT in Table~\ref{tab:scannet50-results}) increases CD to $0.503$ at 1,000 views, showing performance regression. A3D provides a competitive intermediate trade-off (0.452 Chamfer distance; $3.83{\times}$ speedup). This comparison exposes the role of sparse retention: retaining more context favors reconstruction quality, while more aggressive configuration favors runtime. The speedup of 0.7L/0.8S grows from $1.86\times$ at 100 views to $5.60\times$ at 1,000 views, making its efficiency advantage more pronounced for long sequences. Qualitative results are illustrated in Figure~\ref{fig:qual-scannet}, Figure~\ref{fig:qual-unseen} and in appendix~\ref{app:qualitative}.

\textbf{Efficiency across backbones.}
We also extend the runtime comparison to all three backbones. At 1,024 views, \method{} (0.7L/0.8S) achieves $6.2\times$, $5.5\times$, and $7.1\times$ end-to-end speedups 
over dense VGGT, HYWM, and $\pi^3$, respectively, and exceeds A3D and Block Sparse 3D (Figure~\ref{fig:inference-speedup}). The increasing gains with view count are consistent with global attention accounting for a larger fraction of runtime, while attention replacement has less impact when other components dominate.  

\textbf{Additional datasets and models.}
Across Tanks and Temples, ETH3D, and DTU at $N\in\{8,16,32,64\}$, all $0.5L$ configurations remain within $3\%$ of their dense backbones in NC1 and outperform A3D (Appendix~\ref{app:wide-baseline}). We also compare with Fast3R, MUSt3R, and Spann3R~\citep{yang2025fast3r,cabon2025must3r,wang2025spann3r} in Appendix~\ref{app:direct-multiview-comparison}. \method outperforms these models on point-map and camera-pose benchmarks and surpasses Fast3R and Spann3R in both ScanNet quality and speed. These results extend the conservative conversion setting beyond the video-sequence benchmarks.

\subsection{Ablation Studies}

\textbf{Layer-role placement.}
\label{sec:placement-ablation}
At a fixed $0.5$L/$0.8$S budget, sensitivity-guided allocation achieves the highest Re10K AUC@30 on all three backbones (Table~\ref{tab:attention-operator-ablations}(d)), while uniform,
random, and reverse-sensitivity assignments degrade substantially. The replacement budget alone therefore does not determine accuracy: preserving sparse softmax in the appropriate stages matters. Back-sparse nearly matches the guided allocation on $\pi^3$ ($0.864$ vs. $0.866$), consistent with its sensitivity being concentrated later in the network. This supports adapting the allocation to each backbone rather than imposing a shared stage pattern.

\textbf{Attention operators.}
Table~\ref{tab:attention-operator-ablations}(a) separates the effects of fitting and centering. On $\pi^3$, affine fitting raises Re10K AUC@30 from $0.789$ to $0.811$, and centering further improves it to $0.836$, indicating that both components help adapt the linear operator to pretrained representations (detailed comparisons are in Appendix~\ref{app:zero-shot-linear-ablation}). Table~\ref{tab:attention-operator-ablations}(b) shows that landmark summaries are particularly important under sparse selection: adding them to tile-sparse attention raises AUC@30 from $0.811$ to $0.868$ and reduces NRGBD accuracy error from $0.066$ to $0.014$. Thus, retaining compressed context from unselected tiles contributes substantially beyond exact attention over selected tokens. Appendix~\ref{app:additional-ablation-results} reports additional backbone and configuration results.

\textbf{Training objective.}
Table~\ref{tab:attention-operator-ablations}(c) compares training objectives at a fixed training budget. KD-only training underperforms GT-only training on Re10K ($0.716$ vs.\ $0.777$ AUC@30), and adding output KD alone does not close this gap (Table~3c). The complete objective performs best across all three reported metrics, reaching $0.805$ AUC@30 and reducing 7Scenes and NRGBD accuracy errors to $0.015$ and $0.024$. These results support combining task supervision with output and intermediate-feature supervision in the tested recipe. Full results are in (Appendix~\ref{app:kd-ablation}).

\textbf{Calibration-set stability.}
Across five calibration datasets, sensitivity rankings have Spearman correlations of $0.843--0.886$, and the resulting allocations agree with ours on $81.5--88.9\%$ of stages at $0.5L$ and $93.5--97.2\%$ at $0.7L$ (Appendix~\ref{app:calibration-dataset-stability}). This stability suggests that the allocation captures a component of backbone-specific sensitivity that transfers across the tested calibration sets, supporting the use of a fixed allocation without recalibrating for each evaluation dataset.

\section{Conclusion}

We introduced \method, a sensitivity-guided framework for upcycling pretrained 3D reconstruction transformers into sparse--linear hybrids. Across VGGT, HYWM, and $\pi^3$, we identified depth-dependent sensitivity to linear replacement, with more sensitive stages tending to exhibit more concentrated attention. This observation motivates assigning tile-sparse attention to sensitive stages and linear-centered attention to tolerant stages, while landmark summaries preserve unselected context. Closed-form affine fitting enables training-free conversion, with optional post-conversion training incorporating task supervision and output and relational feature distillation. \method offers quality--efficiency trade-offs at 79--94\% mean attention sparsity: conservative configurations preserve accuracy close to dense models, while more aggressive configurations prioritize speed. At 0.7L/0.8S, it delivers respective end-to-end speedups of $6.2\times$, $5.5\times$, and $7.1\times$ at $1{,}024$ views across the three backbones. Input-adaptive allocation and broader hardware support remain future work.

\subsection*{AI Use Statement}
Generative AI tools were used to assist with language editing and the
preparation and debugging of code and tables. AI is used to write implementation details in depth to aid in reproduction by coding agents. The authors reviewed and
verified all AI-assisted content and take full responsibility for the
paper's claims, results, and presentation.

\subsection*{Ethics Statement}
This work uses existing pretrained models and standard research datasets and
does not involve human-subject experiments or the collection of new personal
data.  Scalable 3D reconstruction can nevertheless enable privacy-sensitive
or surveillance applications.  Deployments should respect consent, dataset
licenses, privacy, and applicable legal requirements.

\subsection*{Reproducibility Statement}
The paper and appendix specify the model conversions, layer assignments,
calibration and training procedures, evaluation protocols, and ablations
needed to reproduce the reported results.  We will publicly release the code,
configurations, custom kernels, and trained checkpoints upon publication.

\bibliography{arxiv}
\bibliographystyle{arxiv}

\appendix
\section{Additional Background and Related Work}
\label{app:background-related-work}

\subsection{Inference in 3D Reconstruction Models}
Feed-forward 3D reconstruction models infer scene geometry directly from a set
of images. VGGT, HunyuanWorld-Mirror (HYWM), and $\pi^3$ differ in their tokens and
prediction heads, but follow the same broad computation
\citep{wang2025vggt,liu2026worldmirror,wang2026pi3}. A visual encoder, typically
DINOv2, converts each input view into patch embeddings, which are concatenated
with any model-specific tokens:
\begin{equation}
    \mT_i=E_{\theta}(\mI_i)\in\mathbb{R}^{P_i\times d},
    \qquad
    \mX^0=\operatorname{Concat}_{i=1}^{F}[\mT_i;\mS_i],
    \label{eq:inference-tokenization}
\end{equation}
where $\mS_i$ denotes camera, register, or geometric-conditioning tokens and
$N=\sum_i(P_i+|\mS_i|)$ is the sequence length. The resulting tokens pass
through an aggregator that alternates within-view processing and scene-level
mixing:
\begin{equation}
    \mX^{\ell+\frac12}
      =\operatorname{FrameAttn}_{\ell}(\mX^\ell),
    \qquad
    \mX^{\ell+1}
      =\operatorname{GlobalAttn}_{\ell}(\mX^{\ell+\frac12}),
    \label{eq:inference-aggregation}
\end{equation}
Task-specific heads then decode camera parameters and dense outputs such as
depth, point maps, normals, and tracking features. VGGT appends one camera
token and four register tokens to every view; $\pi^3$ removes
reference-specific embeddings to preserve permutation equivariance; and
WorldMirror can encode camera, intrinsic, and depth priors.

Global attention is the only stage in this pipeline that allows unrestricted
communication across all views. For head $h$ of width $d_h$,
\begin{equation}
    \mQ_h=\mX\mW_h^Q,\qquad
    \mK_h=\mX\mW_h^K,\qquad
    \mV_h=\mX\mW_h^V,
    \label{eq:global-qkv}
\end{equation}
and multi-head global attention is
\begin{equation}
    \mA_h=\softmax\left(\frac{\mQ_h\mK_h^\top}{\sqrt{d_h}}\right),
    \qquad
    \operatorname{GlobalAttn}(\mX)
    =\operatorname{Concat}_{h=1}^{H}(\mA_h\mV_h)\mW^O,
    \label{eq:global-attention}
\end{equation}
with a row-wise softmax. The dense matrix
$\mA_h\in\mathbb{R}^{N\times N}$ supports long-range correspondence and global
geometric consistency, but costs $\mathcal{O}(N^2d)$ computation. At fixed image resolution, compute
grows quadratically with the number of views. Prior work identifies this step
as the main obstacle to scaling feed-forward reconstruction
\citep{shen2026fastvggt,sun2026avggt,wang2026blocksparse,wang2026httm}.

\subsection{Extended Related Work}\label{app:extended_related_works}
\paragraph{Pairwise and direct multi-view reconstruction.}
DUSt3R regresses dense point maps from image pairs without requiring known
camera calibration or poses, but reconstructing an image collection requires
selecting pairs and aligning their local predictions in a common frame
\citep{wang2024dust3r}. The quadratic growth in candidate pairs and the
subsequent global optimization limit scaling to long sequences. MUSt3R extends
this paradigm to symmetric multi-view prediction in a shared coordinate frame
and introduces a multi-layer memory mechanism for scalable offline and online
reconstruction \citep{cabon2025must3r}. Fast3R also generalizes pairwise
reconstruction to unordered image sets: a fusion transformer processes all
views jointly and predicts local and global point maps in one forward pass,
avoiding external global alignment \citep{yang2025fast3r}. Its train-short,
test-long design and distributed inference enable sequences exceeding
$1{,}000$ images. These methods redesign the reconstruction architecture or
inference pathway for direct multi-view processing; \method instead upcycles
existing feed-forward 3D backbones by replacing their dense global-attention
operators.

\paragraph{Token reduction.}
Token merging reduces both dimensions of the attention matrix while leaving
softmax attention itself unchanged. FastVGGT merges redundant tokens while
protecting the reference view, salient regions, and spatial coverage
\citep{shen2026fastvggt}. Its safeguards reflect a central difficulty in 3D
reconstruction: tokens that appear redundant locally may still carry
cross-view evidence needed for correspondence or camera estimation. LiteVGGT
therefore selects geometry-aware anchors and caches merge assignments across
layers, reducing repeated selection overhead while maintaining spatial
structure \citep{shu2026litevggt}. HTTM performs temporal merging separately
for each attention head so that head-specific interaction patterns are not
forced into one shared token partition \citep{wang2026httm}. Co-Me instead
learns a confidence predictor and preferentially merges low-confidence tokens
\citep{chen2026come}. These methods reduce both query and context lengths, so
their savings propagate through subsequent layers, but merging is lossy and
changes the token set consumed by downstream prediction heads.

\paragraph{Key/value and context compression.}
Another strategy keeps every query but compresses the context. We refer to
AVGGT and its architecture-matched HYWM and $\pi^3$ variants collectively as
A3D. A3D replaces early global layers with frame attention, then spatially
subsamples patch keys and values while preserving queries and special tokens
\citep{sun2026avggt}. This preserves a prediction for every query token but
reduces the evidence available to each query. FlashVGGT summarizes each frame
with descriptor tokens and uses cross-attention from image tokens to those
descriptors; caching descriptors enables chunk-recursive inference without
reprocessing the complete history \citep{wang2026flashvggt}. RegimeVGGT
combines saliency-aware merging with protected key/value downsampling and
varies compression across depth, acknowledging that redundancy is not uniform
through the network \citep{you2026regimevggt}. Speed3R similarly combines a
compressed context with selected full-resolution interactions
\citep{ren2026speed3r}. Context-compression methods preserve the query grid,
but their sampling or summary policy determines which cross-view evidence
remains individually accessible.

\paragraph{Sparse and linear replacements.}
The closest methods replace global attention itself. We refer to Block-Sparse
VGGT and its architecture-matched variants as Block Sparse 3D. Block Sparse
3D uses sparse kernels to evaluate adaptively selected query--key blocks,
exploiting the concentration of attention around geometric correspondences
\citep{wang2026blocksparse}. Selected blocks retain exact softmax
interactions, while unselected blocks make no contribution. SAF3R introduces
dynamic sparse attention with head-specific sparse-pattern assignments
\citep{deng2026saf3r}. Lite3R instead combines sparse linear attention with
partial attention distillation, replacing quadratic interactions with a
model-agnostic efficient operator \citep{zhang2026lite3r}. VGG-T$^3$ takes a
different route: it replaces the variable-length scene representation with a
fixed-size MLP adapted at test time, yielding linear scaling in the number of
views \citep{elflein2026vggt3}.

\method differs along both operator selection and information preservation.
Rather than applying one approximation throughout the stack or prescribing a
fixed schedule, it profiles the pretrained model and assigns linear or
tile-sparse attention according to measured stage sensitivity. Tile-sparse
stages preserve exact softmax interactions for selected spatial regions and
represent every omitted region with a landmark under the same normalization.
Linear-centered stages retain the inherited projections and use a closed-form
fit, enabling training-free conversion. Optional relational distillation then
refines the same hybrid architecture without changing its stage assignment.

\input{tables/appendix_direct_multiview_comparison}

\input{tables/appendix_scannet_quality_speed}

\section{Implementation and Evaluation Details}
\label{app:implementation-details}

\input{tables/appendix_implementation_details}

\subsection{Mean Attention Sparsity}
\label{app:mean-sparsity}
Let $F$ be the number of frames, $N$ the total number of tokens, and
$C_\ell$ the number of query--key pairs evaluated by layer $\ell$.  We define
the mean attention sparsity over $L$ layers as
\begin{equation}
    \bar{s}=1-\frac{1}{L}\sum_{\ell=1}^{L}\frac{C_\ell}{N^2}.
\end{equation}
Thus, dense, frame, and linear attention have sparsities $0$,
$1-1/F$, and $1$, respectively.
Special-token terms may be omitted in a simplified patch-dominated
approximation when their contribution is negligible; all numerical examples
and performance--sparsity plots reported here retain them.

For \method, a sparse layer evaluates the keys in the selected spatial tiles,
one landmark for each unselected tile, all special-token keys, and an unselected query's
self key; special-token query rows remain dense.  If $\mathcal{Q}_{\rm sp}$
is the set of special-token queries and $\mathcal{K}_\ell(q)$ is the
resulting key set for query $q$, then
\begin{equation}
    C_\ell^{\rm Zebra}
    =|\mathcal{Q}_{\rm sp}|N+
      \sum_{q\notin\mathcal{Q}_{\rm sp}}|\mathcal{K}_\ell(q)|,
    \qquad
    \bar{s}_{\rm Zebra}
    =1-\frac{1}{L}\sum_{\ell=1}^{L}\frac{C_\ell^{\rm Zebra}}{N^2}.
\end{equation}

For A3D, let $E$ of its $D$ global stages use global-to-frame attention,
and let $V_\sigma$ be the number of keys retained by an SGA stage at
subsampling factor $\sigma$.  Here $V_\sigma$ includes all special-token
keys, the intact first frame when used, the subsampled patch grid, and the
mean-fill token.  Its mean sparsity is
\begin{equation}
    \bar{s}_{\rm A3D}
    =1-\frac{1}{D}\left(\frac{E}{F}
      +(D-E)\frac{V_\sigma}{N}\right).
\end{equation}

For RegimeVGGT, let $M_\ell$ be the number of query tokens remaining after
the layer-$\ell$ merge and $K_\ell$ the number of those tokens retained on
the key--value axis by its phase-shifted grid, including the reference-frame
anchor and all special tokens.  Merge-only middle layers have
$K_\ell=M_\ell$; unmerging occurs after attention and therefore adds no
query--key pairs.  Hence
\begin{equation}
    C_\ell^{\rm Regime}=M_\ell K_\ell,\qquad
    \bar{s}_{\rm Regime}
    =1-\frac{1}{D}\sum_{\ell=1}^{D}\frac{M_\ell K_\ell}{N^2}.
\end{equation}
The geometry-salient top-$\alpha$ protection set is input dependent, so its
overlap with the one-per-$2{\times}2$ merge-destination grid can change
$M_\ell$ and $K_\ell$ slightly across scenes.  Dataset results use the exact
counts accumulated during inference.  A configuration-only calculation uses
the expectation under uniformly located protected patches while retaining the
reference frame, ragged grid boundaries, special tokens, and all four released
layer groups.

\textbf{Example.}
For 300-frame VGGT at $518{\times}378$ resolution, each frame has 999 patch
and 5 special tokens, so $N=301{,}200$.  For \method 0.5L/0.8S ($20\%$
retention), a sparse layer attends 942 selected 64-token tile units, 3,741
landmarks, all 1,500 special tokens, and the self key.  Including dense
special-query rows gives $C_\ell/N^2=0.2215$.  Since 0.5L uses 12 sparse and
12 linear layers,
\begin{equation}
    \bar{s}_{\rm Zebra}
    =1-\frac{12(0.2215)+12(0)}{24}=0.8893.
\end{equation}
For A3D-f4, each non-reference frame retains
$\lceil37/2\rceil\lceil27/2\rceil=266$ patches.  Including the intact
reference frame, special tokens, and one mean-fill token gives
$V_4=82{,}034$.  With $E=9$ G2F stages among $D=24$ stages,
\begin{equation}
    \bar{s}_{\rm A3D}
    =1-\frac{9/300+15(82{,}034/301{,}200)}{24}=0.8285.
\end{equation}
For the canonical RegimeVGGT AB2 setting at $F=10$, $N=10{,}040$.
The expected $(M_\ell,K_\ell)$ counts are $(4046,2345.3)$ for the ten
shallow layers, $(5228.7,5228.7)$ for the four merge-only rank-peak layers,
$(5228.7,3139.5)$ for the remaining four middle layers, and
$(4046,2038.7)$ for the six deep layers.  Therefore
\begin{equation}
  \begin{aligned}
    \bar{s}_{\rm Regime}
    =1-\frac{1}{24(10{,}040)^2}\bigl[
      &10(4046)(2345.3)+4(5228.7)^2\\
      &+4(5228.7)(3139.5)+6(4046)(2038.7)\bigr]
    =0.8680.
  \end{aligned}
\end{equation}

\section{Layer Structure and Allocation Analysis}
\label{app:layer-structure}

\subsection{Depth-Wise Structure Across Backbones}
\label{app:cross-backbone-structure}
\begin{figure*}[t]
    \centering
    \begin{minipage}[t]{0.32\textwidth}
        \centering
        \includegraphics[width=\linewidth]{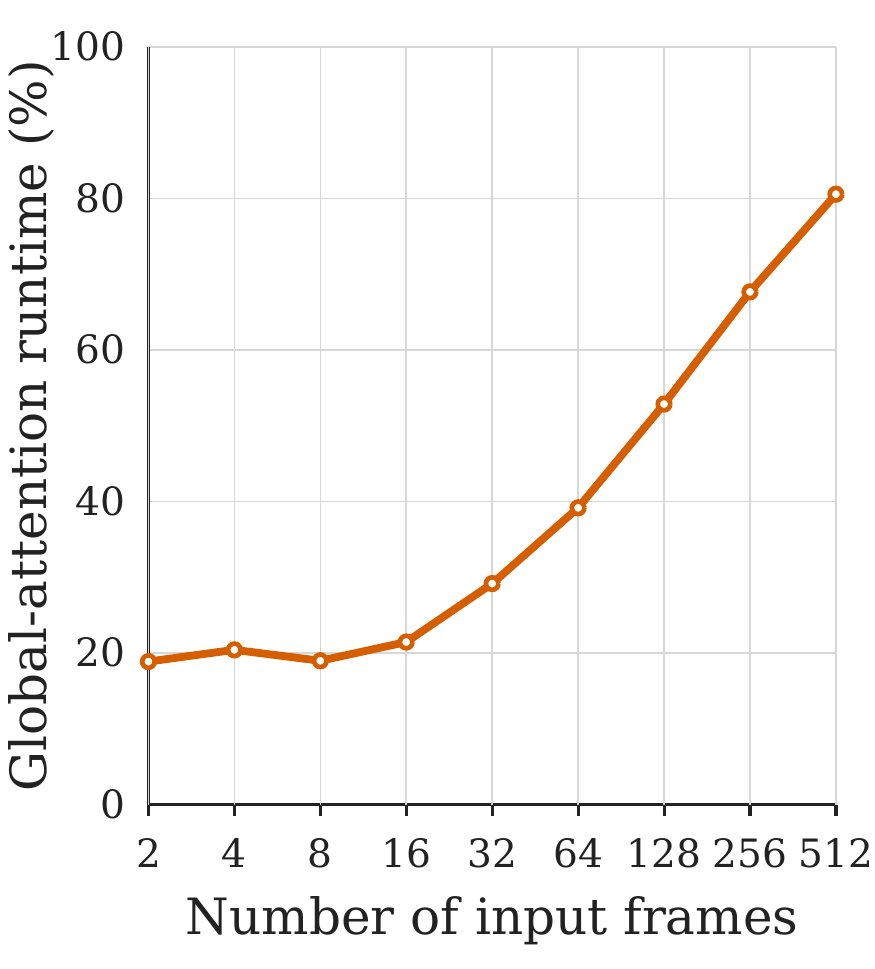}
        \smallskip
        \textbf{(a)} HYWM: Global-attention runtime share.
    \end{minipage}
    \hfill
    \begin{minipage}[t]{0.32\textwidth}
        \centering
        \includegraphics[width=\linewidth]{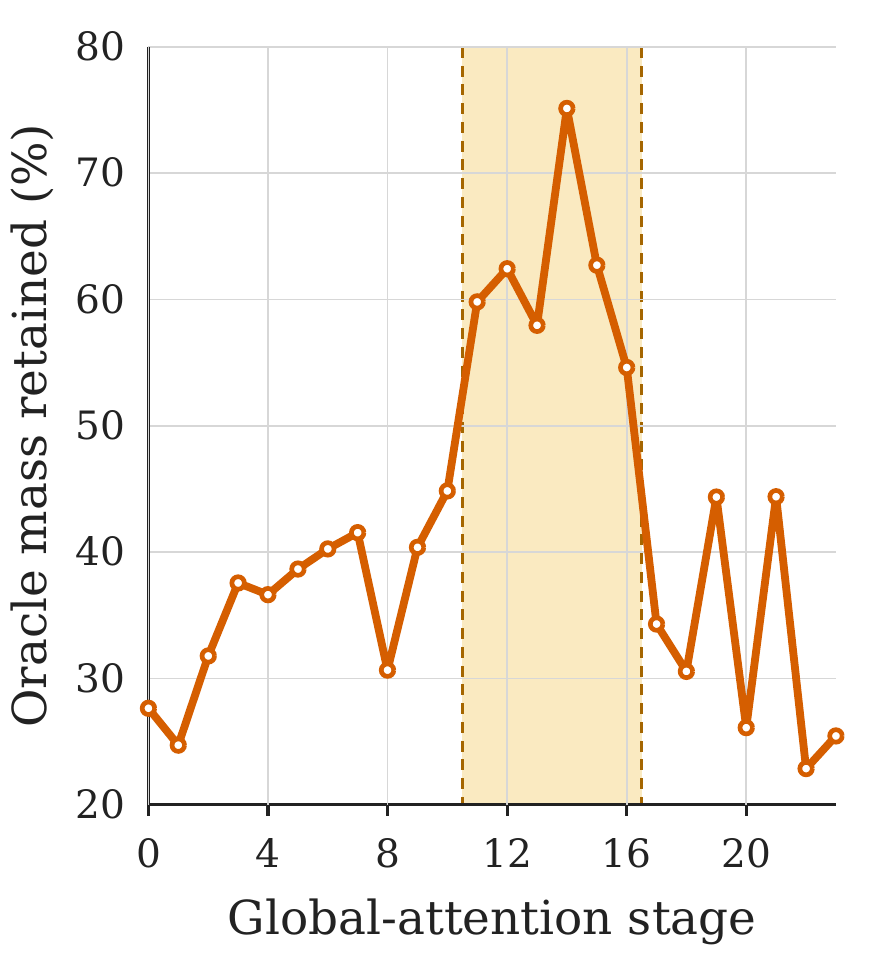}
        \smallskip
        \textbf{(b)} HYWM: Oracle top-$20\%$ block mass.
    \end{minipage}
    \hfill
    \begin{minipage}[t]{0.32\textwidth}
        \centering
        \includegraphics[width=\linewidth]{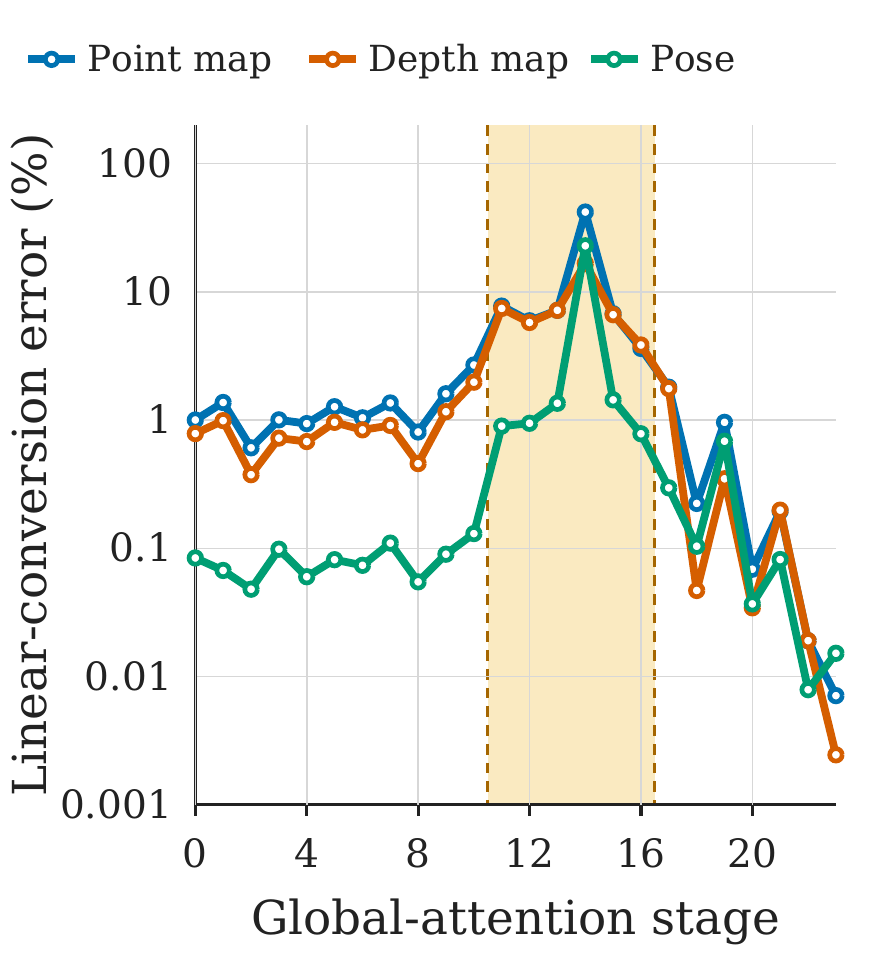}
        \smallskip
        \textbf{(c)} HYWM: Per-stage teacher deviation.
    \end{minipage}

    \medskip

    \begin{minipage}[t]{0.32\textwidth}
        \centering
        \includegraphics[width=\linewidth]{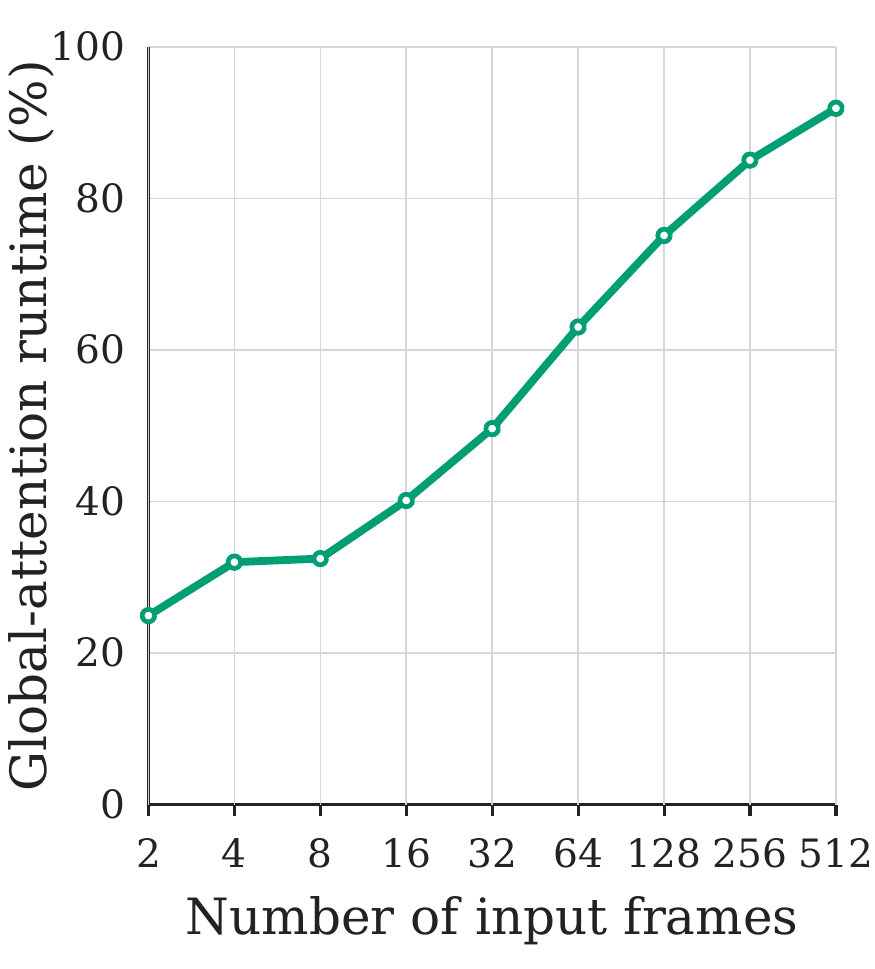}
        \smallskip
        \textbf{(d)} Pi3: Global-attention runtime share.
    \end{minipage}
    \hfill
    \begin{minipage}[t]{0.32\textwidth}
        \centering
        \includegraphics[width=\linewidth]{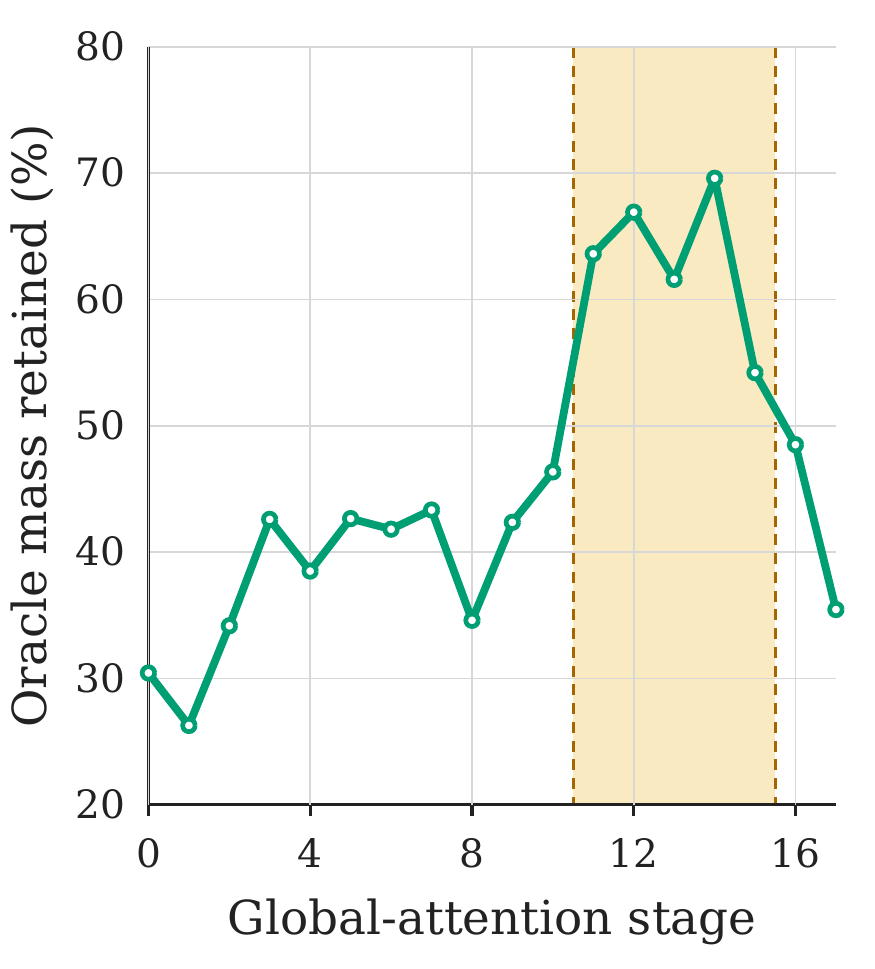}
        \smallskip
        \textbf{(e)} Pi3: Oracle top-$20\%$ block mass.
    \end{minipage}
    \hfill
    \begin{minipage}[t]{0.32\textwidth}
        \centering
        \includegraphics[width=\linewidth]{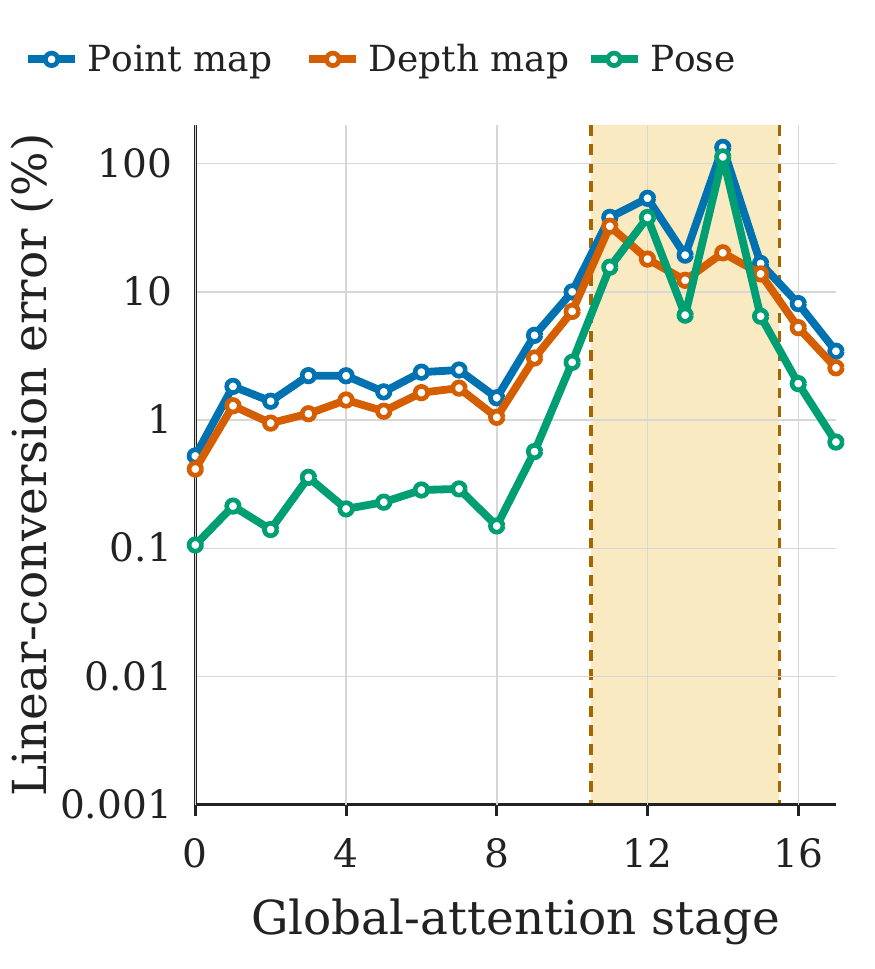}
        \smallskip
        \textbf{(f)} Pi3: Per-stage teacher deviation.
    \end{minipage}
    \caption{Additional background measurements for HYWM (top) and Pi3
    (bottom). Global-attention runtime reaches $80.6\%$ for HYWM and $91.9\%$
    for Pi3 at 512 frames. The middle column reports attention mass retained
    by oracle selection of the highest-mass $20\%$ of $64{\times}64$ token
    blocks. The right column reports relative final-output deviation after
    replacing one global-attention stage with fitted linear-centered
    attention, averaged over 11 held-out validation scenes. Shading marks the
    sensitive middle bands: stages 11--16 for HYWM and 11--15 for Pi3.}
    \label{fig:appendix-background-additional}
\end{figure*}

Figure~\ref{fig:appendix-background-additional} extends the VGGT measurements in
Figure~\ref{fig:background-efficiency} to HYWM and Pi3 using the same
measurement protocol.  Across all three backbones, global attention dominates
long-sequence inference, oracle block selection recovers only part of the
teacher attention mass at a $20\%$ block budget, and linear-centered conversion
is most disruptive in a narrow middle band of global-attention stages.
\ruijing{Figure~\ref{fig:appendix-attention-maps} shows the corresponding tile-level
teacher attention for eight VGGT stages across both linear bands and the
tile-sparse band.}

\begin{revision}
\begin{figure*}[t]
    \centering
    \includegraphics[width=\linewidth]{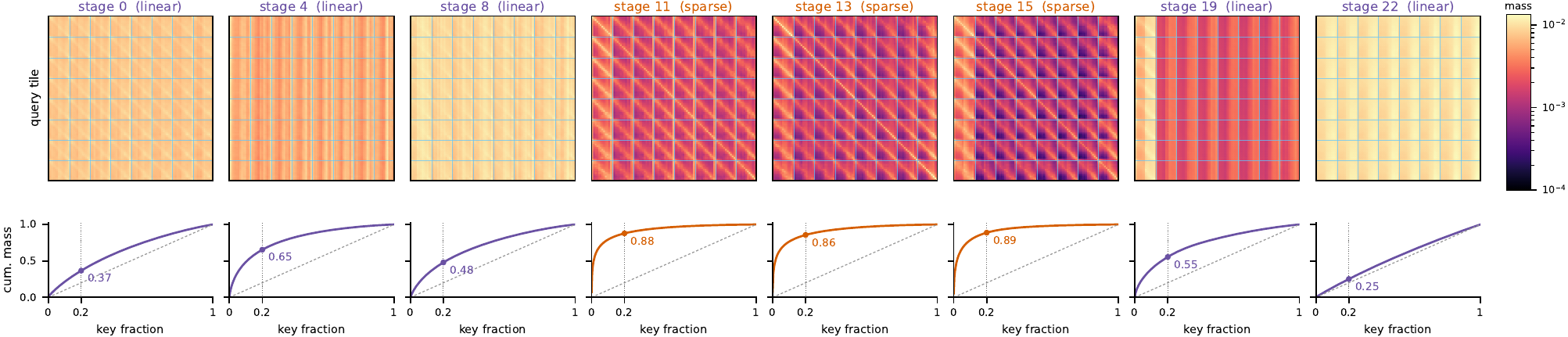}
    \caption{Tile-level teacher attention for eight VGGT global stages under
    the protocol of Figure~\ref{fig:attention-maps}, spanning both linear
    bands and the tile-sparse band.  The transition into concentrated,
    stripe-structured attention at stage $11$ and out of it after stage $15$
    matches the 0.7L assignment in
    Table~\ref{tab:layer-role-configs}.}
    \label{fig:appendix-attention-maps}
\end{figure*}
\end{revision}

\input{tables/appendix_placement_ablation}

\input{tables/appendix_calibration_stability}

\section{Operator and Refinement Ablations}
\label{app:additional-ablation-results}

\input{tables/attention_ablation_appendix}

\section{Extended Results and Wide-Baseline Evaluation}
\label{app:extended-results}

\subsection{Full Main-Benchmark Results}
\label{app:full-main-results}
The following tables report the complete reconstruction and camera-pose metric
breakdowns summarized by Table~\ref{tab:main-combined-results}:
Table~\ref{tab:main-reconstruction-results} reports reconstruction results,
and Table~\ref{tab:main-camera-pose-results} reports camera-pose results.

\input{tables/main_reconstruction_results}

\input{tables/main_camera_pose_results}

\subsection{Additional Performance--Sparsity Results}
\begin{figure*}[t]
    \centering
    \begin{minipage}[t]{0.32\textwidth}
        \centering
        \includegraphics[width=\linewidth]{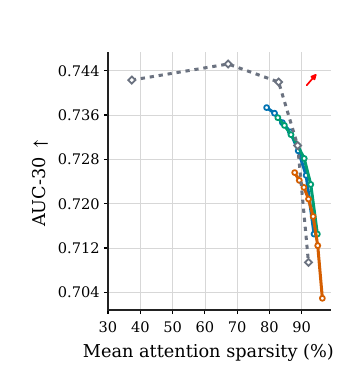}
        \smallskip
        \textbf{(a)} VGGT: RealEstate10K camera pose.
    \end{minipage}
    \hfill
    \begin{minipage}[t]{0.32\textwidth}
        \centering
        \includegraphics[width=\linewidth]{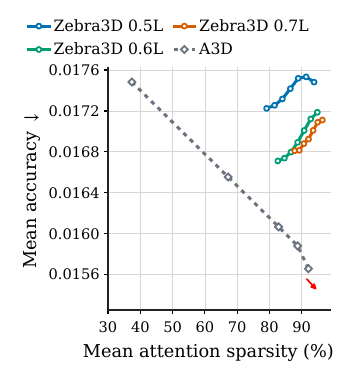}
        \smallskip
        \textbf{(b)} VGGT: 7Scenes point maps.
    \end{minipage}
    \hfill
    \begin{minipage}[t]{0.32\textwidth}
        \centering
        \includegraphics[width=\linewidth]{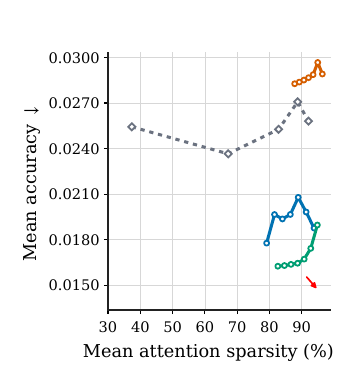}
        \smallskip
        \textbf{(c)} VGGT: Neural RGB-D point maps.
    \end{minipage}
    \medskip

    \begin{minipage}[t]{0.32\textwidth}
        \centering
        \includegraphics[width=\linewidth]{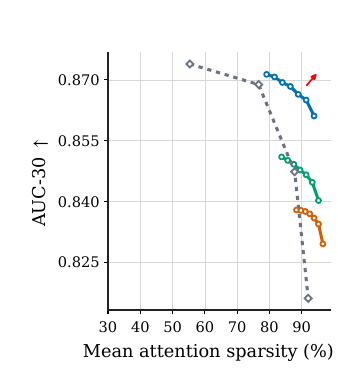}
        \smallskip
        \textbf{(d)} $\pi^3$: RealEstate10K camera pose.
    \end{minipage}
    \hfill
    \begin{minipage}[t]{0.32\textwidth}
        \centering
        \includegraphics[width=\linewidth]{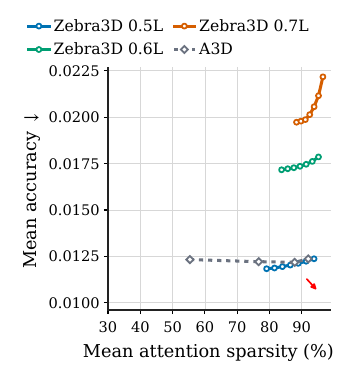}
        \smallskip
        \textbf{(e)} $\pi^3$: 7Scenes point maps.
    \end{minipage}
    \hfill
    \begin{minipage}[t]{0.32\textwidth}
        \centering
        \includegraphics[width=\linewidth]{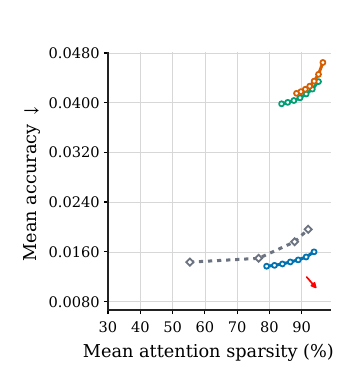}
        \smallskip
        \textbf{(f)} $\pi^3$: Neural RGB-D point maps.
    \end{minipage}
    \caption{Performance versus mean attention sparsity for VGGT and $\pi^3$ and mean attention sparsity is computed at 300 input frames. Zebra3D curves vary sparse retention within
    the 0.5L, 0.6L, and 0.7L placements; gray A3D curves vary the
    subsampling configuration. Higher is better for RealEstate10K AUC-30,
    while lower is better for 7Scenes and Neural RGB-D point-map accuracy
    error. The gold arrow points toward each panel's preferred corner.}
    \label{fig:appendix-performance-sparsity}
\end{figure*}

Figure~\ref{fig:appendix-performance-sparsity} reports the corresponding
300-frame trade-offs for VGGT and $\pi^3$. The sparsity calculation and plotting
protocol are identical across all three backbones.

\input{tables/appendix_wide_baseline_results}

\FloatBarrier
\section{Qualitative Results and Limitations}
\label{app:qualitative-limitations}

\subsection{Additional Qualitative Reconstructions}
\label{app:qualitative}
\begin{revision}
Figure~\ref{fig:appendix-qual-all} extends Figure~\ref{fig:qual-scannet} to
every scene used for qualitative inspection.  Scenes were selected from the
per-scene Chamfer distances already recorded for
Table~\ref{tab:scannet50-results}, before any render was produced, so the
choice is made on the reported metric rather than on appearance.

\begin{figure*}[!htbp]
    \centering
    \includegraphics[width=\linewidth]{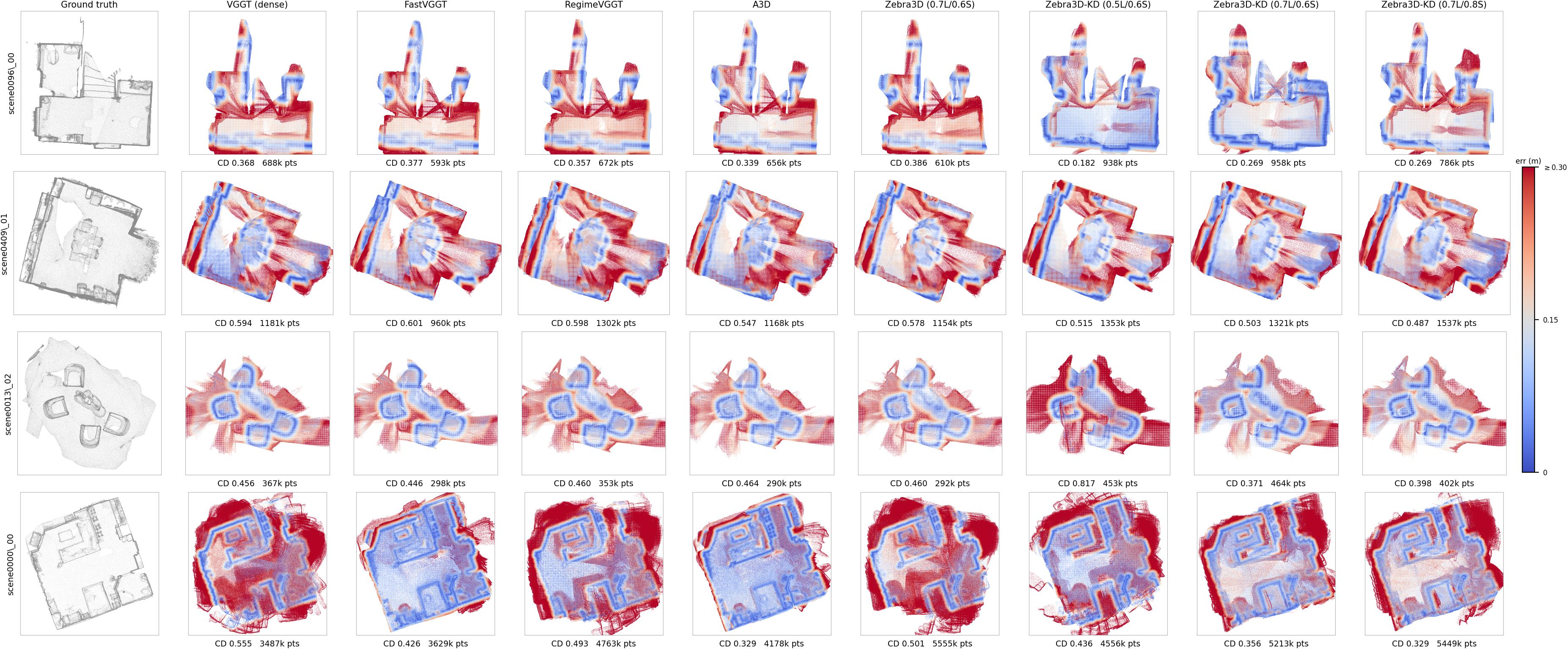}
    \caption{Long-sequence reconstruction at $1{,}000$ views on four
    ScanNet-50 scenes, coloured by distance to the ground-truth mesh
    (blue $0$\,m, red $\geq0.30$\,m).  Protocol is identical to
    Figure~\ref{fig:qual-scannet}.}
    \label{fig:appendix-qual-all}
\end{figure*}

\ruijing{The renders re-fit the evaluation's similarity transform on $1$--$99$
percentile extents, identically for every method, so that a few unprojection
outliers cannot shrink the room toward the centre of the frame; no points are
removed, and every reported Chamfer distance uses the evaluation's own
bounding-box alignment.}

A lower Chamfer distance can in principle be bought by smoothing: points spread
around the true surface score well under a two-sided mean distance, while the
render looks softer.  That is not what happens here.  Averaged over the
\ruijing{four scenes of Figure~\ref{fig:appendix-qual-all}}, \method-KD
($0.7$L/$0.6$S) improves the median per-point error from
\ruijing{$0.217$ to $0.174$}\,m, the $90$th percentile from
\ruijing{$0.536$ to $0.420$}\,m, and the share of points beyond $0.30$\,m from
\ruijing{$33.0\%$ to $23.1\%$}, so the gain is present at every quantile rather
than only in the mean.  Local surface thickness, estimated as the median
distance from a point to the best-fit plane of its $16$ nearest neighbours, is
\ruijing{$0.0114$\,m against $0.0112$\,m} for dense VGGT: the reconstructions
are no thicker.  What does differ is coverage, since
more of the predicted depth clears the shared confidence threshold, which is
why point counts are annotated in the figures.

Point clouds are exported from the evaluation path itself, and the Chamfer
distance recomputed from each exported cloud reproduces the recorded value to
$\pm0.0000$ for dense VGGT, A3D and both \method-KD configurations. FastVGGT
and RegimeVGGT differ by up to $0.005$, which is expected: their token-merging
decisions depend on runtime state and are not bit-reproducible across runs.  No
ordering in Table~\ref{tab:scannet50-results} changes.

\ruijing{Figure~\ref{fig:qual-unseen} adds four scenes from outside the training
data: an outdoor KITTI driving sequence and three Neural RGB-D rooms.  Unlike the ScanNet figures, the clouds are
coloured by the input images rather than by error, since this comparison is
qualitative only.}

\begin{figure*}[!htbp]
    \centering
    \setlength{\tabcolsep}{0.6pt}%
    \newcommand{\qualcell}[2]{\includegraphics[width=0.155\linewidth]{figures/qual_unseen/#1/#2.jpg}}%
    \newcommand{\qualrow}[3]{%
        \rotatebox{90}{\makebox[\dimexpr0.155\linewidth*#3/900\relax][c]{\scriptsize #1}} &
        \qualcell{#2}{input} & \qualcell{#2}{vggt} & \qualcell{#2}{zebra3d_07L08S} &
        \qualcell{#2}{zebra3d_05L06S} & \qualcell{#2}{zebra3d_kd_07L08S} &
        \qualcell{#2}{zebra3d_kd_05L06S}}%
    \begin{tabular}{@{}c@{\hspace{2pt}}cccccc@{}}
    & \scriptsize\shortstack{Input\\(first frame)}
    & \scriptsize\shortstack{VGGT\\(dense)}
    & \scriptsize\shortstack{Zebra3D\\(0.7L/0.8S)}
    & \scriptsize\shortstack{Zebra3D\\(0.5L/0.6S)}
    & \scriptsize\shortstack{Zebra3D-KD\\(0.7L/0.8S)}
    & \scriptsize\shortstack{Zebra3D-KD\\(0.5L/0.6S)} \\[1pt]
    \qualrow{KITTI}{kitti_0023}{483} \\[2pt]
    \qualrow{\shortstack{NRGBD\\kitchen}}{nrgbd_complete_kitchen}{675} \\[2pt]
    \qualrow{\shortstack{NRGBD\\apartment}}{nrgbd_morning_apartment}{616} \\[2pt]
    \qualrow{\shortstack{NRGBD\\whiteroom}}{nrgbd_whiteroom}{672}
    \end{tabular}
    \caption{\ruijing{\textbf{Reconstructions on scenes outside the training data.}
    From top: $40$ consecutive frames of the KITTI driving sequence
    \texttt{2011\_09\_26\_drive\_0023} (camera~02); $40$ frames each of the
    Neural RGB-D scenes \texttt{complete\_kitchen}, \texttt{morning\_apartment}
    and \texttt{whiteroom}.  None of these is in the training data of
    any model shown; VGGT's includes only the synthetic Virtual KITTI.  The first
    column is the first input frame, and every method is rendered from one
    virtual camera placed behind and above it.  The same lowest-confidence $40\%$
    and farthest $3\%$ of pixels are removed from every method, so each keeps
    nearly the same number of points, and each point is drawn as a disc sized to
    the surface its source pixel covers.}}
    \label{fig:qual-unseen}
\end{figure*}

\end{revision}

\FloatBarrier

\subsection{Limitations}
\label{app:limitations}
\method primarily benefits long sequences, where global attention dominates inference cost; speedups remain modest at 8--16 views (Figure~\ref{fig:inference-speedup}). As attention becomes cheaper, the image encoder and prediction heads impose an increasing upper bound on end-to-end acceleration. Our stage allocation is fixed per backbone and derived from marginal, single-stage conversion sensitivity, so it does not model interactions among converted stages or adapt to individual inputs. Although the allocation is stable across our calibration datasets, broader generalization to substantially different architectures and domains remains to be established. Finally, our optimized kernels and benchmarks target Triton on AMD MI355X GPUs; portability and performance on other accelerators require further evaluation. Optional distillation also requires a frozen dense teacher during refinement, although it adds no inference-time cost.

\end{document}

%% file: math_commands.tex
\usepackage{amsmath,amsfonts,bm}

\def\eqref#1{equation~\ref{#1}}

\def\1{\bm{1}}

\def\mA{{\bm{A}}}

\def\mI{{\bm{I}}}

\def\mK{{\bm{K}}}

\def\mQ{{\bm{Q}}}

\def\mS{{\bm{S}}}
\def\mT{{\bm{T}}}

\def\mV{{\bm{V}}}
\def\mW{{\bm{W}}}
\def\mX{{\bm{X}}}

\DeclareMathAlphabet{\mathsfit}{\encodingdefault}{\sfdefault}{m}{sl}
\SetMathAlphabet{\mathsfit}{bold}{\encodingdefault}{\sfdefault}{bx}{n}

\newcommand{\softmax}{\mathrm{softmax}}

%% file: tables/main_combined_results.tex
\begin{table*}[t]
\caption{\textbf{Reconstruction and pose.} Point-map results on 7Scenes/NRGBD
(stride 10) and pose on RE10K/CO3Dv2 (10 views). $\bar{s}_N$ denotes mean
attention sparsity at $N$ views. Bold/underline: best/second-best non-dense.
VGGT + GT-FT is dense VGGT finetuned on the same data as Zebra3D-KD. Full
metrics: Appendix~\ref{app:full-main-results}.}
\label{tab:main-combined-results}
\centering
\small
\setlength{\tabcolsep}{3.0pt}
\renewcommand{\arraystretch}{0.98}
\resizebox{\textwidth}{!}{%
\begin{tabular}{@{}l cc cc cc cc cc@{}}
\toprule
\textbf{Method}
& \multicolumn{2}{c}{\textbf{Mean sparsity (\%)} $\uparrow$}
& \multicolumn{2}{c}{\textbf{7Scenes} $\downarrow$}
& \multicolumn{2}{c}{\textbf{NRGBD} $\downarrow$}
& \multicolumn{2}{c}{\textbf{RE10K} $\uparrow$}
& \multicolumn{2}{c}{\textbf{CO3Dv2} $\uparrow$} \\
\cmidrule(lr){2-3}\cmidrule(lr){4-5}\cmidrule(lr){6-7}
\cmidrule(lr){8-9}\cmidrule(lr){10-11}
& $\bar{s}_{300}$ & $\bar{s}_{10}$
& Acc. mean & Comp. mean
& Acc. mean & Comp. mean
& AUC-30 & RTA-5
& AUC-3 & AUC-30 \\
\midrule
\rowcolor{modelgray}
VGGT & 0 & 0 & 0.017 & 0.027 & 0.016 & 0.011 & 0.744 & 0.543 & 0.566 & 0.900 \\
\rowcolor{modelgray}
\ruijing{VGGT + GT-FT} & \ruijing{0} & \ruijing{0} & \ruijing{0.024} & \ruijing{0.031} & \ruijing{0.024} & \ruijing{0.018} & \ruijing{0.816} & \ruijing{0.674} & \ruijing{0.474} & \ruijing{0.885} \\
FastVGGT & 88 & 83 & \textbf{0.015} & 0.025 & \underline{0.020} & \textbf{0.013} & 0.708 & 0.476 & 0.412 & 0.867 \\
SAF3R & 84 & 78 & 0.018 & \textbf{0.022} & 0.054 & 0.022 & 0.714 & 0.493 & 0.504 & 0.883 \\
RegimeVGGT & \underline{90} & 87 & \underline{0.016} & 0.026 & 0.024 & 0.016 & 0.722 & 0.498 & 0.445 & 0.876 \\
Block Sparse 3D & 70 & 70 & 0.017 & 0.025 & 0.042 & 0.016 & 0.649 & 0.379 & 0.207 & 0.835 \\
Lite3R & 70 & 70 & 0.039 & \underline{0.023} & 0.068 & 0.018 & 0.578 & 0.253 & 0.078 & 0.779 \\
A3D & 83 & 75 & \underline{0.016} & 0.026 & 0.025 & 0.017 & \underline{0.742} & 0.536 & 0.468 & 0.881 \\
\rowcolor{zebralavender}
\textbf{Zebra3D (0.5L/0.8S)} & 89 & \underline{89} & 0.018 & 0.027 & 0.021 & \underline{0.014} & 0.730 & 0.507 & \underline{0.535} & \underline{0.892} \\
\rowcolor{zebralavender}
\textbf{Zebra3D (0.5L/0.6S)} & 79 & 79 & 0.017 & 0.027 & \textbf{0.018} & \textbf{0.013} & 0.737 & 0.526 & \textbf{0.554} & \textbf{0.897} \\
\rowcolor{zebralavender}
\textbf{Zebra3D-KD (0.5L/0.6S)} & 79 & 79 & \textbf{0.015} & 0.026 & 0.024 & 0.018 & \textbf{0.805} & \underline{0.649} & 0.458 & 0.883 \\
\rowcolor{zebralavender}
\textbf{Zebra3D-KD (0.7L/0.8S)} & \textbf{94} & \textbf{94} & \underline{0.016} & 0.025 & 0.025 & 0.019 & \textbf{0.805} & \textbf{0.654} & 0.403 & 0.872 \\
\midrule
\rowcolor{modelgray}
HYWM & 0 & 0 & 0.013 & 0.024 & 0.018 & 0.014 & 0.860 & 0.774 & 0.403 & 0.865 \\
Block Sparse 3D & 70 & 70 & \underline{0.014} & 0.024 & 0.041 & 0.023 & 0.735 & 0.573 & 0.085 & 0.761 \\
A3D & 83 & 75 & \textbf{0.013} & 0.024 & \underline{0.019} & \underline{0.014} & 0.839 & 0.740 & 0.323 & 0.844 \\
RegimeVGGT & \underline{90} & 87 & \textbf{0.013} & 0.024 & \underline{0.019} & \underline{0.014} & 0.840 & 0.741 & 0.315 & 0.843 \\
\rowcolor{zebralavender}
\textbf{Zebra3D (0.5L/0.8S)} & 89 & \underline{89} & \textbf{0.013} & 0.024 & \underline{0.019} & \textbf{0.013} & 0.850 & 0.758 & \underline{0.376} & \underline{0.854} \\
\rowcolor{zebralavender}
\textbf{Zebra3D (0.5L/0.6S)} & 79 & 79 & \textbf{0.013} & 0.025 & \underline{0.019} & \underline{0.014} & \textbf{0.856} & \textbf{0.769} & \textbf{0.400} & \textbf{0.862} \\
\rowcolor{zebralavender}
\textbf{Zebra3D-KD (0.5L/0.6S)} & 79 & 79 & \textbf{0.013} & \underline{0.023} & \textbf{0.018} & \underline{0.014} & \underline{0.854} & \underline{0.766} & 0.332 & 0.849 \\
\rowcolor{zebralavender}
\textbf{Zebra3D-KD (0.7L/0.8S)} & \textbf{94} & \textbf{94} & \underline{0.014} & \textbf{0.022} & \underline{0.019} & \underline{0.014} & 0.847 & 0.752 & 0.295 & 0.838 \\
\midrule
\rowcolor{modelgray}
$\pi^3$ & 0 & 0 & 0.012 & 0.017 & 0.014 & 0.009 & 0.877 & 0.809 & 0.582 & 0.903 \\
Block Sparse 3D & 70 & 70 & 0.020 & \textbf{0.016} & 0.049 & 0.017 & 0.786 & 0.655 & 0.213 & 0.835 \\
A3D & 88 & 82 & \textbf{0.012} & \textbf{0.016} & 0.018 & \underline{0.010} & 0.847 & 0.765 & 0.457 & 0.880 \\
\rowcolor{zebralavender}
\textbf{Zebra3D (0.5L/0.8S)} & \underline{89} & \underline{89} & \textbf{0.012} & \underline{0.018} & \underline{0.015} & \underline{0.010} & \underline{0.866} & \underline{0.792} & \underline{0.542} & \underline{0.895} \\
\rowcolor{zebralavender}
\textbf{Zebra3D (0.5L/0.6S)} & 79 & 79 & \textbf{0.012} & \underline{0.018} & \textbf{0.014} & \textbf{0.009} & \textbf{0.871} & \textbf{0.800} & \textbf{0.563} & \textbf{0.900} \\
\rowcolor{zebralavender}
\textbf{Zebra3D (0.7L/0.8S)} & \textbf{94} & \textbf{94} & 0.021 & 0.020 & 0.043 & 0.015 & 0.836 & 0.726 & 0.256 & 0.841 \\
\rowcolor{zebralavender}
\textbf{Zebra3D-KD (0.5L/0.6S)} & 79 & 79 & \textbf{0.012} & \underline{0.018} & \underline{0.015} & \textbf{0.009} & 0.855 & 0.773 & 0.527 & \underline{0.895} \\
\rowcolor{zebralavender}
\ruijing{\textbf{Zebra3D-KD (0.7L/0.8S)}} & \ruijing{\textbf{94}} & \ruijing{\textbf{94}} & \ruijing{\underline{0.013}} & \ruijing{\underline{0.018}} & \ruijing{0.017} & \ruijing{\underline{0.010}} & \ruijing{0.844} & \ruijing{0.751} & \ruijing{0.440} & \ruijing{0.876} \\
\bottomrule
\end{tabular}%
}
\end{table*}

%% file: tables/main_scannet50_results.tex
\begin{table}[!t]
\caption{\textbf{ScanNet-50 long-sequence reconstruction.} Chamfer distance
(CD; $\downarrow$) and end-to-end speedup. Bold/underline:
best/second-best non-dense. VGGT + GT-FT is dense VGGT finetuned on
the same data as Zebra3D-KD.
See Appendix~\ref{app:scannet50-protocol} for the detailed evaluation protocol.}
\label{tab:scannet50-results}
\centering
\small
\setlength{\tabcolsep}{3.5pt}
\renewcommand{\arraystretch}{0.96}
\resizebox{\textwidth}{!}{%
\begin{tabular}{@{}l cc cc cc cc@{}}
\toprule
\textbf{Method}
& \multicolumn{2}{c}{\textbf{1,000 frames}}
& \multicolumn{2}{c}{\textbf{500 frames}}
& \multicolumn{2}{c}{\textbf{300 frames}}
& \multicolumn{2}{c}{\textbf{100 frames}} \\
\cmidrule(lr){2-3}\cmidrule(lr){4-5}\cmidrule(lr){6-7}\cmidrule(lr){8-9}
& \textbf{CD} $\downarrow$ & \textbf{Speedup} $\uparrow$
& \textbf{CD} $\downarrow$ & \textbf{Speedup} $\uparrow$
& \textbf{CD} $\downarrow$ & \textbf{Speedup} $\uparrow$
& \textbf{CD} $\downarrow$ & \textbf{Speedup} $\uparrow$ \\
\midrule
\rowcolor{modelgray}
VGGT & 0.466 & 1.00 & 0.465 & 1.00 & 0.451 & 1.00 & 0.440 & 1.00 \\
\rowcolor{modelgray}
VGGT + GT-FT & 0.503 & 1.00 & 0.496 & 1.00 & 0.486 & 1.00 & 0.474 & 1.00 \\
FastVGGT & 0.464 & \underline{3.97} & 0.462 & \underline{3.40} & 0.450 & \underline{2.89} & 0.435 & 1.81 \\
RegimeVGGT & 0.478 & 3.50 & 0.468 & 3.03 & 0.453 & 2.59 & 0.439 & 1.70 \\
A3D & 0.452 & 3.83 & 0.445 & 3.38 & 0.438 & \textbf{2.95} & \underline{0.427} & \textbf{2.00} \\
Block Sparse 3D & 0.471 & 2.33 & 0.464 & 2.11 & 0.465 & 1.90 & 0.446 & 1.42 \\
\rowcolor{zebralavender}
\textbf{Zebra3D-KD (0.5L/0.8S)} & \underline{0.449} & 3.78 & \textbf{0.443} & 3.34 & 0.438 & 1.74 & \underline{0.427} & 1.50 \\
\rowcolor{zebralavender}
\textbf{Zebra3D-KD (0.5L/0.6S)} & 0.451 & 2.30 & \underline{0.444} & 2.20 & \underline{0.435} & 1.38 & 0.430 & 1.30 \\
\rowcolor{zebralavender}
\textbf{Zebra3D-KD (0.7L/0.8S)} & 0.460 & \textbf{5.60} & 0.449 & \textbf{4.59} & 0.447 & 2.51 & 0.439 & \underline{1.86} \\
\rowcolor{zebralavender}
\textbf{Zebra3D-KD (0.7L/0.6S)} & \textbf{0.447} & 3.54 & 0.447 & 3.21 & \textbf{0.429} & 2.02 & \textbf{0.424} & 1.67 \\
\bottomrule
\end{tabular}%
}
\end{table}

%% file: tables/attention_ablation_main.tex
\begin{table}[!t]
\caption{\textbf{Ablations.}
\textbf{(a)} Linear-attention initialization;
\textbf{(b)} sparse attention with landmark summaries;
\textbf{(c)} KD objectives for VGGT;
\textbf{(d)} layer placement at fixed budget.
Additional ablations are reported in Appendix~\ref{app:additional-ablation-results}.}
\label{tab:attention-operator-ablations}
\centering
\scriptsize
\setlength{\tabcolsep}{3.5pt}
\renewcommand{\arraystretch}{0.96}
\begin{minipage}[t]{0.495\textwidth}
\centering
\textbf{(a) Linear attention (0.7L/0.8S) on $\pi^3$}
\smallskip
\resizebox{\linewidth}{!}{%
\begin{tabular}{@{}lccc@{}}
\toprule
\textbf{\shortstack{Attention choice}} & \textbf{\shortstack{RE10K\\AUC-30 $\uparrow$}} & \textbf{\shortstack{7S Acc. $\downarrow$}} & \textbf{\shortstack{NRGBD\\Acc. $\downarrow$}} \\
\midrule
Frame attention & 0.655 & 0.031 & 0.051 \\
Linear, no fit & 0.789 & 0.028 & 0.071 \\
Linear, uncentered fit & 0.811 & 0.024 & 0.065 \\
\rowcolor{zebralavender}
Linear-centered + fit & \textbf{0.836} & \textbf{0.021} & \textbf{0.043} \\
\bottomrule
\end{tabular}%
}
\end{minipage}\hfill
\begin{minipage}[t]{0.495\textwidth}
\centering
\textbf{(b) Sparse attention (0.5L/0.75S) on $\pi^3$}
\smallskip
\resizebox{\linewidth}{!}{%
\begin{tabular}{@{}lccc@{}}
\toprule
\textbf{\shortstack{Attention choice}} & \textbf{\shortstack{RE10K\\AUC-30 $\uparrow$}} & \textbf{\shortstack{7S Acc. $\downarrow$}} & \textbf{\shortstack{NRGBD\\Acc. $\downarrow$}} \\
\midrule
SGA & 0.847 & \textbf{0.011} & 0.017 \\
Block sparse & 0.808 & 0.024 & 0.058 \\
Tile sparse & 0.811 & 0.027 & 0.066 \\
Block sparse + landmark & 0.864 & 0.013 & 0.024 \\
\rowcolor{zebralavender}
Tile sparse + landmark & \textbf{0.868} & 0.012 & \textbf{0.014} \\
\bottomrule
\end{tabular}%
}
\end{minipage}
\par\medskip
\begin{minipage}[t]{0.495\textwidth}
\centering
\textbf{(c) Knowledge distillation (VGGT at 0.5L/0.6S)}
\smallskip
\resizebox{\linewidth}{!}{%
\begin{tabular}{@{}lccc@{}}
\toprule
\textbf{\shortstack{Training objective}}
& \textbf{\shortstack{7S Acc. $\downarrow$}}
& \textbf{\shortstack{NRGBD\\Acc. $\downarrow$}}
& \textbf{\shortstack{RE10K\\AUC-30 $\uparrow$}} \\
\midrule
GT only & 0.023 & 0.046 & 0.777 \\
KD only & 0.023 & 0.045 & 0.716 \\
GT + output KD & 0.023 & 0.041 & 0.736 \\
GT + feature KD & 0.029 & 0.047 & 0.756 \\
\rowcolor{zebralavender}
RelKD & \textbf{0.015} & \textbf{0.024} & \textbf{0.805} \\
\bottomrule
\end{tabular}%
}
\end{minipage}\hfill
\begin{minipage}[t]{0.495\textwidth}
\centering
\textbf{(d) Layer placement at fixed 0.5L/0.8S budget}
\smallskip
\resizebox{\linewidth}{!}{%
\begin{tabular}{@{}lccc@{}}
\toprule
\textbf{Placement} & \textbf{VGGT} & \textbf{HYWM} & \textbf{$\pi^3$} \\
\midrule
Back-sparse & 0.590 & 0.609 & 0.864 \\
Front-sparse & 0.208 & 0.200 & 0.033 \\
Uniform & 0.115 & 0.142 & 0.148 \\
Reverse-sensitivity & 0.114 & 0.137 & 0.033 \\
Random, mean of 3 seeds & 0.318 & 0.216 & 0.063 \\
\rowcolor{zebralavender}
Sensitivity-guided & \textbf{0.730} & \textbf{0.850} & \textbf{0.866} \\
\bottomrule
\end{tabular}%
}
\end{minipage}
\end{table}

%% file: tables/appendix_direct_multiview_comparison.tex
\subsection{Comparison with Direct Multi-View Reconstruction Models}
\label{app:direct-multiview-comparison}
Tables~\ref{tab:direct-multiview-scannet50}
and~\ref{tab:direct-multiview-main-benchmarks} compare \method with three
architectures designed for direct multi-view reconstruction. MUSt3R predicts
views in a shared coordinate frame using a multi-layer memory
\citep{cabon2025must3r}; Spann3R incrementally maintains an external spatial
memory~\citep{wang2025spann3r}; and Fast3R jointly processes an unordered
image set in one forward pass~\citep{yang2025fast3r}.

MUSt3R provides the highest long-sequence speedup but substantially higher
reconstruction error. At 1,000 frames, \methodns-KD 0.7L/0.8S reaches
$5.60\times$ speedup with CD $0.460$, while the 0.7L/0.6S configuration
achieves the best CD ($0.447$) at $3.54\times$. Fast3R and Spann3R improve
accuracy over MUSt3R but remain less accurate than dense VGGT and both
\methodns-KD operating points.

Table~\ref{tab:direct-multiview-main-benchmarks} extends the comparison to the
main reconstruction and pose benchmarks. Every \method 0.5L/0.6S backbone
outperforms the three direct multi-view models on all reported 7Scenes and
NRGBD point-map metrics and on all CO3Dv2 AUC thresholds. On RE10K,
\methodns-HYWM and \methodns-Pi3 also exceed MUSt3R in AUC-30 and RTA-30, while
MUSt3R remains competitive in RRA-30. Zebra3D-$\pi^3$ gives the strongest overall
pose results, reaching $0.871$ RE10K AUC-30 and $0.900$ CO3Dv2 AUC-30.

\begin{table}[t]
\caption{\textbf{Direct multi-view reconstruction on ScanNet-50.} Chamfer
distance (CD; $\downarrow$) and end-to-end speedup relative to dense VGGT on
one MI355X. Bold/underline denote the best/second-best non-dense results.
Spann3R\textsuperscript{*} was evaluated without \texttt{torch.compile}.}
\label{tab:direct-multiview-scannet50}
\centering
\small
\setlength{\tabcolsep}{3.5pt}
\renewcommand{\arraystretch}{0.96}
\resizebox{\textwidth}{!}{%
\begin{tabular}{@{}l cc cc cc cc@{}}
\toprule
\textbf{Method}
& \multicolumn{2}{c}{\textbf{1,000 frames}}
& \multicolumn{2}{c}{\textbf{500 frames}}
& \multicolumn{2}{c}{\textbf{300 frames}}
& \multicolumn{2}{c}{\textbf{100 frames}} \\
\cmidrule(lr){2-3}\cmidrule(lr){4-5}\cmidrule(lr){6-7}\cmidrule(lr){8-9}
& \textbf{CD} $\downarrow$ & \textbf{Speedup} $\uparrow$
& \textbf{CD} $\downarrow$ & \textbf{Speedup} $\uparrow$
& \textbf{CD} $\downarrow$ & \textbf{Speedup} $\uparrow$
& \textbf{CD} $\downarrow$ & \textbf{Speedup} $\uparrow$ \\
\midrule
\rowcolor{modelgray}
VGGT & 0.466 & 1.00 & 0.465 & 1.00 & 0.451 & 1.00 & 0.440 & 1.00 \\
MUSt3R & 0.958 & \textbf{15.7} & 0.937 & \textbf{7.07} & 0.922 & \textbf{3.85} & 0.897 & 0.91 \\
Spann3R\textsuperscript{*} & 0.605 & 0.50 & 0.604 & 0.27 & 0.588 & 0.18 & 0.567 & 0.09 \\
Fast3R & 0.588 & 1.75 & 0.558 & 1.64 & 0.547 & 1.52 & 0.524 & \underline{1.38} \\
\rowcolor{zebralavender}
\textbf{Zebra3D-KD (0.7L/0.8S)} & \underline{0.460} & \underline{5.60} & \underline{0.449} & \underline{4.59} & \underline{0.447} & \underline{2.51} & \underline{0.439} & \textbf{1.86} \\
\rowcolor{zebralavender}
\textbf{Zebra3D-KD (0.7L/0.6S)} & \textbf{0.447} & 3.54 & \textbf{0.447} & 3.21 & \textbf{0.429} & 2.02 & \textbf{0.424} & 1.67 \\
\bottomrule
\end{tabular}%
}
\end{table}

\begin{table*}[t]
\caption{\textbf{Direct multi-view comparison on the main benchmarks.}
Point-map accuracy, completeness, and NC1 on 7Scenes and NRGBD; camera-pose
metrics on RE10K and CO3Dv2. Accuracy and completeness are lower-is-better;
all other metrics are higher-is-better. Zebra3D uses the training-free
0.5L/0.6S configuration for each backbone.}
\label{tab:direct-multiview-main-benchmarks}
\centering
\scriptsize
\setlength{\tabcolsep}{2.5pt}
\renewcommand{\arraystretch}{1.04}
\resizebox{\textwidth}{!}{%
\begin{tabular}{@{}l ccc ccc ccc cccc@{}}
\toprule
\textbf{Method}
& \multicolumn{3}{c}{\textbf{7Scenes}}
& \multicolumn{3}{c}{\textbf{NRGBD}}
& \multicolumn{3}{c}{\textbf{RE10K}}
& \multicolumn{4}{c}{\textbf{CO3Dv2}} \\
\cmidrule(lr){2-4}\cmidrule(lr){5-7}\cmidrule(lr){8-10}\cmidrule(lr){11-14}
& \textbf{Acc.} $\downarrow$ & \textbf{Comp.} $\downarrow$ & \textbf{NC1} $\uparrow$
& \textbf{Acc.} $\downarrow$ & \textbf{Comp.} $\downarrow$ & \textbf{NC1} $\uparrow$
& \textbf{AUC-30} $\uparrow$ & \textbf{RRA-30} $\uparrow$ & \textbf{RTA-30} $\uparrow$
& \textbf{AUC-3} $\uparrow$ & \textbf{AUC-5} $\uparrow$
& \textbf{AUC-15} $\uparrow$ & \textbf{AUC-30} $\uparrow$ \\
\midrule
Fast3R & 0.035 & 0.043 & 0.522 & 0.074 & 0.030 & 0.539
& 0.605 & 0.984 & 0.825 & 0.226 & 0.357 & 0.642 & 0.767 \\
MUSt3R & 0.045 & 0.321 & 0.547 & 0.070 & 0.127 & 0.654
& 0.796 & 1.000 & 0.934 & 0.356 & 0.496 & 0.742 & 0.839 \\
Spann3R & 0.124 & 0.110 & 0.553 & 0.248 & 0.173 & 0.578
& 0.292 & 0.991 & 0.542 & 0.006 & 0.025 & 0.208 & 0.412 \\
\midrule
\rowcolor{zebralavender}
\textbf{Zebra3D-VGGT (0.5L, 0.6S)}
& 0.017 & 0.027 & 0.606 & 0.018 & 0.013 & 0.746
& 0.737 & 0.999 & 0.919 & 0.554 & 0.665 & 0.834 & 0.897 \\
\rowcolor{zebralavender}
\textbf{Zebra3D-HYWM (0.5L, 0.6S)}
& 0.013 & 0.025 & 0.608 & 0.019 & 0.014 & 0.744
& 0.856 & 0.999 & 0.953 & 0.400 & 0.541 & 0.774 & 0.862 \\
\rowcolor{zebralavender}
\textbf{Zebra3D-Pi3 (0.5L, 0.6S)}
& 0.012 & 0.018 & 0.606 & 0.014 & 0.009 & 0.731
& 0.871 & 1.000 & 0.955 & 0.563 & 0.673 & 0.839 & 0.900 \\
\bottomrule
\end{tabular}%
}
\end{table*}

%% file: tables/appendix_scannet_quality_speed.tex
\subsection{ScanNet-50 Quality--Speed Trade-off}
\label{app:scannet-quality-speed}
\begin{figure}[!htbp]
\centering
\includegraphics[width=0.78\linewidth]{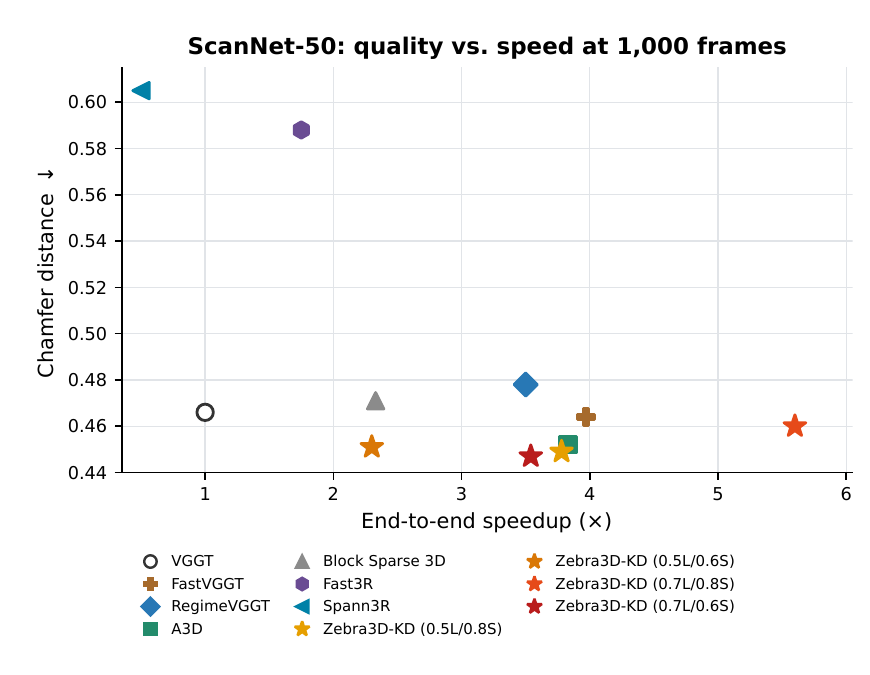}
\caption{\textbf{ScanNet-50 quality--speed trade-off at $1{,}000$ frames.}
Chamfer distance versus end-to-end speedup relative to dense VGGT. Lower
Chamfer distance and higher speedup are preferred.}
\label{fig:scannet1000-quality-speed}
\end{figure}

Figure~\ref{fig:scannet1000-quality-speed} visualizes the 1,000-frame results
from Table~\ref{tab:scannet50-results}, augmented with Fast3R and Spann3R from
Table~\ref{tab:direct-multiview-scannet50}. The two \methodns-KD 0.7L
configurations define complementary operating points: 0.7L/0.6S achieves the
lowest Chamfer distance ($0.447$) at $3.54\times$ speedup, while 0.7L/0.8S
reaches $5.60\times$ speedup with a Chamfer distance of $0.460$.

%% file: tables/appendix_implementation_details.tex
\subsection{Backbones and conversion}
We use the released checkpoints \texttt{facebook/VGGT-1B}~\citep{wang2025vggt},
\texttt{tencent/HunyuanWorld-Mirror}~\citep{liu2026worldmirror}, and
\texttt{yyfz233/Pi3}~\citep{wang2026pi3}.  Conversion
preserves every compatible pretrained parameter, including the input encoder,
$QKV$ and output projections, normalization layers, positional encoding, and
prediction heads. We rank global stages by the single-stage linear-conversion error described in
Equation~\ref{eq:layer-conversion-error}, then assign the requested fraction to
linear-centered attention; the remaining stages use tile-sparse attention.
\ruijing{Table~\ref{tab:layer-role-configs} lists the resulting zero-based
0.5L and 0.7L assignments, which match the released configurations. Every global-attention stage is replaced by either linear-centered or tile-sparse attention.}

\input{tables/appendix_layer_role_configurations}

In the $x$L/$y$S shorthand of Section~\ref{sec:experiments}, the selector
retains a key-tile fraction $\rho=1-y$; the four reported operating points
are
\begin{center}
\small
\begin{tabular}{@{}lccc@{}}
\toprule
Shorthand & $x$ & $y$ & $\rho{=}1-y$ \\
\midrule
$0.5$L/$0.6$S & $0.5$ & $0.6$ & $0.4$ \\
$0.5$L/$0.8$S & $0.5$ & $0.8$ & $0.2$ \\
$0.7$L/$0.6$S & $0.7$ & $0.6$ & $0.4$ \\
$0.7$L/$0.8$S & $0.7$ & $0.8$ & $0.2$ \\
\bottomrule
\end{tabular}
\end{center}

\subsection{Baselines}
The dense baselines are the unmodified released VGGT, HYWM, and $\pi^3$
models~\citep{wang2025vggt,liu2026worldmirror,wang2026pi3}.  We evaluate
FastVGGT at its \texttt{m0} operating point~\citep{shen2026fastvggt} and SAF3R
at its \texttt{t4} operating point using the published head-to-pattern
assignments~\citep{deng2026saf3r}.  A3D denotes AVGGT-f4 on VGGT and the same
frame-attention/key--value subsampling operator ported without architectural
changes to HYWM and $\pi^3$~\citep{sun2026avggt}.  \ruijing{A3D converts the
first global stages to frame attention and runs the rest as subsampled global
attention (SGA): every query is kept, but it attends only to the special
tokens, the intact reference frame, a spatially subsampled grid of patch keys
and values from the other frames, and one mean-fill token that averages the
dropped keys and values. AVGGT releases no code, so A3D is our reimplementation of the published f4 setting; the same operator is applied unchanged to HYWM and $\pi^3$.} Block Sparse 3D similarly denotes the published adaptive block-sparse operator and our
architecture-matched ports at the sp70 operating point
\citep{wang2026blocksparse}.  We evaluate RegimeVGGT at its released canonical
AB2 setting on VGGT and, because HYWM has the same 24-stage alternating
aggregator, apply the operator unchanged to HYWM; the latter is an
architecture-matched extension not reported by the RegimeVGGT
paper~\citep{you2026regimevggt}.  Lite3R is evaluated from its released
configuration~\citep{zhang2026lite3r}.  We report a method only on backbones
that its released implementation or our architecture-matched operator supports.
Every baseline is evaluated through the same data loaders, frame selections,
alignment, and metric code as \method.

\subsection{Linear initialization}
\label{app:linear-init}
For each linear stage, we first run the frozen dense model on 13 held-out
multi-view calibration scenes.  Teacher softmax outputs and converted linear
outputs are accumulated across scenes, tokens, heads, and views.  We then fit
the centered affine correction in Equation~\ref{eq:linear-projection-fit} by a
regularized least-squares solve in float64.  The fit requires one teacher
forward pass and one solve per converted stage; it does not update inherited
weights by gradient descent.  These fitted projections, together with copied
dense-model weights, define the training-free Zebra3D models.

\subsection{Triton implementation of hybrid attention}
\label{app:triton}
We implement Zebra3D's linear-centered and tile-sparse operators in Triton
to reduce intermediate memory traffic and execute their computations
efficiently on the GPU.  For linear-centered attention, we fuse Q/K
normalization, positional rotation, and channel-softmax feature mapping into
the kernels that accumulate the linear-attention state and compute the
readout.  This avoids materializing intermediate normalized Q/K tensors
between separate preprocessing and attention kernels.

Sparse stages partition each image into $8{\times}8$-patch tiles, or 64 tokens
per full tile.  We pack the full key/value sequence into tile-major order
before selecting key blocks independently for each head and query block.
The retention parameter is $\rho{=}0.2$ or $0.4$, corresponding to
$y{=}0.8$ or $0.6$.  Unselected packed blocks contribute one key--value
landmark each; special-token keys remain visible, special-token query rows
retain dense attention, and every patch query retains its self key.
Training and evaluation use the packed \texttt{triton} implementation
with 64-token query and key blocks, head-level selection, one view per
selection group.

The operator first constructs a permutation from the model's view-major token
layout to tile-major order.  Complete spatial tiles occupy aligned 64-token
blocks; tiles truncated at the image boundary are concatenated in a ragged
region and may share or span packed blocks.  Camera and register tokens are
appended as special tokens.  The permutation, its inverse, the tile-to-block
map, and special-token indices are cached for each frame count and patch grid.
This avoids rebuilding layout-dependent index tensors and repeating the
device synchronizations needed to construct them.  At inference, a fused
packing launch makes $Q$ contiguous and permutation-packs $K$ and $V$
together.  The attention kernel can then read repeatedly used $K/V$ blocks
with coalesced memory accesses, while query and output rows are mapped through
the cached permutation.

A fused statistics kernel reads BF16 $Q$, $K$, and $V$ and accumulates in
FP32.  For each packed block and head, it emits the mean query and key used
by the selector, a mean key--value landmark over non-special tokens, and the
log occupancy.  These blocks coincide with spatial tiles in the aligned
region but can combine boundary-tile tokens in the ragged region.  Each query
block ranks key blocks by the mean-query/mean-key dot product and stores the
selected indices together with a chosen-block mask.  The selection count is
$\lceil\rho N/64\rceil$, capped by the number of eligible non-special blocks
and kept at least one when an eligible block exists.

The main Triton program implements FlashAttention-style online softmax.  It
streams four contributions without constructing an
$N{\times}N$ score matrix: non-special tokens in selected key blocks, every
special-token key, the query's own key when its block was not selected, and
one occupancy-weighted landmark for each unselected block.  The landmark
logit receives a $\log n_b$ offset, where $n_b$ is the number of valid
non-special tokens represented by that landmark.  The chosen-block mask
excludes selected blocks from the landmark tail, while special-token masks
exclude special keys from the selected-block stream.  When an explicit self
key is added, it is not subtracted from its block's landmark summary.
Softmax state and output accumulators remain in FP32;
BF16 matrix products use the accelerator's matrix instructions.  The kernel
works in base-two exponent units, folding $\log_2 e$ into the attention scale
and using the hardware \texttt{exp2} instruction.  The shipped forward launch
uses two warps, two pipeline stages, and a 64-column landmark tile.  Special
query rows are evaluated exactly by a separate dense-row kernel and written
into the same output tensor.  This kernel distributes key tiles across
workers and combines their partial softmax results after rescaling to a
common maximum.  On supported GPU layouts, it uses four warps, two pipeline
stages, and explicit base-two exponential evaluation while retaining
attention over every key.

Training uses a custom autograd bridge for BF16 $[B,H,N,64]$ $Q/K/V$,
64-token blocks, per-head selection, and a landmark tail.  The forward retains the selected-block indices, chosen masks, landmark
statistics, output, and row-wise log-sum-exp.  Separate Triton kernels compute
$dQ$ and the owner-accumulated $dK/dV$ in FP32, followed by dedicated gradient
kernels for exact special-query rows; gradients are cast back to the input
dtype at the boundary.  The sparse call is an explicit compiler graph break so
that raw Triton launches are not captured or rewritten, while the surrounding
backbone remains compilable.  The reported 3,000-step KD runs use this
autograd bridge, whereas inference calls the forward kernel directly.

\subsection{Relational knowledge distillation}
\label{app:relkd}

All \methodns-KD models are initialized from the corresponding training-free
0.5L or 0.7L conversion
(Table~\ref{tab:layer-role-configs}), using copied dense weights and the
float64 linear-centered projection fit
(Appendix~\ref{app:linear-init}).
Each student is refined for $3{,}000$ optimizer steps with all parameters
trainable. The original unconverted checkpoint serves as a frozen FP32
teacher, while the student uses BF16 mixed precision.

\paragraph{Distillation loss definitions.}
For the output modalities shared by teacher and student, we use
\begin{equation}
\mathcal{L}_{\mathrm{out}}
=\sum_{u\in\mathcal{U}}\alpha_u
\frac{\left\|M_u\odot(y_S^{(u)}-y_T^{(u)})\right\|_1}
{\left\|M_u\odot y_T^{(u)}\right\|_1+\epsilon},
\label{eq:output-distillation}
\end{equation}
where $M_u$ masks invalid entries. Each geometric-map term has weight
$1$, and each included camera term has weight $0.5$; the terms available
for each backbone are listed below. Intermediate feature matching uses
\begin{equation}
\mathcal{L}_1=\frac{1}{K}\sum_{k=1}^{K}
\frac{\|h_S^{(k)}-h_T^{(k)}\|_1}{\|h_T^{(k)}\|_1+\epsilon}.
\label{eq:feat-l1}
\end{equation}
For cosine and relational matching at sampled linear-centered stages,
we first remove special tokens and normalize patch features separately
for each view:
\begin{equation}
\overline{h}_v=\frac{1}{T}\sum_{j=1}^{T}h_{v,j},
\qquad
z_{v,i}=\frac{h_{v,i}-\overline{h}_v}
{\|h_{v,i}-\overline{h}_v\|_2+\epsilon}.
\label{eq:relkd-normalization}
\end{equation}
The cosine term is
$\mathcal{L}_{\mathrm{cos}}
=\mathbb{E}_{v,i}[1-z_{S,v,i}^{\top}z_{T,v,i}]$.
Both cosine and relational losses are averaged over the sampled stages.

\paragraph{Output matching.}
\begin{revision}
Equation~\ref{eq:output-distillation} is evaluated only for modalities
emitted by both models; missing heads are omitted.
For HYWM, output KD matches \texttt{pts3d}, \texttt{depth}, and
\texttt{normals} with unit weight and the 9D pose encoding
\texttt{camera\_params} with weight $0.5$.
For VGGT, it matches \texttt{point\_map} and \texttt{depth\_map} with unit
weight and \texttt{camera\_params}, \texttt{extrinsic}, and
\texttt{intrinsic}, each with weight $0.5$; the latter two are decoded from
\texttt{camera\_params}.
For $\pi^3$, it matches \texttt{point\_map} and \texttt{depth\_map} with
unit weight and the $3{\times}4$ world-to-camera \texttt{extrinsic} with
weight $0.5$; \texttt{camera\_params} and \texttt{intrinsic} are absent.
\end{revision}

\paragraph{Feature taps.}
\begin{revision}
For HYWM, feature-L1 matching
(Equation~\ref{eq:feat-l1}) uses the DPT \texttt{token\_list}
at \texttt{intermediate\_idxs} $4,11,17,23$.
For VGGT, it uses all 24 global-block outputs; for $\pi^3$, it uses all
18 global-stage outputs, including tile-sparse stages.
All feature-L1 taps are evaluated on every microbatch.

Centered cosine and relational matching instead use only the
linear-centered stages specified in
Table~\ref{tab:layer-role-configs}.
At optimizer step $t$, we sample $m=4$ such stages without replacement
using a deterministic permutation seeded by $10007+t$.
Both losses are averaged over the sampled stages; tile-sparse stages are
never candidates. For $\pi^3$, global stage $i$ corresponds to decoder block
$2i+1$. The four accumulated microbatches share the same step-seeded stage
subset, while token pairs are resampled on each microbatch.
\end{revision}

\paragraph{Cross-view pair sampling.}
\begin{revision}
At each sampled stage, features are reshaped to $[B,V,T,d]$,
camera and register tokens are removed, and patch features are normalized
as in Equation~\ref{eq:relkd-normalization}.
For each batch item and sampled stage, we sample $A=192$ patch-token anchors.
For each anchor, $P=64$ candidate indices are sampled independently.
A same-view draw is deterministically remapped by one view stride, and the
teacher retains the $k=8$ most similar candidates from this pool.
We additionally draw $R=24$ candidates by the same procedure, without
deduplicating them against the top-$k$ set.
Equation~\ref{eq:relkd} averages Smooth-L1 over the resulting
$32$ pairs per anchor.
The teacher selects the high-similarity candidates, and teacher and
student relations are evaluated on identical token pairs.
We use PyTorch Smooth-L1 with transition parameter
$\beta_{\mathrm{Huber}}=1$; if $V<2$, the relation term is zero.
We do not distill attention matrices, query/key alignments, or
tile-selection scores.
\end{revision}

\paragraph{Distillation schedule.}
\begin{revision}
\ruijing{The distillation weight decays linearly from $1.0$ to $0.6$ over the
$3{,}000$ optimizer updates:}
\begin{equation}
\lambda_{\mathrm{KD}}(t)
=
1.0-0.4\,\frac{t}{3000},
\qquad 0\leq t<3000,
\end{equation}
\ruijing{where $t$ counts optimizer updates.}
\end{revision}

\subsection{Training data and optimization}
\label{app:kd-training}

\paragraph{Data and augmentation.}
All KD runs use the same $20\%$ preprocessed subset of the
HunyuanWorld-Mirror training mixture.
The constituent datasets and their sampling weights are:
HyperSim~\citep{roberts2021hypersim} (400),
TartanAir~\citep{wang2020tartanair} (300),
Matterport3D~\citep{chang2017matterport3d} (250),
DL3DV-10K~\citep{ling2024dl3dv} (250),
CO3D training split~\citep{reizenstein2021co3d} (250),
BlendedMVS~\citep{yao2020blendedmvs} (200),
UnrealStereo4K~\citep{tosi2021smdnets} (150),
MVS-Synth~\citep{huang2018deepmvs} (150),
WildRGB-D~\citep{xia2024wildrgbd} (100),
ARKitScenes~\citep{baruch2021arkitscenes} (200),
MapFree~\citep{arnold2022mapfree} (150), and
MegaDepth~\citep{li2018megadepth} (150).
Each GPU receives up to $16$ views per microbatch.
Images have aspect ratios in $[0.5,2.0]$ and resolutions of
$100$k--$250$k pixels.
Spatial cropping and photometric augmentation follow the common
multi-view data pipeline.
Color jitter is applied with probability $0.9$, with brightness,
contrast, and saturation magnitudes of $0.5$ and hue magnitude of $0.1$;
the pipeline also includes random grayscale augmentation.

\paragraph{Calibration scenes.}
The reference stage assignment and the closed-form projection fit use 13 multi-view scenes shipped with the HunyuanWorld-Mirror examples.
No gradient update is run on them, and none of them enters a reported metric table.
Appendix~\ref{app:calibration-dataset-stability} repeats the sensitivity profiling on 7Scenes, DTU, RealEstate10K, NRGBD, and CO3Dv2 and compares those rankings with this fixed reference.
It does not change the assignment.

\paragraph{Task supervision.}
\begin{revision}
Task supervision in Equation~\ref{eq:kd-objective} is backbone-specific,
because the three backbones do not predict the same quantities.

HYWM and VGGT use the shared camera, point-map, and depth objectives with
top-level weights $5.0$, $1.0$, and $1.0$, respectively.
These weights are inherited unchanged from the released HYWM and VGGT
training configurations rather than tuned for this work.  In raw magnitude
the camera term is $4$--$5\times$ smaller than the point term, so the factor
of $5.0$ aligns the two contributions rather than prioritizing pose: over
early training the weighted camera and point gradients stay within
$1.06$--$1.18\times$ of each other, and this ratio is nearly identical for
the dense and hybrid runs, so the weighting does not confound the
dense--hybrid comparisons of Section~\ref{sec:experiments}.
We depart from the released VGGT configuration in one respect: it disables
the point loss by default, whereas we enable all three terms.
Point and depth losses use $\gamma=1$, $\alpha=0.2$, and a
valid-range setting of $0.98$.
The point loss uses a normal-based gradient penalty, while the depth
loss uses a spatial-gradient penalty.
Normal-map and Gaussian-splat rendering task losses are disabled; for HYWM,
normal predictions nevertheless remain in output distillation with unit
weight, and the normal head is not frozen.

For $\pi^3$ we use its own published objective, because it predicts local
point maps and camera-to-world transforms rather than the world-coordinate
points, nine-vector pose encoding, and depth confidence that the shared
losses expect.  Local point maps are supervised by an inverse-depth-weighted
L1 against local ground truth after a scale $s^\star$ obtained by a detached
weighted alignment on $4{,}096$ resampled valid points.  A geometric normal
term built from local point neighborhoods is enabled on the high- and
mid-quality datasets of the mixture.  Cameras are supervised on all relative
pose pairs, using a Huber term on translations scaled by the same $s^\star$
and a rotation-angle term; the effective weights are $1$ on the point terms,
$10$ on translations, and $0.1$ on rotations.  Confidence supervision is
disabled, as are the shared camera, point, and depth losses.
\end{revision}

\paragraph{Optimization.}
Training uses eight AMD MI355X GPUs with DistributedDataParallel,
BF16 student mixed precision, synchronized batch normalization, and
four-step gradient accumulation.
Each GPU processes at most $16$ views per microbatch; the dynamic sequence
length sets the sequence batch size to $\lfloor16/V\rfloor$.
Thus $512$ views per optimizer update is an upper bound attained only when
every GPU microbatch fills the 16-view budget.
Training sequences contain at most $16$ views; evaluations on longer
sequences test generalization beyond this training sequence length.

We use AdamW with $(\beta_1,\beta_2)=(0.9,0.95)$,
$\epsilon=10^{-8}$, weight decay $0.1$, and gradient-norm clipping
at $1.0$.
The optimizer's \texttt{pretrained}, \texttt{backbone}, and
\texttt{new} parameter groups use peak learning rates
$10^{-5}$, $3\times10^{-5}$, and $10^{-4}$, respectively.
The \texttt{new} group contains $\operatorname{proj}_{\ell}$
and any mixer-only tensors.
All groups are trainable.
A $250$-step linear warmup is followed by cosine decay to $10\%$
of each group's peak learning rate by step $3{,}000$.
We use random seed $42$, save checkpoints every $500$ steps,
and perform no in-loop validation.
Jobs run in the ROCm environment with Triton $3.7.1$.

\paragraph{\ruijing{Fine-tuned dense control.}}
\ruijing{The VGGT + GT-FT reference reported in the main tables is
produced by this same procedure applied to the unconverted dense VGGT
checkpoint: the same $20\%$ HunyuanWorld-Mirror subset, the same augmentation
pipeline, the same $3{,}000$ steps with $250$-step warmup and cosine decay,
the same optimizer groups and peak learning rates, and the same seed.
The only differences are that no attention operator is replaced and that the
objective contains the task term alone, with all distillation terms disabled;
no frozen teacher is instantiated.
This control therefore matches the Zebra3D-KD training budget and supervision
exactly and differs only in architecture and in the presence of the relational
terms, which is what makes it usable both as a supervision-matched reference
for the main tables and as the dense counterpart of the GT-only row in
Table~\ref{tab:attention-operator-ablations}(c).}

\subsection{Evaluation datasets}
All indoor point-map evaluations use $518{\times}378$ inputs and fixed sequence
maps.  For 7Scenes~\citep{shotton2013seven} and NRGBD~\citep{azinovic2022neuralrgbd}, stride 10 selects every tenth frame; the
maps contain 18 and 9 test sequences, respectively. Depth, validity
masks, intrinsics, and camera poses are read from each benchmark's official
representation before the shared crop-and-resize transform.

RealEstate10K~\citep{zhou2018stereomagnification} pose evaluation uses a fixed
seed-42 map containing 1,719 test clips and ten frames per clip at
$518{\times}378$.  CO3Dv2~\citep{reizenstein2021co3d} follows the VGGT
ten-frame protocol: sequences are selected with NumPy seed 0 from 41
categories, images are resized to width 518 with aspect-preserving center crop
and white padding, and results are macro-averaged over categories.

\subsection{ScanNet-50 long-sequence protocol}
\label{app:scannet50-protocol}
Table~\ref{tab:scannet50-results} follows the FastVGGT ScanNet-50 setting: the
same 50 scenes and the official frame-index maps at 1,000, 500, 300, and 100
frames.  Images are resized to width 518 with the height snapped to the patch
size of 14, which gives $518{\times}392$ inputs for ScanNet.  Chamfer distance
is computed against the ground-truth mesh after voxel downsampling at 0.05,
clipping at 0.5, and sampling 100k points, and is normalized by the
bounding-box diagonal.  For each scene we take the median over repeated forwards,
then the mean over the 50 scenes.
Chamfer distance uses these $518{\times}392$ inputs.
Speedup is measured once per mixer configuration at $518{\times}378$ with the harness in Appendix~\ref{app:runtime}, and the same number is reported for the refined and training-free students that share that configuration.
Figure~\ref{fig:inference-speedup} uses this harness at 1,024 frames.

\subsection{Metrics and alignment}
For point maps, we first estimate a similarity transform with Umeyama alignment
on valid corresponding pixels and refine it with ICP.  We then report
prediction-to-ground-truth distance as accuracy, ground-truth-to-prediction
distance as completeness, and normal consistency in the accuracy direction as
NC1; both means and medians are computed per scene before dataset aggregation.
For RealEstate10K, relative rotation and translation accuracies are computed
over all frame pairs at thresholds of $5^\circ$, $15^\circ$, and $30^\circ$.
AUC-30 integrates the cumulative distribution of the larger rotation or
translation error up to $30^\circ$.  CO3Dv2 uses the same relative-pose
construction and reports AUC at $3^\circ$, $5^\circ$, $15^\circ$, and
$30^\circ$.  Pose arithmetic and metric accumulation use float64.

\subsection{Runtime measurement}
\label{app:runtime}
Each end-to-end measurement uses one MI355X with BF16 inference and standard Inductor compilation, with graph breaks and eager fallback enabled.  Each method and frame count is measured in a fresh
process with a separate compiler cache.  We use five warmup runs and 20
device-event measurements, except at exactly 1,024 frames, where we use two
warmup runs and five measurements.  We report compute
speedup by dividing the separately measured dense median by the method
median for the same backbone and frame count.  The measured call includes
the model forward pass and all enabled geometry-prediction heads, but excludes
input loading and preprocessing.  HYWM Gaussian-splat outputs are disabled
for every method, and no teacher computation is present at inference.

%% file: tables/appendix_layer_role_configurations.tex
\begin{table}[t]
    \caption{Layer-role assignments for the all-efficient Zebra3D variants.}
    \label{tab:layer-role-configs}
    \centering
    \small
    \setlength{\tabcolsep}{4pt}
    \renewcommand{\arraystretch}{1.12}
    \begin{tabular}{@{}llp{0.35\linewidth}p{0.32\linewidth}@{}}
        \toprule
        \textbf{Backbone} & \textbf{Variant} &
        \textbf{Linear-centered stages} & \textbf{Tile-sparse stages} \\
        \midrule
        VGGT & 0.5L & 1--6, 8, 18, 20--23 & 0, 7, 9--17, 19 \\
             & 0.7L & 0--9, 17--23 & 10--16 \\
        \midrule
        HYWM & 0.5L & 1--2, 4--6, 8--9, 18, 20--23 &
                       0, 3, 7, 10--17, 19 \\
             & 0.7L & 0--10, 17--18, 20--23 & 11--16, 19 \\
        \midrule
        $\pi^3$ & 0.5L & 0--2, 4--9 & 3, 10--17 \\
                & 0.7L & 0--10, 16--17 & 11--15 \\
        \bottomrule
    \end{tabular}
\end{table}

%% file: tables/appendix_placement_ablation.tex
\subsection{Layer-Role Placement Ablation}
\label{app:placement-ablation}
Table~\ref{tab:app-placement-ablation} reports every camera-pose metric for
the placement study summarized in
Table~\ref{tab:attention-operator-ablations}(d). All rows use the
$0.5$L/$0.8$S budget, so each backbone keeps the same number of
linear and tile-sparse stages and the same closed-form initialization; only
the set of stages given the linear role changes. No row receives gradient
updates, so the sensitivity-guided rows
coincide with the training-free $0.5$L/$0.8$S entries of
Table~\ref{tab:main-combined-results}.

\begin{table*}[t]
\caption{Layer-role placement ablation on RealEstate10K and CO3Dv2 camera
pose at $0.5$L/$0.8$S. Best per backbone in \textbf{bold}.}
\label{tab:app-placement-ablation}
\centering
\scriptsize
\setlength{\tabcolsep}{3pt}
\renewcommand{\arraystretch}{1.04}
\resizebox{\textwidth}{!}{%
\begin{tabular}{@{}ll ccccccc cccc@{}}
\toprule
& & \multicolumn{7}{c}{\textbf{RealEstate10K Camera Pose}}
& \multicolumn{4}{c}{\textbf{CO3Dv2 Camera Pose}} \\
\cmidrule(lr){3-9}\cmidrule(lr){10-13}
\textbf{Backbone} & \textbf{Placement}
& \textbf{RRA-5} $\uparrow$ & \textbf{RRA-15} $\uparrow$
& \textbf{RRA-30} $\uparrow$ & \textbf{RTA-5} $\uparrow$
& \textbf{RTA-15} $\uparrow$ & \textbf{RTA-30} $\uparrow$
& \textbf{AUC-30} $\uparrow$ & \textbf{AUC-3} $\uparrow$
& \textbf{AUC-5} $\uparrow$ & \textbf{AUC-15} $\uparrow$
& \textbf{AUC-30} $\uparrow$ \\
\midrule
\rowcolor{zebralavender}
VGGT & \textbf{Sensitivity-guided (ours)} & \textbf{0.935} & \textbf{0.995} & \textbf{0.999} & \textbf{0.508} & \textbf{0.832} & \textbf{0.916} & \textbf{0.730} & \textbf{0.535} & \textbf{0.650} & \textbf{0.827} & \textbf{0.892} \\
 & Back-sparse & 0.901 & 0.984 & 0.993 & 0.313 & 0.676 & 0.831 & 0.590 & 0.052 & 0.111 & 0.355 & 0.553 \\
 & Front-sparse & 0.541 & 0.878 & 0.961 & 0.041 & 0.218 & 0.453 & 0.208 & 0.000 & 0.000 & 0.004 & 0.023 \\
 & Uniform & 0.485 & 0.842 & 0.948 & 0.017 & 0.116 & 0.294 & 0.115 & 0.000 & 0.000 & 0.006 & 0.030 \\
 & Reversed & 0.383 & 0.755 & 0.910 & 0.016 & 0.119 & 0.312 & 0.114 & 0.000 & 0.000 & 0.002 & 0.014 \\
 & Random (seed 101) & 0.783 & 0.947 & 0.983 & 0.080 & 0.330 & 0.585 & 0.311 & 0.002 & 0.009 & 0.089 & 0.220 \\
 & Random (seed 202) & 0.884 & 0.977 & 0.990 & 0.242 & 0.613 & 0.791 & 0.534 & 0.017 & 0.047 & 0.252 & 0.471 \\
 & Random (seed 303) & 0.377 & 0.752 & 0.907 & 0.016 & 0.114 & 0.303 & 0.108 & 0.000 & 0.000 & 0.003 & 0.017 \\
\midrule
\rowcolor{zebralavender}
HYWM & \textbf{Sensitivity-guided (ours)} & \textbf{0.984} & \textbf{0.997} & \textbf{0.999} & \textbf{0.758} & \textbf{0.907} & \textbf{0.951} & \textbf{0.850} & \textbf{0.376} & \textbf{0.521} & \textbf{0.763} & \textbf{0.854} \\
 & Back-sparse & 0.924 & 0.984 & 0.994 & 0.357 & 0.684 & 0.835 & 0.609 & 0.031 & 0.076 & 0.308 & 0.511 \\
 & Front-sparse & 0.574 & 0.857 & 0.951 & 0.041 & 0.212 & 0.442 & 0.200 & 0.000 & 0.000 & 0.006 & 0.029 \\
 & Uniform & 0.490 & 0.813 & 0.932 & 0.027 & 0.150 & 0.340 & 0.142 & 0.000 & 0.000 & 0.010 & 0.043 \\
 & Reversed & 0.506 & 0.815 & 0.933 & 0.023 & 0.143 & 0.340 & 0.137 & 0.000 & 0.000 & 0.003 & 0.020 \\
 & Random (seed 101) & 0.506 & 0.810 & 0.931 & 0.030 & 0.167 & 0.384 & 0.159 & 0.000 & 0.000 & 0.005 & 0.023 \\
 & Random (seed 202) & 0.825 & 0.969 & 0.992 & 0.109 & 0.386 & 0.621 & 0.356 & 0.005 & 0.021 & 0.202 & 0.415 \\
 & Random (seed 303) & 0.367 & 0.705 & 0.869 & 0.028 & 0.148 & 0.337 & 0.132 & 0.000 & 0.000 & 0.010 & 0.043 \\
\midrule
\rowcolor{zebralavender}
$\pi^3$ & \textbf{Sensitivity-guided (ours)} & \textbf{0.988} & \textbf{0.998} & \textbf{1.000} & \textbf{0.792} & \textbf{0.917} & \textbf{0.953} & \textbf{0.866} & \textbf{0.542} & \textbf{0.657} & \textbf{0.831} & \textbf{0.895} \\
 & Back-sparse & 0.987 & \textbf{0.998} & \textbf{1.000} & 0.786 & \textbf{0.917} & \textbf{0.953} & 0.864 & 0.541 & 0.656 & 0.830 & \textbf{0.895} \\
 & Front-sparse & 0.111 & 0.417 & 0.659 & 0.005 & 0.043 & 0.158 & 0.033 & 0.000 & 0.000 & 0.001 & 0.012 \\
 & Uniform & 0.518 & 0.827 & 0.926 & 0.023 & 0.146 & 0.357 & 0.148 & 0.001 & 0.004 & 0.052 & 0.158 \\
 & Reversed & 0.107 & 0.404 & 0.649 & 0.005 & 0.044 & 0.160 & 0.033 & 0.000 & 0.000 & 0.002 & 0.012 \\
 & Random (seed 101) & 0.225 & 0.640 & 0.842 & 0.004 & 0.038 & 0.140 & 0.040 & 0.000 & 0.000 & 0.001 & 0.012 \\
 & Random (seed 202) & 0.189 & 0.528 & 0.748 & 0.011 & 0.081 & 0.240 & 0.067 & 0.000 & 0.000 & 0.005 & 0.028 \\
 & Random (seed 303) & 0.401 & 0.831 & 0.951 & 0.007 & 0.067 & 0.226 & 0.081 & 0.000 & 0.000 & 0.003 & 0.023 \\
\bottomrule
\end{tabular}%
}
\end{table*}

%% file: tables/appendix_calibration_stability.tex
\subsection{Calibration-Set Stability of Layer Rankings}
\label{app:calibration-stability}
\label{app:calibration-dataset-stability}
\ruijing{We test whether the stage assignment depends on the data used to measure
conversion sensitivity. For each backbone, the single-stage profiling of
Equation~\ref{eq:layer-conversion-error} is repeated with each of five
calibration datasets, and the resulting stage ranking and operator assignment
are compared with the reference Zebra3D ones
(Tables~\ref{tab:app-calibration-spearman} and~\ref{tab:app-calibration-agreement}).
Rankings remain strongly correlated with the reference, with Spearman's $\rho$
ranging from $0.7765$ to $0.9365$ and dataset means from $0.8425$ to $0.8865$.
Mean assignment agreement ranges from $81.5\%$ to $88.9\%$ at 0.5L and from
$93.5\%$ to $97.2\%$ at 0.7L, so the more linear configuration is especially
stable. The division into linear-tolerant and sensitive stages is therefore not
specific to one calibration collection.}

\begin{table}[t]
\caption{Spearman correlation between layer-sensitivity rankings obtained
from each calibration dataset and the reference Zebra3D rankings.}
\label{tab:app-calibration-spearman}
\centering
\small
\setlength{\tabcolsep}{3.5pt}
\renewcommand{\arraystretch}{1.04}
\begin{tabular}{@{}lcccc@{}}
\toprule
\textbf{Calibration dataset} & \textbf{VGGT} & \textbf{HYWM}
& \textbf{$\pi^3$} & \textbf{Mean} \\
\midrule
7Scenes & 0.9322 & 0.8965 & 0.8308 & 0.8865 \\
DTU & 0.8852 & 0.9278 & 0.8369 & 0.8833 \\
Re10K & 0.9183 & 0.7765 & 0.8328 & 0.8425 \\
NRGBD & 0.8878 & 0.8635 & 0.8225 & 0.8579 \\
CO3Dv2 & 0.9365 & 0.9235 & 0.7874 & 0.8825 \\
\bottomrule
\end{tabular}
\end{table}

\begin{table}[t]
\caption{Agreement with the reference 0.5L and 0.7L operator assignments,
measured as the fraction of stages receiving the same linear or tile-sparse
label.}
\label{tab:app-calibration-agreement}
\centering
\small
\setlength{\tabcolsep}{3.5pt}
\renewcommand{\arraystretch}{1.04}
\begin{tabular}{@{}lcccccccc@{}}
\toprule
& \multicolumn{4}{c}{\textbf{0.5L}} & \multicolumn{4}{c}{\textbf{0.7L}} \\
\cmidrule(lr){2-5}\cmidrule(lr){6-9}
\textbf{Calibration dataset} & \textbf{VGGT} & \textbf{HYWM}
& \textbf{$\pi^3$} & \textbf{Mean} & \textbf{VGGT} & \textbf{HYWM}
& \textbf{$\pi^3$} & \textbf{Mean} \\
\midrule
7Scenes & 91.7\% & 83.3\% & 88.9\% & 88.0\% & 91.7\% & 100.0\% & 100.0\% & 97.2\% \\
DTU & 83.3\% & 91.7\% & 77.8\% & 84.3\% & 91.7\% & 91.7\% & 100.0\% & 94.4\% \\
Re10K & 91.7\% & 75.0\% & 88.9\% & 85.2\% & 91.7\% & 100.0\% & 100.0\% & 97.2\% \\
NRGBD & 83.3\% & 83.3\% & 100.0\% & 88.9\% & 91.7\% & 100.0\% & 100.0\% & 97.2\% \\
CO3Dv2 & 83.3\% & 83.3\% & 77.8\% & 81.5\% & 91.7\% & 100.0\% & 88.9\% & 93.5\% \\
\bottomrule
\end{tabular}
\end{table}

%% file: tables/attention_ablation_appendix.tex
\subsection{Zero-Shot Linear-Attention Ablation}
\label{app:zero-shot-linear-ablation}
Tables~\ref{tab:app-linear-7scenes}--\ref{tab:app-linear-re10k} report the
zero-shot linear-attention ablations across all three backbones and the 0.5L
and 0.7L assignments. Lavender marks linear-centered attention, and the best
result within each backbone--assignment group is bold.
Linear-centered attention achieves the best RealEstate10K AUC-30 in every
backbone--assignment group, with especially large gains over frame attention
at 0.7L. Point-map results are more metric-dependent, but fitted linear
attention generally avoids the severe 0.7L degradation of frame attention;
centering provides the strongest overall robustness across reconstruction and
pose.

\begin{table*}[t]
\caption{Linear-attention ablation on 7Scenes.}
\label{tab:app-linear-7scenes}
\centering
\scriptsize
\setlength{\tabcolsep}{3pt}
\renewcommand{\arraystretch}{1.04}
\resizebox{\textwidth}{!}{%
\begin{tabular}{@{}llcccccccc@{}}
\toprule
\textbf{Configuration} & \textbf{Attention choice} & \textbf{Acc. mean $\downarrow$} & \textbf{Acc. med. $\downarrow$} & \textbf{Comp. mean $\downarrow$} & \textbf{Comp. med. $\downarrow$} & \textbf{NC1 mean $\uparrow$} & \textbf{NC1 med. $\uparrow$} & \textbf{NC2 mean $\uparrow$} & \textbf{NC2 med. $\uparrow$} \\
\midrule
VGGT (0.5L/0.8S) & Frame attention & 0.019 & 0.007 & \textbf{0.025} & \textbf{0.009} & 0.605 & 0.664 & 0.600 & 0.654 \\
 & Linear, no fit & 0.018 & 0.007 & 0.027 & 0.012 & \textbf{0.606} & \textbf{0.665} & 0.602 & 0.658 \\
 & Linear, uncentered fit & 0.018 & 0.006 & 0.028 & 0.012 & 0.605 & 0.665 & 0.604 & 0.661 \\
\rowcolor{zebralavender}
 & Linear-centered + fit & \textbf{0.018} & \textbf{0.006} & 0.027 & 0.011 & 0.605 & 0.664 & \textbf{0.604} & \textbf{0.661} \\
\addlinespace[2pt]
VGGT (0.7L/0.8S) & Frame attention & 0.099 & 0.062 & 0.033 & 0.010 & \textbf{0.609} & \textbf{0.674} & 0.591 & 0.641 \\
 & Linear, no fit & 0.018 & 0.007 & 0.025 & 0.010 & 0.606 & 0.666 & 0.601 & 0.656 \\
 & Linear, uncentered fit & \textbf{0.016} & \textbf{0.006} & 0.026 & 0.011 & 0.607 & 0.667 & \textbf{0.606} & \textbf{0.664} \\
\rowcolor{zebralavender}
 & Linear-centered + fit & 0.017 & 0.006 & \textbf{0.025} & \textbf{0.008} & 0.607 & 0.669 & 0.603 & 0.660 \\
\addlinespace[2pt]
HYWM (0.5L/0.8S) & Frame attention & 0.014 & 0.006 & \textbf{0.022} & \textbf{0.007} & 0.607 & 0.667 & 0.602 & 0.659 \\
 & Linear, no fit & 0.013 & 0.005 & 0.025 & 0.010 & 0.608 & 0.669 & 0.607 & 0.666 \\
 & Linear, uncentered fit & 0.013 & 0.005 & 0.025 & 0.010 & \textbf{0.608} & \textbf{0.669} & 0.608 & 0.668 \\
\rowcolor{zebralavender}
 & Linear-centered + fit & \textbf{0.013} & \textbf{0.005} & 0.024 & 0.010 & 0.607 & 0.668 & \textbf{0.608} & \textbf{0.668} \\
\addlinespace[2pt]
HYWM (0.7L/0.8S) & Frame attention & 0.098 & 0.059 & 0.031 & 0.009 & \textbf{0.614} & \textbf{0.683} & 0.598 & 0.652 \\
 & Linear, no fit & \textbf{0.012} & 0.005 & 0.023 & 0.008 & 0.610 & 0.672 & 0.609 & 0.669 \\
 & Linear, uncentered fit & 0.013 & 0.005 & \textbf{0.023} & \textbf{0.008} & 0.610 & 0.672 & \textbf{0.609} & \textbf{0.669} \\
\rowcolor{zebralavender}
 & Linear-centered + fit & 0.013 & \textbf{0.005} & 0.023 & 0.009 & 0.609 & 0.671 & 0.609 & 0.669 \\
\addlinespace[2pt]
$\pi^3$ (0.5L/0.8S) & Frame attention & 0.013 & 0.005 & \textbf{0.017} & \textbf{0.006} & 0.606 & 0.666 & 0.598 & 0.652 \\
 & Linear, no fit & 0.012 & 0.004 & 0.019 & 0.008 & 0.606 & 0.666 & 0.601 & 0.657 \\
 & Linear, uncentered fit & 0.012 & \textbf{0.004} & 0.019 & 0.007 & \textbf{0.606} & \textbf{0.666} & \textbf{0.602} & \textbf{0.658} \\
\rowcolor{zebralavender}
 & Linear-centered + fit & \textbf{0.012} & 0.004 & 0.018 & 0.007 & 0.606 & 0.665 & 0.601 & 0.657 \\
\addlinespace[2pt]
$\pi^3$ (0.7L/0.8S) & Frame attention & 0.031 & 0.014 & \textbf{0.018} & 0.006 & \textbf{0.625} & \textbf{0.698} & \textbf{0.585} & \textbf{0.632} \\
 & Linear, no fit & 0.028 & 0.013 & 0.021 & 0.006 & 0.620 & 0.691 & 0.579 & 0.622 \\
 & Linear, uncentered fit & 0.024 & 0.010 & 0.019 & \textbf{0.006} & 0.619 & 0.689 & 0.582 & 0.626 \\
\rowcolor{zebralavender}
 & Linear-centered + fit & \textbf{0.021} & \textbf{0.008} & 0.020 & 0.006 & 0.613 & 0.679 & 0.585 & 0.630 \\
\bottomrule
\end{tabular}%
}
\end{table*}

\begin{table*}[t]
\caption{Linear-attention ablation on Neural RGB-D.}
\label{tab:app-linear-nrgbd}
\centering
\scriptsize
\setlength{\tabcolsep}{3pt}
\renewcommand{\arraystretch}{1.04}
\resizebox{\textwidth}{!}{%
\begin{tabular}{@{}llcccccccc@{}}
\toprule
\textbf{Configuration} & \textbf{Attention choice} & \textbf{Acc. mean $\downarrow$} & \textbf{Acc. med. $\downarrow$} & \textbf{Comp. mean $\downarrow$} & \textbf{Comp. med. $\downarrow$} & \textbf{NC1 mean $\uparrow$} & \textbf{NC1 med. $\uparrow$} & \textbf{NC2 mean $\uparrow$} & \textbf{NC2 med. $\uparrow$} \\
\midrule
VGGT (0.5L/0.8S) & Frame attention & 0.020 & 0.013 & \textbf{0.010} & \textbf{0.003} & 0.730 & 0.881 & 0.713 & 0.866 \\
 & Linear, no fit & 0.019 & 0.011 & 0.012 & 0.003 & 0.737 & 0.886 & 0.745 & 0.899 \\
 & Linear, uncentered fit & \textbf{0.017} & \textbf{0.010} & 0.011 & 0.003 & \textbf{0.743} & \textbf{0.891} & \textbf{0.757} & \textbf{0.906} \\
\rowcolor{zebralavender}
 & Linear-centered + fit & 0.021 & 0.013 & 0.014 & 0.004 & 0.739 & 0.883 & 0.744 & 0.881 \\
\addlinespace[2pt]
VGGT (0.7L/0.8S) & Frame attention & 0.201 & 0.143 & 0.041 & 0.011 & 0.672 & 0.829 & 0.695 & 0.857 \\
 & Linear, no fit & 0.036 & 0.020 & 0.019 & 0.007 & 0.719 & 0.875 & 0.722 & 0.878 \\
 & Linear, uncentered fit & \textbf{0.024} & \textbf{0.013} & \textbf{0.015} & \textbf{0.004} & \textbf{0.733} & \textbf{0.883} & \textbf{0.741} & \textbf{0.894} \\
\rowcolor{zebralavender}
 & Linear-centered + fit & 0.029 & 0.017 & 0.018 & 0.005 & 0.721 & 0.872 & 0.720 & 0.867 \\
\addlinespace[2pt]
HYWM (0.5L/0.8S) & Frame attention & 0.021 & 0.014 & \textbf{0.012} & \textbf{0.004} & 0.720 & 0.868 & 0.708 & 0.858 \\
 & Linear, no fit & 0.020 & 0.013 & 0.014 & 0.005 & 0.732 & 0.874 & 0.744 & 0.890 \\
 & Linear, uncentered fit & 0.020 & 0.013 & 0.014 & 0.005 & 0.734 & 0.876 & 0.751 & 0.898 \\
\rowcolor{zebralavender}
 & Linear-centered + fit & \textbf{0.019} & \textbf{0.012} & 0.013 & 0.005 & \textbf{0.740} & \textbf{0.880} & \textbf{0.762} & \textbf{0.906} \\
\addlinespace[2pt]
HYWM (0.7L/0.8S) & Frame attention & 0.201 & 0.150 & 0.043 & 0.012 & 0.681 & 0.848 & 0.695 & 0.859 \\
 & Linear, no fit & \textbf{0.022} & \textbf{0.013} & \textbf{0.012} & \textbf{0.003} & 0.721 & 0.866 & 0.728 & 0.876 \\
 & Linear, uncentered fit & 0.029 & 0.019 & 0.014 & 0.005 & 0.720 & 0.865 & 0.728 & 0.877 \\
\rowcolor{zebralavender}
 & Linear-centered + fit & 0.022 & 0.015 & 0.014 & 0.005 & \textbf{0.729} & \textbf{0.871} & \textbf{0.744} & \textbf{0.895} \\
\addlinespace[2pt]
$\pi^3$ (0.5L/0.8S) & Frame attention & 0.016 & 0.009 & 0.010 & 0.003 & 0.723 & 0.860 & 0.713 & 0.854 \\
 & Linear, no fit & 0.015 & 0.008 & 0.010 & 0.002 & 0.725 & 0.861 & 0.718 & 0.857 \\
 & Linear, uncentered fit & 0.015 & 0.008 & 0.010 & 0.002 & 0.725 & 0.861 & \textbf{0.720} & \textbf{0.858} \\
\rowcolor{zebralavender}
 & Linear-centered + fit & \textbf{0.015} & \textbf{0.008} & \textbf{0.010} & \textbf{0.002} & \textbf{0.726} & \textbf{0.862} & 0.718 & 0.858 \\
\addlinespace[2pt]
$\pi^3$ (0.7L/0.8S) & Frame attention & 0.051 & 0.032 & \textbf{0.012} & 0.004 & \textbf{0.710} & \textbf{0.871} & 0.677 & 0.824 \\
 & Linear, no fit & 0.071 & 0.044 & 0.023 & 0.005 & 0.700 & 0.857 & \textbf{0.680} & \textbf{0.828} \\
 & Linear, uncentered fit & 0.065 & 0.040 & 0.021 & 0.005 & 0.702 & 0.857 & 0.675 & 0.820 \\
\rowcolor{zebralavender}
 & Linear-centered + fit & \textbf{0.043} & \textbf{0.027} & 0.015 & \textbf{0.004} & 0.708 & 0.862 & 0.678 & 0.822 \\
\bottomrule
\end{tabular}%
}
\end{table*}

\begin{table*}[t]
\caption{Linear-attention ablation on RealEstate10K camera pose.}
\label{tab:app-linear-re10k}
\centering
\scriptsize
\setlength{\tabcolsep}{3pt}
\renewcommand{\arraystretch}{1.04}
\resizebox{\textwidth}{!}{%
\begin{tabular}{@{}llccccccc@{}}
\toprule
\textbf{Configuration} & \textbf{Attention choice} & \textbf{RRA-5 $\uparrow$} & \textbf{RRA-15 $\uparrow$} & \textbf{RRA-30 $\uparrow$} & \textbf{RTA-5 $\uparrow$} & \textbf{RTA-15 $\uparrow$} & \textbf{RTA-30 $\uparrow$} & \textbf{AUC-30 $\uparrow$} \\
\midrule
VGGT (0.5L/0.8S) & Frame attention & 0.932 & 0.994 & 0.999 & 0.481 & 0.818 & 0.907 & 0.714 \\
 & Linear, no fit & 0.928 & 0.994 & 0.999 & 0.471 & 0.822 & 0.912 & 0.714 \\
 & Linear, uncentered fit & 0.928 & 0.995 & \textbf{0.999} & 0.475 & 0.820 & 0.910 & 0.713 \\
\rowcolor{zebralavender}
 & Linear-centered + fit & \textbf{0.935} & \textbf{0.995} & \textbf{0.999} & \textbf{0.508} & \textbf{0.832} & \textbf{0.916} & \textbf{0.730} \\
\addlinespace[2pt]
VGGT (0.7L/0.8S) & Frame attention & 0.909 & 0.994 & 0.999 & 0.309 & 0.682 & 0.846 & 0.596 \\
 & Linear, no fit & 0.921 & 0.994 & 0.999 & 0.440 & 0.819 & 0.915 & 0.707 \\
 & Linear, uncentered fit & 0.920 & 0.993 & 0.999 & 0.457 & 0.821 & 0.914 & 0.710 \\
\rowcolor{zebralavender}
 & Linear-centered + fit & \textbf{0.932} & \textbf{0.995} & \textbf{1.000} & \textbf{0.459} & \textbf{0.828} & \textbf{0.918} & \textbf{0.718} \\
\addlinespace[2pt]
HYWM (0.5L/0.8S) & Frame attention & 0.979 & 0.996 & 0.998 & 0.708 & 0.890 & 0.942 & 0.823 \\
 & Linear, no fit & 0.984 & \textbf{0.997} & 0.999 & 0.741 & 0.903 & 0.948 & 0.840 \\
 & Linear, uncentered fit & 0.984 & \textbf{0.997} & 0.999 & 0.745 & 0.903 & 0.948 & 0.842 \\
\rowcolor{zebralavender}
 & Linear-centered + fit & \textbf{0.984} & 0.997 & \textbf{0.999} & \textbf{0.758} & \textbf{0.907} & \textbf{0.951} & \textbf{0.850} \\
\addlinespace[2pt]
HYWM (0.7L/0.8S) & Frame attention & 0.959 & 0.986 & 0.993 & 0.486 & 0.766 & 0.877 & 0.690 \\
 & Linear, no fit & 0.977 & 0.997 & 0.999 & 0.685 & 0.885 & 0.939 & 0.811 \\
 & Linear, uncentered fit & 0.977 & 0.997 & 0.999 & 0.692 & 0.887 & 0.939 & 0.814 \\
\rowcolor{zebralavender}
 & Linear-centered + fit & \textbf{0.983} & \textbf{0.998} & \textbf{1.000} & \textbf{0.739} & \textbf{0.903} & \textbf{0.948} & \textbf{0.839} \\
\addlinespace[2pt]
$\pi^3$ (0.5L/0.8S) & Frame attention & 0.986 & 0.997 & 0.999 & 0.783 & 0.916 & 0.952 & 0.862 \\
 & Linear, no fit & \textbf{0.988} & 0.998 & 1.000 & 0.788 & 0.917 & 0.952 & 0.864 \\
 & Linear, uncentered fit & 0.988 & 0.998 & \textbf{1.000} & 0.791 & 0.917 & 0.952 & 0.865 \\
\rowcolor{zebralavender}
 & Linear-centered + fit & 0.988 & \textbf{0.998} & 1.000 & \textbf{0.792} & \textbf{0.917} & \textbf{0.953} & \textbf{0.867} \\
\addlinespace[2pt]
$\pi^3$ (0.7L/0.8S) & Frame attention & 0.977 & 0.994 & 0.997 & 0.443 & 0.732 & 0.841 & 0.655 \\
 & Linear, no fit & 0.980 & 0.997 & 1.000 & 0.641 & 0.866 & 0.931 & 0.789 \\
 & Linear, uncentered fit & 0.983 & 0.997 & 0.999 & 0.682 & 0.883 & 0.936 & 0.811 \\
\rowcolor{zebralavender}
 & Linear-centered + fit & \textbf{0.985} & \textbf{0.998} & \textbf{1.000} & \textbf{0.725} & \textbf{0.901} & \textbf{0.948} & \textbf{0.836} \\
\bottomrule
\end{tabular}%
}
\end{table*}

\subsection{Training-Based Linear-Attention Ablation}
\label{app:trained-mixer-ablation}
\ruijing{Tables~\ref{tab:app-trained-mixer-pointmap} and~\ref{tab:app-trained-mixer-pose}}
compare HYWM mixers trained for 3,000 steps with the same data, optimization,
and distillation recipe. The comparison covers Mamba-2 \citep{dao2024transformers}, linear-centered
attention, Gated Delta Net \citep{yang2025gated}, and Sparse Linear Attn \citep{zhang2026sla}. Lavender marks the
linear-centered operator used by Zebra3D, and bold marks the best result in
each column\ruijing{ within each dataset}.
No mixer dominates every metric: Sparse Linear Attn performs best on NRGBD,
while Gated Delta Net leads several 7Scenes metrics. Linear-centered attention
achieves the best AUC-30 on both pose benchmarks ($0.852$ on RealEstate10K and
$0.849$ on CO3Dv2), making it the strongest balanced choice across
reconstruction and pose.

\begin{table*}[t]
\caption{Training-based linear-attention ablation on
7Scenes~\citep{shotton2013seven} and Neural
RGB-D~\citep{azinovic2022neuralrgbd}.}
\label{tab:app-trained-mixer-pointmap}
\centering
\scriptsize
\setlength{\tabcolsep}{3pt}
\renewcommand{\arraystretch}{1.04}
\resizebox{\textwidth}{!}{%
\begin{tabular}{@{}lcccccccc@{}}
\toprule
\textbf{Mixer} & \textbf{Acc. mean $\downarrow$} & \textbf{Acc. med. $\downarrow$}
& \textbf{Comp. mean $\downarrow$} & \textbf{Comp. med. $\downarrow$}
& \textbf{NC1 mean $\uparrow$} & \textbf{NC1 med. $\uparrow$}
& \textbf{NC2 mean $\uparrow$} & \textbf{NC2 med. $\uparrow$} \\
\midrule
\multicolumn{9}{l}{\textbf{7Scenes}} \\
Mamba-2 & 0.014 & 0.006 & 0.023 & 0.009 & 0.609 & 0.670 & 0.604 & 0.662 \\
\rowcolor{zebralavender}
Linear-centered & \textbf{0.013} & 0.005 & 0.023 & 0.009 & 0.607 & 0.667 & 0.606 & 0.664 \\
Gated Delta Net & 0.014 & 0.005 & \textbf{0.022} & \textbf{0.008}
& \textbf{0.609} & \textbf{0.670} & \textbf{0.607} & \textbf{0.666} \\
Sparse Linear Attn & 0.013 & \textbf{0.005} & 0.024 & 0.009 & 0.608 & 0.668 & 0.606 & 0.665 \\
\midrule
\multicolumn{9}{l}{\textbf{Neural RGB-D}} \\
Mamba-2 & 0.019 & 0.012 & 0.013 & 0.005 & 0.741 & 0.881 & 0.749 & 0.895 \\
\rowcolor{zebralavender}
Linear-centered & 0.019 & 0.012 & 0.014 & 0.005 & 0.746 & 0.885 & 0.769 & 0.910 \\
Gated Delta Net & 0.020 & 0.012 & 0.014 & 0.005 & 0.742 & 0.881 & 0.751 & 0.896 \\
Sparse Linear Attn & \textbf{0.016} & \textbf{0.010} & \textbf{0.013} & \textbf{0.004}
& \textbf{0.749} & \textbf{0.888} & \textbf{0.775} & \textbf{0.916} \\
\bottomrule
\end{tabular}%
}
\end{table*}

\begin{table*}[t]
\caption{Training-based linear-attention ablation on
RealEstate10K~\citep{zhou2018stereomagnification} and
CO3Dv2~\citep{reizenstein2021co3d} camera pose.}
\label{tab:app-trained-mixer-pose}
\centering
\scriptsize
\setlength{\tabcolsep}{3pt}
\renewcommand{\arraystretch}{1.04}
\resizebox{\textwidth}{!}{%
\begin{tabular}{@{}lccccccccccc@{}}
\toprule
& \multicolumn{7}{c}{\textbf{RealEstate10K}} & \multicolumn{4}{c}{\textbf{CO3Dv2}} \\
\cmidrule(lr){2-8}\cmidrule(lr){9-12}
\textbf{Mixer} & \textbf{RRA-5 $\uparrow$} & \textbf{RRA-15 $\uparrow$}
& \textbf{RRA-30 $\uparrow$} & \textbf{RTA-5 $\uparrow$}
& \textbf{RTA-15 $\uparrow$} & \textbf{RTA-30 $\uparrow$}
& \textbf{AUC-30 $\uparrow$} & \textbf{AUC-3 $\uparrow$}
& \textbf{AUC-5 $\uparrow$} & \textbf{AUC-15 $\uparrow$}
& \textbf{AUC-30 $\uparrow$} \\
\midrule
Mamba-2 & 0.985 & 0.998 & \textbf{1.000} & 0.746 & 0.907 & 0.952 & 0.845 & \textbf{0.345} & \textbf{0.496} & \textbf{0.752} & 0.848 \\
\rowcolor{zebralavender}
Linear-centered & 0.985 & 0.998 & 1.000 & \textbf{0.762} & \textbf{0.912} & \textbf{0.953} & \textbf{0.852} & 0.330 & 0.482 & 0.748 & \textbf{0.849} \\
Gated Delta Net & 0.984 & \textbf{0.998} & \textbf{1.000} & 0.711 & 0.900 & 0.948 & 0.828 & 0.305 & 0.461 & 0.735 & 0.839 \\
Sparse Linear Attn & \textbf{0.987} & 0.998 & 1.000 & 0.741 & 0.898 & 0.944 & 0.838 & 0.325 & 0.478 & 0.745 & 0.846 \\
\bottomrule
\end{tabular}%
}
\end{table*}

\subsection{Sparse-Attention Ablation}
\label{app:detailed-attention-ablations}
\label{app:sparse-attention-ablation}
Tables~\ref{tab:app-sparse-7scenes}--\ref{tab:app-sparse-re10k}
report the sparse-attention ablations across all three backbones at 0.5L and
0.7L, both with 0.75S. Lavender marks the operator selected for Zebra3D, and
the best result within each backbone--assignment group is bold.
Overall, the landmark tail improves sparse-attention quality, and tile sparse
with landmarks provides the strongest aggregate trade-off. It generally
outperforms block sparse and surpasses the SGA operator used by
A3D~\citep{sun2026avggt} on NRGBD and RealEstate10K, although SGA
remains competitive on some 7Scenes metrics.

\begin{table*}[t]
\caption{Sparse-attention ablation on 7Scenes.}
\label{tab:app-sparse-7scenes}
\centering
\scriptsize
\setlength{\tabcolsep}{3pt}
\renewcommand{\arraystretch}{1.04}
\resizebox{\textwidth}{!}{%
\begin{tabular}{@{}llcccccccc@{}}
\toprule
\textbf{Configuration} & \textbf{Attention choice} & \textbf{Acc. mean $\downarrow$} & \textbf{Acc. med. $\downarrow$} & \textbf{Comp. mean $\downarrow$} & \textbf{Comp. med. $\downarrow$} & \textbf{NC1 mean $\uparrow$} & \textbf{NC1 med. $\uparrow$} & \textbf{NC2 mean $\uparrow$} & \textbf{NC2 med. $\uparrow$} \\
\midrule
VGGT (0.5L/0.75S) & SGA & \textbf{0.016} & \textbf{0.005} & \textbf{0.026} & 0.010 & 0.605 & 0.664 & \textbf{0.605} & \textbf{0.664} \\
 & Block sparse & 0.018 & 0.006 & 0.026 & \textbf{0.009} & 0.606 & \textbf{0.666} & 0.603 & 0.660 \\
 & Tile sparse & 0.018 & 0.006 & 0.027 & 0.010 & 0.604 & 0.664 & 0.604 & 0.661 \\
 & Block sparse + landmark & 0.018 & 0.006 & 0.027 & 0.010 & \textbf{0.606} & 0.665 & 0.603 & 0.660 \\
\rowcolor{zebralavender}
 & Tile sparse + landmark & 0.017 & 0.006 & 0.027 & 0.011 & 0.605 & 0.664 & 0.605 & 0.662 \\
\addlinespace[2pt]
VGGT (0.7L/0.75S) & SGA & \textbf{0.016} & \textbf{0.006} & \textbf{0.023} & \textbf{0.007} & 0.607 & 0.668 & \textbf{0.604} & \textbf{0.661} \\
 & Block sparse & 0.018 & 0.006 & 0.025 & 0.008 & 0.607 & 0.668 & 0.602 & 0.658 \\
 & Tile sparse & 0.017 & 0.006 & 0.025 & 0.008 & 0.606 & 0.667 & 0.602 & 0.659 \\
 & Block sparse + landmark & 0.017 & 0.006 & 0.025 & 0.008 & \textbf{0.608} & \textbf{0.670} & 0.603 & 0.660 \\
\rowcolor{zebralavender}
 & Tile sparse + landmark & 0.017 & 0.006 & 0.025 & 0.008 & 0.607 & 0.669 & 0.604 & 0.661 \\
\addlinespace[2pt]
HYWM (0.5L/0.75S) & SGA & 0.013 & 0.005 & 0.024 & 0.009 & 0.608 & 0.669 & 0.609 & 0.669 \\
 & Block sparse & 0.014 & 0.005 & 0.024 & \textbf{0.008} & 0.606 & 0.667 & 0.606 & 0.665 \\
 & Tile sparse & 0.013 & 0.005 & 0.025 & 0.009 & 0.605 & 0.664 & 0.606 & 0.665 \\
 & Block sparse + landmark & 0.013 & 0.005 & \textbf{0.024} & 0.009 & \textbf{0.609} & \textbf{0.670} & \textbf{0.609} & \textbf{0.669} \\
\rowcolor{zebralavender}
 & Tile sparse + landmark & \textbf{0.013} & \textbf{0.005} & 0.025 & 0.010 & 0.608 & 0.668 & 0.608 & 0.668 \\
\addlinespace[2pt]
HYWM (0.7L/0.75S) & SGA & 0.013 & 0.005 & \textbf{0.022} & 0.008 & 0.609 & 0.671 & 0.609 & 0.669 \\
 & Block sparse & 0.013 & 0.005 & 0.023 & \textbf{0.008} & 0.609 & 0.671 & 0.608 & 0.669 \\
 & Tile sparse & 0.013 & 0.005 & 0.023 & 0.008 & 0.609 & 0.670 & 0.609 & 0.669 \\
 & Block sparse + landmark & \textbf{0.012} & \textbf{0.005} & 0.023 & 0.008 & \textbf{0.610} & \textbf{0.673} & 0.609 & 0.669 \\
\rowcolor{zebralavender}
 & Tile sparse + landmark & 0.013 & 0.005 & 0.023 & 0.009 & 0.609 & 0.671 & \textbf{0.609} & \textbf{0.670} \\
\addlinespace[2pt]
$\pi^3$ (0.5L/0.75S) & SGA & \textbf{0.011} & \textbf{0.004} & 0.016 & 0.005 & 0.605 & 0.665 & 0.600 & 0.655 \\
 & Block sparse & 0.024 & 0.010 & 0.017 & 0.004 & 0.625 & 0.700 & 0.586 & 0.632 \\
 & Tile sparse & 0.027 & 0.013 & \textbf{0.014} & \textbf{0.004} & \textbf{0.628} & \textbf{0.705} & 0.590 & 0.639 \\
 & Block sparse + landmark & 0.013 & 0.005 & 0.017 & 0.005 & 0.610 & 0.673 & 0.595 & 0.647 \\
\rowcolor{zebralavender}
 & Tile sparse + landmark & 0.012 & 0.004 & 0.018 & 0.007 & 0.606 & 0.665 & \textbf{0.602} & \textbf{0.658} \\
\addlinespace[2pt]
$\pi^3$ (0.7L/0.75S) & SGA & 0.022 & 0.009 & \textbf{0.018} & \textbf{0.005} & \textbf{0.617} & \textbf{0.686} & \textbf{0.585} & \textbf{0.631} \\
 & Block sparse & 0.024 & 0.010 & 0.021 & 0.006 & 0.613 & 0.678 & 0.583 & 0.628 \\
 & Tile sparse & \textbf{0.020} & 0.008 & 0.019 & 0.006 & 0.615 & 0.683 & 0.585 & 0.631 \\
 & Block sparse + landmark & 0.025 & 0.010 & 0.022 & 0.006 & 0.610 & 0.674 & 0.582 & 0.627 \\
\rowcolor{zebralavender}
 & Tile sparse + landmark & 0.020 & \textbf{0.008} & 0.020 & 0.006 & 0.614 & 0.680 & 0.585 & 0.631 \\
\bottomrule
\end{tabular}%
}
\end{table*}

\begin{table*}[t]
\caption{Sparse-attention ablation on Neural RGB-D.}
\label{tab:app-sparse-nrgbd}
\centering
\scriptsize
\setlength{\tabcolsep}{3pt}
\renewcommand{\arraystretch}{1.04}
\resizebox{\textwidth}{!}{%
\begin{tabular}{@{}llcccccccc@{}}
\toprule
\textbf{Configuration} & \textbf{Attention choice} & \textbf{Acc. mean $\downarrow$} & \textbf{Acc. med. $\downarrow$} & \textbf{Comp. mean $\downarrow$} & \textbf{Comp. med. $\downarrow$} & \textbf{NC1 mean $\uparrow$} & \textbf{NC1 med. $\uparrow$} & \textbf{NC2 mean $\uparrow$} & \textbf{NC2 med. $\uparrow$} \\
\midrule
VGGT (0.5L/0.75S) & SGA & 0.029 & 0.022 & 0.030 & 0.009 & 0.713 & 0.845 & 0.720 & 0.859 \\
 & Block sparse & 0.038 & 0.023 & 0.016 & 0.004 & 0.708 & 0.853 & 0.714 & 0.864 \\
 & Tile sparse & 0.039 & 0.034 & 0.020 & 0.006 & 0.708 & 0.847 & 0.707 & 0.854 \\
 & Block sparse + landmark & 0.034 & 0.020 & 0.015 & 0.004 & 0.734 & 0.884 & 0.747 & \textbf{0.898} \\
\rowcolor{zebralavender}
 & Tile sparse + landmark & \textbf{0.018} & \textbf{0.011} & \textbf{0.013} & \textbf{0.003} & \textbf{0.743} & \textbf{0.888} & \textbf{0.750} & 0.892 \\
\addlinespace[2pt]
VGGT (0.7L/0.75S) & SGA & 0.040 & 0.030 & 0.028 & 0.009 & 0.716 & 0.864 & 0.704 & 0.843 \\
 & Block sparse & 0.029 & 0.018 & \textbf{0.016} & 0.004 & 0.714 & 0.867 & 0.729 & 0.882 \\
 & Tile sparse & 0.030 & 0.017 & 0.018 & 0.005 & 0.718 & 0.865 & 0.719 & 0.863 \\
 & Block sparse + landmark & 0.031 & 0.018 & 0.017 & \textbf{0.004} & 0.717 & \textbf{0.872} & \textbf{0.731} & \textbf{0.889} \\
\rowcolor{zebralavender}
 & Tile sparse + landmark & \textbf{0.029} & \textbf{0.017} & 0.018 & 0.005 & \textbf{0.722} & 0.871 & 0.719 & 0.866 \\
\addlinespace[2pt]
HYWM (0.5L/0.75S) & SGA & 0.020 & 0.012 & 0.014 & \textbf{0.004} & 0.738 & 0.879 & 0.758 & 0.906 \\
 & Block sparse & 0.036 & 0.023 & 0.022 & 0.006 & 0.700 & 0.839 & 0.719 & 0.861 \\
 & Tile sparse & 0.032 & 0.021 & 0.018 & 0.005 & 0.701 & 0.846 & 0.701 & 0.849 \\
 & Block sparse + landmark & 0.022 & 0.014 & 0.014 & 0.005 & 0.736 & 0.879 & 0.753 & 0.904 \\
\rowcolor{zebralavender}
 & Tile sparse + landmark & \textbf{0.019} & \textbf{0.012} & \textbf{0.013} & 0.005 & \textbf{0.741} & \textbf{0.881} & \textbf{0.764} & \textbf{0.907} \\
\addlinespace[2pt]
HYWM (0.7L/0.75S) & SGA & 0.024 & 0.015 & \textbf{0.013} & \textbf{0.003} & 0.726 & 0.872 & 0.743 & \textbf{0.895} \\
 & Block sparse & 0.031 & 0.020 & 0.016 & 0.005 & 0.719 & 0.863 & 0.728 & 0.879 \\
 & Tile sparse & \textbf{0.022} & \textbf{0.014} & 0.013 & 0.004 & 0.719 & 0.861 & 0.729 & 0.879 \\
 & Block sparse + landmark & 0.032 & 0.020 & 0.016 & 0.005 & 0.721 & 0.866 & 0.734 & 0.884 \\
\rowcolor{zebralavender}
 & Tile sparse + landmark & 0.023 & 0.015 & 0.014 & 0.005 & \textbf{0.730} & \textbf{0.872} & \textbf{0.745} & 0.895 \\
\addlinespace[2pt]
$\pi^3$ (0.5L/0.75S) & SGA & 0.017 & 0.010 & 0.010 & 0.003 & 0.717 & 0.857 & 0.686 & 0.822 \\
 & Block sparse & 0.058 & 0.040 & 0.019 & 0.005 & 0.689 & 0.840 & 0.677 & 0.818 \\
 & Tile sparse & 0.066 & 0.047 & 0.016 & 0.004 & 0.695 & 0.854 & 0.677 & 0.825 \\
 & Block sparse + landmark & 0.024 & 0.014 & 0.012 & 0.003 & 0.721 & \textbf{0.866} & 0.702 & 0.844 \\
\rowcolor{zebralavender}
 & Tile sparse + landmark & \textbf{0.014} & \textbf{0.008} & \textbf{0.010} & \textbf{0.002} & \textbf{0.728} & 0.864 & \textbf{0.723} & \textbf{0.862} \\
\addlinespace[2pt]
$\pi^3$ (0.7L/0.75S) & SGA & 0.043 & 0.028 & \textbf{0.014} & 0.004 & 0.708 & 0.864 & 0.675 & 0.820 \\
 & Block sparse & 0.053 & 0.033 & 0.023 & 0.004 & 0.701 & 0.853 & 0.675 & 0.821 \\
 & Tile sparse & \textbf{0.042} & 0.027 & 0.016 & 0.004 & \textbf{0.709} & \textbf{0.864} & \textbf{0.677} & \textbf{0.823} \\
 & Block sparse + landmark & 0.054 & 0.031 & 0.022 & 0.004 & 0.702 & 0.855 & 0.675 & 0.819 \\
\rowcolor{zebralavender}
 & Tile sparse + landmark & 0.043 & \textbf{0.027} & 0.015 & \textbf{0.004} & 0.708 & 0.863 & 0.677 & 0.821 \\
\bottomrule
\end{tabular}%
}
\end{table*}

\begin{table*}[t]
\caption{Sparse-attention ablation on RealEstate10K camera pose.}
\label{tab:app-sparse-re10k}
\centering
\scriptsize
\setlength{\tabcolsep}{3pt}
\renewcommand{\arraystretch}{1.04}
\resizebox{\textwidth}{!}{%
\begin{tabular}{@{}llccccccc@{}}
\toprule
\textbf{Configuration} & \textbf{Attention choice} & \textbf{RRA-5 $\uparrow$} & \textbf{RRA-15 $\uparrow$} & \textbf{RRA-30 $\uparrow$} & \textbf{RTA-5 $\uparrow$} & \textbf{RTA-15 $\uparrow$} & \textbf{RTA-30 $\uparrow$} & \textbf{AUC-30 $\uparrow$} \\
\midrule
VGGT (0.5L/0.75S) & SGA & \textbf{0.949} & \textbf{0.996} & \textbf{0.999} & \textbf{0.519} & \textbf{0.841} & \textbf{0.924} & \textbf{0.739} \\
 & Block sparse & 0.945 & 0.993 & 0.998 & 0.466 & 0.811 & 0.906 & 0.709 \\
 & Tile sparse & 0.943 & 0.995 & 0.999 & 0.492 & 0.824 & 0.910 & 0.722 \\
 & Block sparse + landmark & 0.939 & 0.993 & 0.998 & 0.487 & 0.825 & 0.912 & 0.722 \\
\rowcolor{zebralavender}
 & Tile sparse + landmark & 0.938 & 0.995 & 0.999 & 0.516 & 0.834 & 0.917 & 0.733 \\
\addlinespace[2pt]
VGGT (0.7L/0.75S) & SGA & \textbf{0.943} & \textbf{0.996} & \textbf{1.000} & \textbf{0.471} & \textbf{0.836} & \textbf{0.925} & \textbf{0.727} \\
 & Block sparse & 0.939 & 0.995 & 0.999 & 0.442 & 0.819 & 0.914 & 0.710 \\
 & Tile sparse & 0.939 & 0.996 & 0.999 & 0.465 & 0.830 & 0.918 & 0.720 \\
 & Block sparse + landmark & 0.935 & 0.994 & 0.999 & 0.437 & 0.817 & 0.913 & 0.708 \\
\rowcolor{zebralavender}
 & Tile sparse + landmark & 0.933 & 0.995 & 1.000 & 0.464 & 0.830 & 0.919 & 0.721 \\
\addlinespace[2pt]
HYWM (0.5L/0.75S) & SGA & \textbf{0.985} & \textbf{0.998} & \textbf{1.000} & 0.745 & 0.903 & 0.949 & 0.843 \\
 & Block sparse & 0.984 & 0.997 & 0.998 & 0.727 & 0.897 & 0.944 & 0.833 \\
 & Tile sparse & 0.985 & 0.998 & 1.000 & 0.721 & 0.892 & 0.942 & 0.828 \\
 & Block sparse + landmark & 0.982 & 0.996 & 0.999 & 0.746 & 0.906 & 0.949 & 0.844 \\
\rowcolor{zebralavender}
 & Tile sparse + landmark & 0.985 & 0.998 & 0.999 & \textbf{0.763} & \textbf{0.908} & \textbf{0.951} & \textbf{0.852} \\
\addlinespace[2pt]
HYWM (0.7L/0.75S) & SGA & \textbf{0.985} & 0.998 & \textbf{1.000} & 0.722 & 0.896 & 0.945 & 0.830 \\
 & Block sparse & 0.982 & 0.996 & 0.999 & 0.713 & 0.895 & 0.942 & 0.826 \\
 & Tile sparse & 0.984 & \textbf{0.998} & 1.000 & 0.716 & 0.893 & 0.943 & 0.826 \\
 & Block sparse + landmark & 0.981 & 0.996 & 0.999 & 0.721 & 0.900 & 0.946 & 0.831 \\
\rowcolor{zebralavender}
 & Tile sparse + landmark & 0.984 & 0.998 & 1.000 & \textbf{0.744} & \textbf{0.904} & \textbf{0.949} & \textbf{0.841} \\
\addlinespace[2pt]
$\pi^3$ (0.5L/0.75S) & SGA & 0.986 & 0.998 & 1.000 & 0.767 & 0.908 & 0.947 & 0.847 \\
 & Block sparse & 0.987 & \textbf{0.998} & \textbf{1.000} & 0.689 & 0.880 & 0.933 & 0.808 \\
 & Tile sparse & 0.988 & 0.998 & 1.000 & 0.693 & 0.878 & 0.935 & 0.811 \\
 & Block sparse + landmark & 0.987 & 0.998 & 1.000 & 0.784 & \textbf{0.921} & \textbf{0.954} & 0.864 \\
\rowcolor{zebralavender}
 & Tile sparse + landmark & \textbf{0.988} & 0.998 & 1.000 & \textbf{0.795} & 0.919 & 0.954 & \textbf{0.868} \\
\addlinespace[2pt]
$\pi^3$ (0.7L/0.75S) & SGA & 0.969 & 0.993 & 0.997 & 0.621 & 0.826 & 0.896 & 0.759 \\
 & Block sparse & 0.978 & 0.998 & 1.000 & 0.684 & 0.888 & 0.944 & 0.815 \\
 & Tile sparse & 0.986 & \textbf{0.998} & \textbf{1.000} & 0.677 & 0.878 & 0.934 & 0.807 \\
 & Block sparse + landmark & 0.981 & 0.997 & 1.000 & 0.709 & \textbf{0.902} & \textbf{0.952} & 0.831 \\
\rowcolor{zebralavender}
 & Tile sparse + landmark & \textbf{0.986} & \textbf{0.998} & \textbf{1.000} & \textbf{0.727} & 0.901 & 0.948 & \textbf{0.837} \\
\bottomrule
\end{tabular}%
}
\end{table*}

\subsection{Knowledge-Distillation Ablation}
\label{app:kd-ablation}
\ruijing{Tables~\ref{tab:app-kd-pointmap} and~\ref{tab:app-kd-re10k}} ablate the 3,000-step
training objective for VGGT with the 0.5L/0.6S configuration. ``GT only'' uses
task supervision alone, and ``KD only'' uses teacher distillation without
ground-truth supervision. ``GT + output KD'' adds final-prediction matching
to the task loss, while ``GT + feature KD'' adds intermediate-feature matching.
RelKD combines task supervision with output, feature-L1, token-cosine, and
cross-view relational losses and is the complete Zebra3D-KD objective.
Lavender marks the reported Zebra3D-KD objective, and bold marks the best
result in each column\ruijing{ within each dataset}.

\begin{table*}[t]
\caption{Knowledge-distillation ablation on
7Scenes~\citep{shotton2013seven} and Neural
RGB-D~\citep{azinovic2022neuralrgbd}.}
\label{tab:app-kd-pointmap}
\centering
\scriptsize
\setlength{\tabcolsep}{3pt}
\renewcommand{\arraystretch}{1.04}
\resizebox{\textwidth}{!}{%
\begin{tabular}{@{}lcccccccc@{}}
\toprule
\textbf{Training objective} & \textbf{Acc. mean $\downarrow$}
& \textbf{Acc. med. $\downarrow$} & \textbf{Comp. mean $\downarrow$}
& \textbf{Comp. med. $\downarrow$} & \textbf{NC1 mean $\uparrow$}
& \textbf{NC1 med. $\uparrow$} & \textbf{NC2 mean $\uparrow$}
& \textbf{NC2 med. $\uparrow$} \\
\midrule
\multicolumn{9}{l}{\textbf{7Scenes}} \\
GT only & 0.023 & 0.008 & 0.023 & 0.007 & 0.612 & 0.678 & 0.597 & 0.650 \\
KD only & 0.023 & 0.010 & \textbf{0.020} & \textbf{0.004} & 0.611 & 0.674 & 0.580 & 0.623 \\
GT + output KD & 0.023 & 0.010 & 0.021 & 0.005 & \textbf{0.620} & \textbf{0.690} & 0.596 & 0.648 \\
GT + feature KD & 0.029 & 0.012 & 0.023 & 0.007 & 0.616 & 0.683 & 0.592 & 0.641 \\
\rowcolor{zebralavender}
RelKD & \textbf{0.015} & \textbf{0.005} & 0.026 & 0.011 & 0.608 & 0.668 & \textbf{0.606} & \textbf{0.665} \\
\midrule
\multicolumn{9}{l}{\textbf{Neural RGB-D}} \\
GT only & 0.046 & 0.027 & 0.018 & 0.005 & 0.699 & 0.848 & 0.685 & 0.828 \\
KD only & 0.045 & 0.031 & 0.014 & \textbf{0.004} & 0.696 & 0.839 & 0.687 & 0.825 \\
GT + output KD & 0.041 & 0.029 & 0.015 & \textbf{0.004} & 0.701 & 0.858 & 0.695 & 0.847 \\
GT + feature KD & 0.047 & 0.031 & \textbf{0.011} & \textbf{0.004} & 0.705 & 0.866 & 0.693 & 0.848 \\
\rowcolor{zebralavender}
RelKD & \textbf{0.024} & \textbf{0.013} & 0.018 & 0.005
& \textbf{0.751} & \textbf{0.888} & \textbf{0.762} & \textbf{0.896} \\
\bottomrule
\end{tabular}%
}
\end{table*}

\begin{table*}[t]
\caption{Knowledge-distillation ablation on RealEstate10K camera
pose~\citep{zhou2018stereomagnification}.}
\label{tab:app-kd-re10k}
\centering
\scriptsize
\setlength{\tabcolsep}{3pt}
\renewcommand{\arraystretch}{1.04}
\resizebox{\textwidth}{!}{%
\begin{tabular}{@{}lccccccc@{}}
\toprule
\textbf{Training objective} & \textbf{RRA-5 $\uparrow$}
& \textbf{RRA-15 $\uparrow$} & \textbf{RRA-30 $\uparrow$}
& \textbf{RTA-5 $\uparrow$} & \textbf{RTA-15 $\uparrow$}
& \textbf{RTA-30 $\uparrow$} & \textbf{AUC-30 $\uparrow$} \\
\midrule
GT only & \textbf{0.980} & 0.997 & 1.000 & 0.586 & 0.870 & 0.939 & 0.777 \\
KD only & 0.961 & 0.998 & 1.000 & 0.449 & 0.823 & 0.916 & 0.716 \\
GT + output KD & 0.973 & 0.997 & 0.999 & 0.488 & 0.838 & 0.922 & 0.736 \\
GT + feature KD & 0.977 & 0.998 & \textbf{1.000} & 0.542 & 0.854 & 0.929 & 0.756 \\
\rowcolor{zebralavender}
RelKD & 0.975 & \textbf{0.998} & 1.000 & \textbf{0.649}
& \textbf{0.895} & \textbf{0.950} & \textbf{0.805} \\
\bottomrule
\end{tabular}%
}
\end{table*}

%% file: tables/main_reconstruction_results.tex
\begin{table}[!htbp]
\caption{Full quantitative point-map reconstruction results on 7Scenes and
NRGBD at stride 10. Dense backbones are excluded when marking the best
(\textbf{bold}) and second-best (\underline{underlined}) non-dense results.
In the shorthand $x$L/$y$S, $x$ is the fraction of model layers using linear
attention and $y$ is the sparsity of attention rows in the remaining layers.
\ruijing{VGGT + GT-only FT is dense VGGT fine-tuned with the same data,
schedule, and ground-truth supervision as Zebra3D-KD.}}
\label{tab:main-reconstruction-results}
\centering
\small
\setlength{\tabcolsep}{3.2pt}
\renewcommand{\arraystretch}{0.96}
\resizebox{\textwidth}{!}{%
\begin{tabular}{@{}l@{\hspace{8pt}}c@{\hspace{8pt}}cc cc cc@{\hspace{12pt}}cc cc cc@{}}
\toprule
\textbf{Method}
& \shortstack{\textbf{Mean Attn.}\\\textbf{Sparsity} $\uparrow$}
& \multicolumn{6}{c}{\textbf{7Scenes}}
& \multicolumn{6}{c}{\textbf{NRGBD}} \\
\cmidrule(lr){3-8}\cmidrule(lr){9-14}
& & \multicolumn{2}{c}{\textbf{Acc} $\downarrow$} &
  \multicolumn{2}{c}{\textbf{Comp} $\downarrow$} &
  \multicolumn{2}{c}{\textbf{NC1} $\uparrow$}
& \multicolumn{2}{c}{\textbf{Acc} $\downarrow$} &
  \multicolumn{2}{c}{\textbf{Comp} $\downarrow$} &
  \multicolumn{2}{c}{\textbf{NC1} $\uparrow$} \\
& & \textit{Mean} & \textit{Med.} & \textit{Mean} & \textit{Med.} &
  \textit{Mean} & \textit{Med.}
& \textit{Mean} & \textit{Med.} & \textit{Mean} & \textit{Med.} &
  \textit{Mean} & \textit{Med.} \\
\midrule
\rowcolor{modelgray}
VGGT & 0\% & 0.017 & 0.005 & 0.027 & 0.011 & 0.607 & 0.667 & 0.016 & 0.009 & 0.011 & 0.003 & 0.761 & 0.901 \\
\rowcolor{modelgray}
\ruijing{VGGT + GT-only FT} & \ruijing{0\%} & \ruijing{0.024} & \ruijing{0.008} & \ruijing{0.031} & \ruijing{0.014} & \ruijing{0.604} & \ruijing{0.662} & \ruijing{0.024} & \ruijing{0.013} & \ruijing{0.018} & \ruijing{0.006} & \ruijing{0.754} & \ruijing{0.890} \\
FastVGGT & 88\% & \textbf{0.015} & \textbf{0.005} & 0.025 & 0.009 & 0.607 & 0.667 & \underline{0.020} & \underline{0.013} & \textbf{0.013} & \underline{0.004} & 0.731 & 0.871 \\
SAF3R & 84\% & 0.018 & 0.008 & \textbf{0.022} & \textbf{0.005} & \underline{0.611} & \underline{0.675} & 0.054 & 0.034 & 0.022 & 0.005 & 0.694 & 0.847 \\
RegimeVGGT & \underline{90\%} & 0.016 & \textbf{0.005} & 0.026 & 0.010 & 0.606 & 0.667 & 0.024 & 0.014 & 0.016 & 0.005 & 0.736 & 0.874 \\
Block Sparse 3D & 70\% & 0.017 & \underline{0.006} & 0.025 & \underline{0.007} & 0.608 & 0.670 & 0.042 & 0.029 & \underline{0.016} & \underline{0.004} & 0.696 & 0.837 \\
Lite3R & 70\% & 0.039 & 0.020 & \underline{0.023} & \textbf{0.005} & \textbf{0.619} & \textbf{0.688} & 0.068 & 0.041 & 0.018 & 0.005 & 0.689 & 0.843 \\
A3D & 83\% & \underline{0.016} & \textbf{0.005} & 0.026 & 0.010 & 0.607 & 0.668 & 0.025 & 0.014 & 0.017 & 0.005 & 0.735 & 0.877 \\
\rowcolor{zebralavender}
\textbf{Zebra3D (0.5L/0.6S)} & 79\% & 0.017 & \underline{0.006} & 0.027 & 0.012 & 0.606 & 0.665 & \textbf{0.018} & \textbf{0.011} & \textbf{0.013} & \textbf{0.003} & \underline{0.746} & \textbf{0.891} \\
\rowcolor{zebralavender}
\textbf{Zebra3D-KD (0.5L/0.6S)} & 79\% & \textbf{0.015} & \textbf{0.005} & 0.026 & 0.011 & 0.608 & 0.668 & 0.024 & \underline{0.013} & 0.018 & 0.005 & \textbf{0.751} & \underline{0.888} \\
\rowcolor{zebralavender}
\textbf{Zebra3D-KD (0.7L/0.8S)} & \textbf{94\%} & \underline{0.016} & \underline{0.006} & 0.025 & 0.011 & 0.608 & 0.669 & 0.025 & 0.015 & 0.019 & 0.006 & 0.739 & 0.879 \\
\midrule
\rowcolor{modelgray}
HYWM & 0\% & 0.013 & 0.005 & 0.024 & 0.010 & 0.608 & 0.668 & 0.018 & 0.011 & 0.014 & 0.004 & 0.750 & 0.890 \\
Block Sparse 3D & 70\% & \underline{0.014} & \underline{0.006} & 0.024 & \textbf{0.007} & \underline{0.606} & 0.667 & 0.041 & 0.026 & \underline{0.023} & 0.006 & 0.688 & 0.828 \\
A3D & 83\% & \textbf{0.013} & \textbf{0.005} & 0.024 & 0.009 & \textbf{0.608} & \textbf{0.669} & \underline{0.019} & \textbf{0.011} & \textbf{0.014} & \textbf{0.004} & 0.740 & 0.882 \\
RegimeVGGT & \underline{90\%} & \textbf{0.013} & \textbf{0.005} & 0.024 & 0.009 & 0.607 & 0.667 & 0.019 & 0.011 & \textbf{0.014} & \underline{0.004} & 0.734 & 0.873 \\
\rowcolor{zebralavender}
\textbf{Zebra3D (0.5L/0.6S)} & 79\% & \textbf{0.013} & \textbf{0.005} & 0.025 & 0.010 & \textbf{0.608} & \textbf{0.669} & \underline{0.019} & \underline{0.012} & \textbf{0.014} & \underline{0.005} & \underline{0.744} & \underline{0.883} \\
\rowcolor{zebralavender}
\textbf{Zebra3D-KD (0.5L/0.6S)} & 79\% & \textbf{0.013} & \textbf{0.005} & \underline{0.023} & 0.009 & \textbf{0.608} & \textbf{0.669} & \textbf{0.018} & \textbf{0.011} & \textbf{0.014} & \underline{0.005} & \textbf{0.749} & \textbf{0.887} \\
\rowcolor{zebralavender}
\textbf{Zebra3D-KD (0.7L/0.8S)} & \textbf{94\%} & \underline{0.014} & \textbf{0.005} & \textbf{0.022} & \underline{0.008} & \textbf{0.608} & \underline{0.668} & \underline{0.019} & 0.013 & \textbf{0.014} & \underline{0.005} & 0.737 & 0.877 \\
\midrule
\rowcolor{modelgray}
$\pi^3$ & 0\% & 0.012 & 0.004 & 0.017 & 0.007 & 0.607 & 0.668 & 0.014 & 0.007 & 0.009 & 0.003 & 0.737 & 0.872 \\
Block Sparse 3D & 70\% & 0.020 & 0.008 & \textbf{0.016} & \textbf{0.004} & \textbf{0.622} & \textbf{0.695} & 0.049 & 0.033 & 0.017 & 0.004 & 0.693 & 0.844 \\
A3D & \underline{88\%} & \textbf{0.012} & \textbf{0.004} & \textbf{0.016} & \underline{0.005} & 0.607 & 0.667 & 0.018 & 0.010 & \underline{0.010} & \underline{0.003} & 0.716 & 0.857 \\
\rowcolor{zebralavender}
\textbf{Zebra3D (0.5L/0.6S)} & 79\% & \textbf{0.012} & \textbf{0.004} & \underline{0.018} & 0.007 & 0.606 & 0.666 & \textbf{0.014} & \textbf{0.007} & \textbf{0.009} & \textbf{0.002} & \textbf{0.731} & \textbf{0.866} \\
\rowcolor{zebralavender}
\textbf{Zebra3D (0.7L/0.8S)} & \textbf{94\%} & 0.021 & 0.008 & 0.020 & 0.006 & \underline{0.613} & \underline{0.679} & 0.043 & 0.027 & 0.015 & 0.004 & 0.708 & \underline{0.863} \\
\rowcolor{zebralavender}
\textbf{Zebra3D-KD (0.5L/0.6S)} & 79\% & \textbf{0.012} & \textbf{0.004} & \underline{0.018} & 0.007 & 0.604 & 0.663 & \underline{0.015} & \underline{0.008} & \textbf{0.009} & \textbf{0.002} & \underline{0.728} & 0.861 \\
\rowcolor{zebralavender}
\ruijing{\textbf{Zebra3D-KD (0.7L/0.8S)}} & \ruijing{\textbf{94\%}} & \ruijing{\underline{0.013}} & \ruijing{\underline{0.005}} & \ruijing{\underline{0.018}} & \ruijing{0.006} & \ruijing{0.605} & \ruijing{0.665} & \ruijing{0.017} & \ruijing{0.009} & \ruijing{\underline{0.010}} & \ruijing{\underline{0.003}} & \ruijing{0.718} & \ruijing{0.854} \\
\bottomrule
\end{tabular}%
}
\end{table}

%% file: tables/main_camera_pose_results.tex
\begin{table}[!htbp]
\caption{Full camera-pose estimation results on RealEstate10K and CO3Dv2 using
10 input frames. Higher is better. Dense backbones are excluded when marking the
best (\textbf{bold}) and second-best (\underline{underlined}) non-dense
results; dashes denote unavailable measurements. In the shorthand $x$L/$y$S,
$x$ is the fraction of model layers using linear attention and $y$ is the
sparsity of attention rows in the remaining layers.
\ruijing{VGGT + GT-only FT is dense VGGT fine-tuned with the same data,
schedule, and ground-truth supervision as Zebra3D-KD.}}
\label{tab:main-camera-pose-results}
\centering
\small
\setlength{\tabcolsep}{3.2pt}
\renewcommand{\arraystretch}{0.96}
\resizebox{\textwidth}{!}{%
\begin{tabular}{@{}l@{\hspace{8pt}}c@{\hspace{8pt}}ccccccc@{\hspace{12pt}}cccc@{}}
\toprule
\textbf{Method}
& \shortstack{\textbf{Mean Attn.}\\\textbf{Sparsity} $\uparrow$}
& \multicolumn{7}{c}{\textbf{RealEstate10K Camera Pose}}
& \multicolumn{4}{c}{\textbf{CO3Dv2 Camera Pose}} \\
\cmidrule(lr){3-9}\cmidrule(lr){10-13}
& & \textbf{RRA-5} $\uparrow$ & \textbf{RRA-15} $\uparrow$
& \textbf{RRA-30} $\uparrow$ & \textbf{RTA-5} $\uparrow$
& \textbf{RTA-15} $\uparrow$ & \textbf{RTA-30} $\uparrow$
& \textbf{AUC-30} $\uparrow$ & \textbf{AUC-3} $\uparrow$
& \textbf{AUC-5} $\uparrow$ & \textbf{AUC-15} $\uparrow$
& \textbf{AUC-30} $\uparrow$ \\
\midrule
\rowcolor{modelgray}
VGGT & 0\% & 0.947 & 0.996 & 0.999 & 0.543 & 0.843 & 0.920 & 0.744 & 0.566 & 0.675 & 0.839 & 0.900 \\
\rowcolor{modelgray}
\ruijing{VGGT + GT-only FT} & \ruijing{0\%} & \ruijing{0.980} & \ruijing{0.998} & \ruijing{1.000} & \ruijing{0.674} & \ruijing{0.901} & \ruijing{0.951} & \ruijing{0.816} & \ruijing{0.474} & \ruijing{0.609} & \ruijing{0.812} & \ruijing{0.885} \\
FastVGGT & 83\% & 0.936 & 0.994 & 0.999 & 0.476 & 0.808 & 0.904 & 0.708 & 0.412 & 0.555 & 0.784 & 0.867 \\
SAF3R & 78\% & 0.946 & 0.995 & 0.999 & 0.493 & 0.809 & 0.902 & 0.714 & \underline{0.504} & \underline{0.626} & \underline{0.814} & \underline{0.883} \\
RegimeVGGT & \underline{87\%} & 0.939 & 0.995 & 0.999 & 0.498 & 0.824 & 0.911 & 0.722 & 0.445 & 0.582 & 0.797 & 0.876 \\
Block Sparse 3D & 70\% & 0.937 & 0.994 & 0.998 & 0.379 & 0.745 & 0.869 & 0.649 & 0.207 & 0.390 & 0.721 & 0.835 \\
Lite3R & 70\% & 0.933 & \underline{0.996} & 0.999 & 0.253 & 0.666 & 0.843 & 0.578 & 0.078 & 0.220 & 0.621 & 0.779 \\
A3D & 75\% & 0.952 & \underline{0.996} & 0.999 & 0.536 & 0.839 & 0.919 & 0.742 & 0.468 & 0.602 & 0.806 & 0.881 \\
\rowcolor{zebralavender}
\textbf{Zebra3D (0.5L/0.6S)} & 79\% & 0.941 & \underline{0.996} & 0.999 & 0.526 & 0.839 & 0.919 & 0.737 & \textbf{0.554} & \textbf{0.665} & \textbf{0.834} & \textbf{0.897} \\
\rowcolor{zebralavender}
\textbf{Zebra3D-KD (0.5L/0.6S)} & 79\% & \textbf{0.975} & \textbf{0.998} & \underline{1.000} & \underline{0.649} & \textbf{0.895} & \textbf{0.950} & \underline{0.805} & 0.458 & 0.598 & 0.808 & \underline{0.883} \\
\rowcolor{zebralavender}
\textbf{Zebra3D-KD (0.7L/0.8S)} & \textbf{94\%} & \underline{0.975} & \textbf{0.998} & \textbf{1.000} & \textbf{0.654} & \underline{0.894} & \underline{0.947} & \textbf{0.805} & 0.403 & 0.556 & 0.789 & 0.872 \\
\midrule
\rowcolor{modelgray}
HYWM & 0\% & 0.986 & 0.998 & 1.000 & 0.774 & 0.915 & 0.955 & 0.860 & 0.403 & 0.544 & 0.777 & 0.865 \\
Block Sparse 3D & 70\% & 0.981 & 0.996 & 0.998 & 0.573 & 0.810 & 0.890 & 0.735 & 0.085 & 0.199 & 0.590 & 0.761 \\
A3D & 75\% & 0.984 & 0.998 & 0.999 & 0.740 & 0.900 & 0.947 & 0.839 & 0.323 & 0.474 & 0.743 & 0.844 \\
RegimeVGGT & \underline{87\%} & 0.985 & \textbf{0.998} & \textbf{1.000} & 0.741 & 0.902 & 0.947 & 0.840 & 0.315 & 0.468 & 0.740 & 0.843 \\
\rowcolor{zebralavender}
\textbf{Zebra3D (0.5L/0.6S)} & 79\% & \textbf{0.986} & 0.997 & 0.999 & \textbf{0.769} & \underline{0.911} & \underline{0.953} & \textbf{0.856} & \textbf{0.400} & \textbf{0.541} & \textbf{0.774} & \textbf{0.862} \\
\rowcolor{zebralavender}
\textbf{Zebra3D-KD (0.5L/0.6S)} & 79\% & \underline{0.986} & \underline{0.998} & \underline{1.000} & \underline{0.766} & \textbf{0.913} & \textbf{0.954} & \underline{0.854} & \underline{0.332} & \underline{0.481} & \underline{0.748} & \underline{0.849} \\
\rowcolor{zebralavender}
\textbf{Zebra3D-KD (0.7L/0.8S)} & \textbf{94\%} & 0.984 & 0.998 & 0.999 & 0.752 & 0.907 & 0.949 & 0.847 & 0.295 & 0.447 & 0.731 & 0.838 \\
\midrule
\rowcolor{modelgray}
$\pi^3$ & 0\% & 0.988 & 0.998 & 1.000 & 0.809 & 0.926 & 0.958 & 0.877 & 0.582 & 0.687 & 0.844 & 0.903 \\
Block Sparse 3D & 70\% & \underline{0.987} & \textbf{0.998} & \textbf{1.000} & 0.655 & 0.855 & 0.918 & 0.786 & 0.213 & 0.387 & 0.720 & 0.835 \\
A3D & \underline{82\%} & 0.985 & \textbf{0.998} & \textbf{1.000} & 0.765 & \underline{0.909} & \underline{0.950} & 0.847 & 0.457 & 0.593 & 0.804 & 0.880 \\
\rowcolor{zebralavender}
\textbf{Zebra3D (0.5L/0.6S)} & 79\% & \textbf{0.988} & \textbf{0.998} & \textbf{1.000} & \textbf{0.800} & \textbf{0.921} & \textbf{0.955} & \textbf{0.871} & \textbf{0.563} & \textbf{0.673} & \textbf{0.839} & \textbf{0.900} \\
\rowcolor{zebralavender}
\textbf{Zebra3D (0.7L/0.8S)} & \textbf{94\%} & 0.985 & \textbf{0.998} & \textbf{1.000} & 0.726 & 0.901 & 0.948 & 0.836 & 0.256 & 0.425 & 0.731 & 0.841 \\
\rowcolor{zebralavender}
\textbf{Zebra3D-KD (0.5L/0.6S)} & 79\% & \textbf{0.988} & \textbf{0.998} & \textbf{1.000} & \underline{0.773} & 0.907 & 0.948 & \underline{0.855} & \underline{0.527} & \underline{0.649} & \underline{0.829} & \underline{0.895} \\
\rowcolor{zebralavender}
\ruijing{\textbf{Zebra3D-KD (0.7L/0.8S)}} & \ruijing{\textbf{94\%}} & \ruijing{\underline{0.987}} & \ruijing{\textbf{0.998}} & \ruijing{\textbf{1.000}} & \ruijing{0.751} & \ruijing{0.901} & \ruijing{0.945} & \ruijing{0.844} & \ruijing{0.440} & \ruijing{0.581} & \ruijing{0.797} & \ruijing{0.876} \\
\bottomrule
\end{tabular}%
}
\end{table}

%% file: tables/appendix_wide_baseline_results.tex
\subsection{Wide-Baseline Point-Map Reconstruction}
\label{app:wide-baseline}
The benchmarks of Section~\ref{sec:experiments} are dominated by video-like
capture: 7Scenes, NRGBD, and ScanNet-50 are handheld sequences, and
RealEstate10K and CO3Dv2 supply densely sampled trajectories.  Consecutive
views therefore share a narrow baseline, the setting in which
attention is most likely to concentrate on a few neighboring frames.  This
appendix evaluates the converted models outside that setting, on
Tanks and Temples~\citep{knapitsch2017tanks} (large-scale outdoor and indoor
scenes) and ETH3D~\citep{schoeps2017eth3d} (indoor and outdoor DSLR captures),
with DTU~\citep{anaes2016dtu} retained as an object-centric turntable control.
\ruijing{Table~\ref{tab:wide-baseline} averages} over
scenes and over view counts $N\in\{8,16,32,64\}$, so each entry spans both
sparse and moderately dense coverage of the same scene.  All rows share the
inputs, preprocessing, and metric definitions of
Appendix~\ref{app:implementation-details}, and \ruijing{each converted model is
compared against that backbone's own dense model on the same scenes}, which
isolates the cost of conversion from the backbone's absolute quality.

\paragraph{Conversion cost at the $0.5$L operating points.}
Normal consistency is the metric most sensitive to fine-grained
correspondence, and it is where a widened baseline should hurt first.  Across
all three benchmarks and all three backbones, every $0.5$L configuration stays
within $3\%$ of its dense backbone in NC1; the largest gap is $2.9\%$ (HYWM
$0.5$L/$0.8$S on Tanks and Temples, $0.6790$ versus $0.6992$), and most are
inside $2\%$.  Point-map errors follow the same pattern and occasionally
improve on dense: on ETH3D, every VGGT $0.5$L configuration beats dense in both accuracy
and completeness, with Zebra3D-KD $0.5$L/$0.6$S reaching $0.2438$\,m accuracy
and $0.1816$\,m completeness against $0.2898$\,m and $0.3321$\,m for dense
VGGT.  Tanks and Temples is the hardest of the three, and the damage is
concentrated in completeness on HYWM ($+0.53$\,m at $0.5$L/$0.6$S, reduced to
$+0.04$\,m after distillation), while the same operating point stays within
$+0.06$\,m on VGGT and $+0.03$\,m on $\pi^3$.

\paragraph{Where the budget does break.}
The $0.7$L configurations are less forgiving here than on the video benchmarks,
and $\pi^3$ degrades most: $0.7$L/$0.8$S loses $0.076$ NC1 on Tanks and
Temples ($10.4\%$ relative) and $0.054$ on ETH3D ($6.4\%$), and on DTU its
accuracy grows from $1.0550$\,mm to $1.9583$\,mm at $0.7$L/$0.6$S.  This is
consistent with the placement analysis of Section~\ref{sec:placement-ablation}:
$\pi^3$ has the latest sensitive band, so raising the linear fraction
encroaches on it soonest.  We therefore regard $0.5$L as the operating point
that transfers to wide baselines, and $0.7$L as a speed-first setting
validated mainly on the video-like benchmarks.

\begin{table}[t]
\centering
\footnotesize
\setlength{\tabcolsep}{3pt}
\caption{Wide-baseline point-map accuracy on Tanks and Temples (large-scale
outdoor and indoor scenes) and ETH3D (indoor and outdoor DSLR captures), with
DTU (object-centric turntable rig) retained as a narrow-baseline control.
Each entry is averaged over scenes and view counts $N\in\{8,16,32,64\}$.
Accuracy and completeness are lower-is-better, in m for Tanks and Temples and
ETH3D and in mm for DTU; NC1 is higher is better.  Dense rows are excluded
when marking the best (\textbf{bold}) result in each column within a backbone
block.}
\label{tab:wide-baseline}
\resizebox{\textwidth}{!}{%
\begin{tabular}{@{}l ccc ccc ccc@{}}
\toprule
& \multicolumn{3}{c}{\textbf{Tanks and Temples}}
& \multicolumn{3}{c}{\textbf{ETH3D}}
& \multicolumn{3}{c}{\textbf{DTU}} \\
\cmidrule(lr){2-4}\cmidrule(lr){5-7}\cmidrule(lr){8-10}
\textbf{Method}
& \textbf{Acc} $\downarrow$ & \textbf{Comp} $\downarrow$ & \textbf{NC1} $\uparrow$
& \textbf{Acc} $\downarrow$ & \textbf{Comp} $\downarrow$ & \textbf{NC1} $\uparrow$
& \textbf{Acc} $\downarrow$ & \textbf{Comp} $\downarrow$ & \textbf{NC1} $\uparrow$ \\
\midrule
\multicolumn{10}{l}{\textbf{HYWM}} \\
Dense & 0.9865 & 0.4437 & 0.6992 & 0.2591 & 0.1857 & 0.8198 & 0.9622 & 1.4920 & 0.6549 \\
A3D & 1.4735 & 1.7308 & 0.6681 & 0.4634 & 0.4119 & 0.7723 & 0.9990 & \textbf{1.4659} & 0.6512 \\
\textbf{Zebra3D (0.5L/0.6S)} & 1.2380 & 0.9749 & 0.6817 & \textbf{0.2437} & \textbf{0.1681} & \textbf{0.8191} & \textbf{0.9563} & 1.5110 & \textbf{0.6565} \\
\textbf{Zebra3D (0.5L/0.8S)} & 1.3271 & 0.9866 & 0.6790 & 0.2629 & 0.1830 & 0.8151 & 0.9766 & 1.5020 & 0.6559 \\
\textbf{Zebra3D (0.7L/0.6S)} & \textbf{1.1272} & 0.8316 & 0.6795 & 0.2829 & 0.1857 & 0.8059 & 1.0430 & 1.4766 & 0.6557 \\
\textbf{Zebra3D (0.7L/0.8S)} & 1.4817 & 1.6450 & 0.6678 & 0.3073 & 0.2013 & 0.8034 & 1.0691 & 1.5129 & 0.6551 \\
\textbf{Zebra3D-KD (0.5L/0.6S)} & 1.1364 & \textbf{0.4831} & \textbf{0.6889} & 0.2989 & 0.2242 & 0.8101 & 1.1988 & 1.8062 & 0.6549 \\
\textbf{Zebra3D-KD (0.7L/0.8S)} & 1.4398 & 0.7431 & 0.6779 & 0.3017 & 0.2159 & 0.8052 & 1.2152 & 1.7885 & 0.6543 \\
\midrule
\multicolumn{10}{l}{\textbf{VGGT}} \\
Dense & 0.7303 & 0.6909 & 0.7231 & 0.2898 & 0.3321 & 0.8056 & 1.1153 & 1.9545 & 0.6768 \\
A3D & 0.8312 & 0.8278 & 0.6965 & 0.4995 & 0.5108 & 0.7657 & 1.1285 & 1.9411 & 0.6675 \\
\textbf{Zebra3D (0.5L/0.6S)} & \textbf{0.6977} & 0.7481 & \textbf{0.7184} & 0.2605 & 0.2606 & 0.8097 & 1.1165 & 1.9084 & \textbf{0.6739} \\
\textbf{Zebra3D (0.5L/0.8S)} & 0.7732 & 0.8653 & 0.7069 & 0.2513 & 0.2307 & 0.8109 & \textbf{1.1162} & 1.8913 & 0.6720 \\
\textbf{Zebra3D (0.7L/0.6S)} & 0.7083 & \textbf{0.7074} & 0.7062 & 0.2540 & 0.2302 & 0.7994 & 1.2368 & 1.8959 & 0.6610 \\
\textbf{Zebra3D (0.7L/0.8S)} & 0.7441 & 0.7740 & 0.6993 & 0.2499 & 0.2235 & 0.7996 & 1.2465 & 1.8868 & 0.6600 \\
\textbf{Zebra3D-KD (0.5L/0.6S)} & 0.7878 & 0.8131 & 0.7060 & \textbf{0.2438} & \textbf{0.1816} & \textbf{0.8222} & 1.1562 & \textbf{1.8787} & 0.6703 \\
\textbf{Zebra3D-KD (0.7L/0.8S)} & 0.7954 & 0.7609 & 0.6934 & 0.2595 & 0.1942 & 0.8173 & 1.1552 & 1.8844 & 0.6606 \\
\midrule
\multicolumn{10}{l}{\textbf{$\pi^3$}} \\
Dense & 0.7478 & 0.4910 & 0.7311 & 0.1281 & 0.0811 & 0.8472 & 1.0550 & 1.5950 & 0.6463 \\
A3D & 1.1478 & 0.8858 & 0.6832 & 0.3332 & 0.2416 & 0.7807 & 1.3521 & 1.6462 & 0.6530 \\
\textbf{Zebra3D (0.5L/0.6S)} & \textbf{0.7488} & 0.5141 & \textbf{0.7276} & \textbf{0.1322} & \textbf{0.0843} & \textbf{0.8424} & \textbf{1.1458} & 1.6189 & 0.6502 \\
\textbf{Zebra3D (0.5L/0.8S)} & 0.7812 & 0.5679 & 0.7187 & 0.1458 & 0.0957 & 0.8356 & 1.1807 & 1.6009 & 0.6505 \\
\textbf{Zebra3D (0.7L/0.6S)} & 0.9643 & 0.6009 & 0.6627 & 0.2401 & 0.1484 & 0.7956 & 1.9583 & 1.7168 & 0.6595 \\
\textbf{Zebra3D (0.7L/0.8S)} & 1.1376 & 0.9781 & 0.6551 & 0.2448 & 0.1526 & 0.7935 & 2.0262 & 1.7319 & \textbf{0.6607} \\
\textbf{Zebra3D-KD (0.5L/0.6S)} & 0.9205 & \textbf{0.4656} & 0.7141 & 0.1463 & 0.0941 & 0.8335 & 1.1493 & 1.6234 & 0.6454 \\
\textbf{Zebra3D-KD (0.7L/0.8S)} & 1.1929 & 0.7778 & 0.6890 & 0.1966 & 0.1246 & 0.8196 & 1.2652 & \textbf{1.5940} & 0.6516 \\
\bottomrule
\end{tabular}%
}
\end{table}